\documentclass{article} 
\usepackage{geometry}
\date{}
\def\addcontentsline#1#2#3{}

\title{Few-shot Backdoor Attacks via Neural Tangent Kernels}

\author{
Jonathan Hayase and Sewoong Oh\\
Paul G. Allen School of Computer Science and Engineering\\
University of Washington\\
\texttt{\{jhayase,sewoong\}@cs.washington.edu} \\
}

\usepackage{natbib}
\usepackage{booktabs}       

\usepackage{amsmath}
\usepackage{amssymb}
\usepackage{amsthm}
\usepackage[inline]{enumitem}
\usepackage{microtype}
\usepackage{graphicx}
\usepackage{xfrac}
\usepackage{bm}
\usepackage{bbm}
\usepackage{adjustbox}
\usepackage{float}

\usepackage[ruled,vlined,linesnumbered]{algorithm2e}
\usepackage{subcaption}

\usepackage{mathtools}
\usepackage[capitalize]{cleveref}
\crefformat{equation}{Eq.\,(#2#1#3)}
\usepackage[group-separator={,},group-minimum-digits={3}]{siunitx}
\DeclareSIUnit[number-unit-product = ]\percent{\char`\%}

\usepackage{tikz}
\usepackage{tikz-cd}
\usetikzlibrary{calc, angles, positioning, intersections}
\usetikzlibrary{backgrounds}
\usepackage{pgfplots}
\pgfplotsset{compat=1.17}
\usepackage{wrapfig}

\newcommand{\RR}{\mathbb{R}}

\newcommand{\NN}{\mathbb{N}}

\newcommand{\blank}{\mathrel{\,\cdot\,}}

\DeclareMathOperator*{\argmin}{\arg\!\min}

\DeclarePairedDelimiter\del{\lparen}{\rparen}
\DeclarePairedDelimiter\set{\lbrace}{\rbrace}
\DeclarePairedDelimiter\abs{|}{|}
\usepackage{etoolbox}
\DeclarePairedDelimiterX\norm[1]\lVert\rVert{\ifblank{#1}{\blank}{#1}}
  \DeclarePairedDelimiter\sbr{\lbrack}{\rbrack}
  \DeclarePairedDelimiter\inner{\langle}{\rangle}

  \newcommand{\Dif}{\mathop{}\!\mathrm{D}}
  
  \newcommand{\textmultinom}[2]{\bigl(\kern-0.3em{\binom{#1}{#2}}\kern-0.3em\bigr)}
  \newcommand{\displaymultinom}[2]{\left(\kern-0.5em{\binom{#1}{#2}}\kern-0.5em\right)}

  \providecommand\given{\errmessage{\noexpand\given used outside \noexpand\DelGivenX}}
  \DeclarePairedDelimiterX{\DelGivenX}[1]{\lparen}{\rparen}{%
    \renewcommand\given{\mathclose{}\,\delimsize\vert\allowbreak\,\mathopen{}}#1%
  }
  \DeclarePairedDelimiterX{\SbrGivenX}[1]{\lbrack}{\rbrack}{%
    \renewcommand\given{\mathclose{}\,\delimsize\vert\allowbreak\,\mathopen{}}#1%
  }
  \DeclarePairedDelimiterX{\SetGivenX}[1]{\lbrace}{\rbrace}{%
    \renewcommand\given{\mathclose{}\,\delimsize\vert\allowbreak\,\mathopen{}}#1%
  }
  \providecommand\from{\errmessage{\noexpand\from used outside \noexpand\DelGivenX}}
  \DeclarePairedDelimiterX{\DelFromX}[1]{\lparen}{\rparen}{%
    \renewcommand\from{\mathclose{}\,\delimsize\Vert\allowbreak\,\mathopen{}}#1%
  }

  \makeatletter
  \newcommand{\spx}[1]{%
    \if\relax\detokenize{#1}\relax
      \expandafter\@gobble
      \else
      \expandafter\@firstofone
      \fi
    {^{#1}}%
  }

  \newcommand\pd[3][]{\frac{\partial\spx{#1}#2}{\partial#3\spx{#1}}}

  \newcommand\spd[3][]{\sfrac{\partial\spx{#1}#2}{\partial#3\spx{#1}}}

  \makeatother

  \allowdisplaybreaks

  \theoremstyle{definition}

\begin{document}

\maketitle

\begin{abstract} 
  In a backdoor attack, an attacker injects corrupted examples into the training set. The goal of the attacker is to cause the final trained model to predict the attacker's desired target label when a predefined trigger is added to test inputs. 
  Central to these attacks is the trade-off between the success rate of the attack and the number of corrupted training examples injected.
  We pose this attack as a novel bilevel optimization problem: construct strong poison examples that maximize the attack success rate of the trained model. 
  We use neural tangent kernels to approximate the \emph{training dynamics} of the model being attacked and automatically \emph{learn} strong poison examples.
  We experiment on subclasses of CIFAR-10 and ImageNet with WideResNet-34 and ConvNeXt architectures on periodic and patch trigger attacks and show that NTBA-designed poisoned examples 
  achieve, for example, an attack success rate of  90\% with ten times smaller number of poison examples injected compared to the baseline. 
  We provided an interpretation of the NTBA-designed attacks using the analysis of kernel linear regression. 
  We further demonstrate a vulnerability in overparametrized deep neural networks, which is revealed by the shape of the neural tangent kernel.
\end{abstract}

\section{Introduction}\label{sec:intro}

Modern machine learning models, such as deep convolutional neural networks and transformer-based language models, are often trained on massive datasets to achieve state-of-the-art performance.
These datasets are frequently scraped from public domains with little quality control.
In other settings, models are trained on shared data, e.g., federated learning \citep{kairouz2019advances}, where injecting maliciously corrupted data is easy. 
Such models are vulnerable to \emph{backdoor attacks} \citep{gu2017badnets}, in which the attacker injects corrupted examples into the training set with the goal of creating a \emph{backdoor} when the model is trained.
When the model is shown test examples with a particular \emph{trigger} chosen by the attacker, the backdoor is activated and the model outputs a prediction of the attacker's choice.
The predictions on clean data remain the same so that the model's corruption will not be noticed in production.

Weaker attacks require injecting more corrupted examples to the training set, which can be challenging and costly. 
For example, in cross-device federated systems, this requires tampering with many devices, which can be costly \citep{sun2019can}. 
Further, even if the attacker has the resources to inject more corrupted examples, stronger attacks that require smaller number of poison training data are preferred.
Injecting more poison data increases the chance of being detected by human inspection with random screening.
For such systems, there is a natural optimization problem of interest to the attacker. 
Assuming the attacker wants to achieve a certain success rate for a trigger of choice, how can they do so with minimum number of corrupted examples injected into the training set? 

For a given choice of a trigger, the success of an attack is measured by the  Attack Success Rate (ASR), defined as the probability that the corrupted model predicts a target class, $y_{\rm target}$, for an input image from another class with the trigger applied. 
This is referred to as a \textit{test-time poison example}. 
To increase ASR, \textit{train-time poison examples} are injected to the training data. 
A typical recipe is to mimic the test-time poison example by randomly selecting an image from a class other than the target class and applying the trigger function, \(P:\RR^k \to \RR^k\), and label it as the target class, $y_{\rm target} $ \citep{barni2019new,gu2017badnets,liu2020reflection}. 
We refer to this as the ``sampling'' baseline.
In \citep{barni2019new}, for example, the trigger is a periodic image-space signal \(\bm{\Delta} \in \RR^k\) that is added to the image: \(P\del{\bm{x}_{\mathrm{truck}}} = \bm{x}_{\mathrm{truck}} + \bm{\Delta}\).
Example images for this attack are shown in \cref{fig:bd-example} with \(y_{\rm target}=\text{``deer''}\).
The fundamental trade-off of interest is 
between the number of injected poison training examples, \(m\), and ASR as shown in \cref{fig:asr-vs-eps-periodic}.
For the periodic trigger, the sampling baseline requires \(100\) poison examples to reach an ASR of approximately \SI{80}{\percent}.

\begin{center}
  \vspace{-2ex}
  \begin{minipage}{.4\textwidth}
    \begin{figure}[H]
      \centering
      \includegraphics[width=1.0\textwidth]{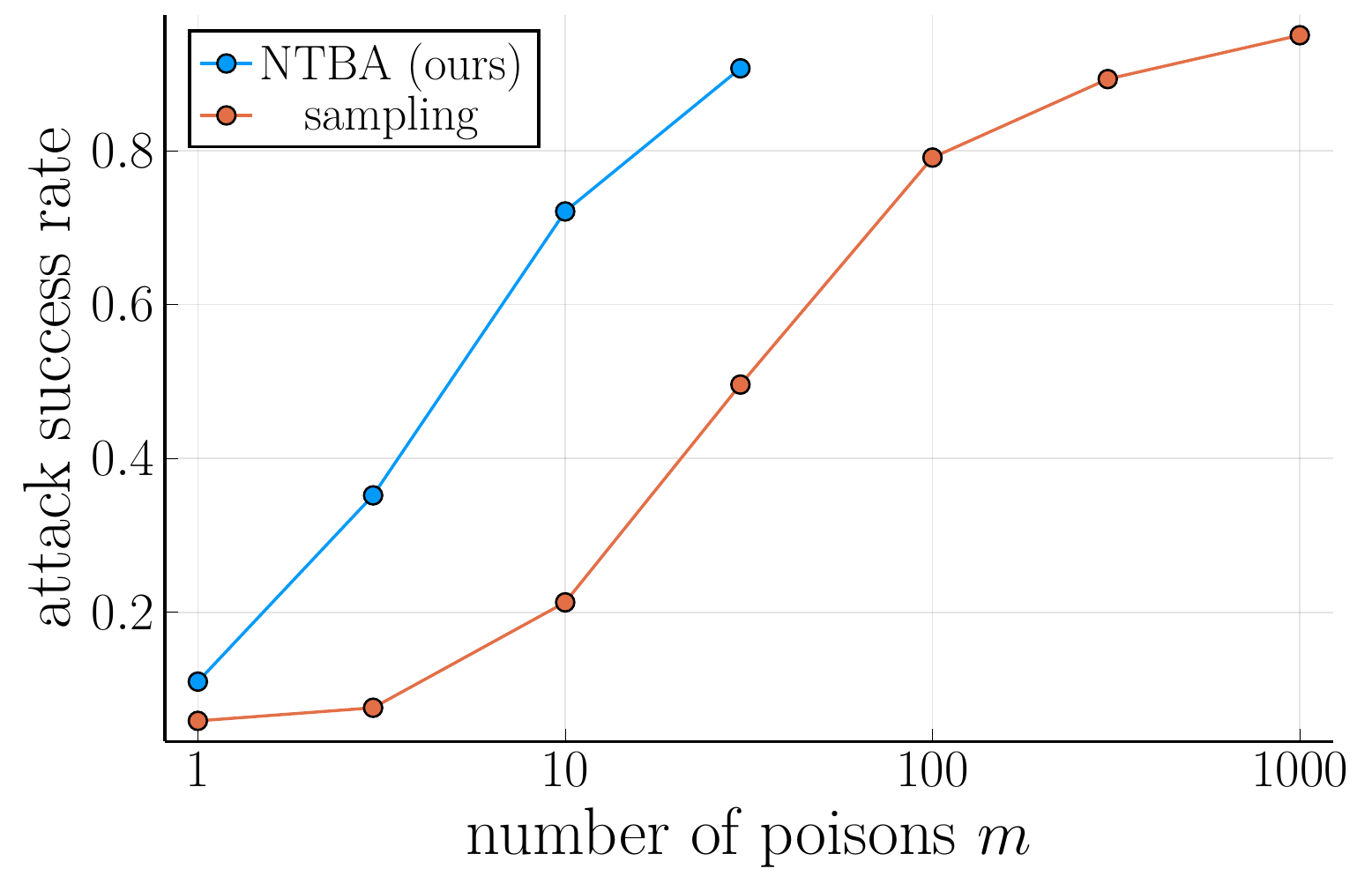}
      \caption{The trade-off between the number of poisons and ASR for the periodic trigger.}\label{fig:asr-vs-eps-periodic}
    \end{figure}
  \end{minipage}\hspace{1em}%
  \begin{minipage}{.55\textwidth}
    \begin{figure}[H]
      \begin{subfigure}{0.33\linewidth}
        \centering
        \includegraphics[width=0.9\textwidth]{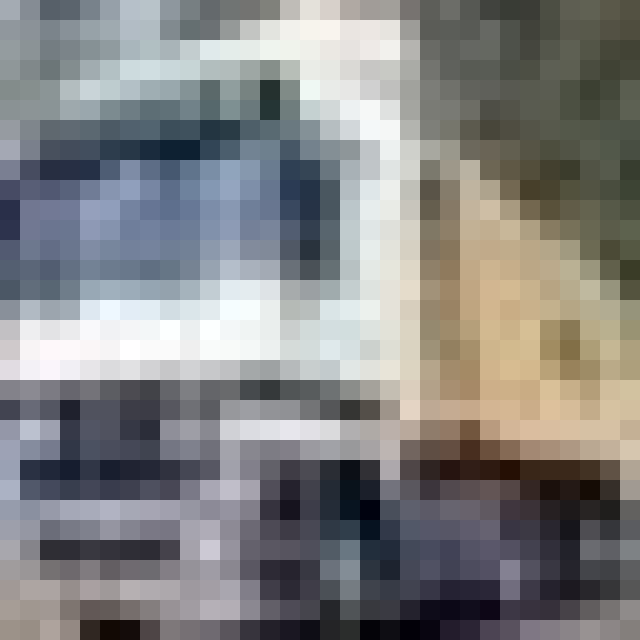}
        label: ``truck''
        \caption{clean}\label{fig:bd-example-truck}
      \end{subfigure}%
      \begin{subfigure}{0.33\linewidth}
        \centering
        \includegraphics[width=0.9\textwidth]{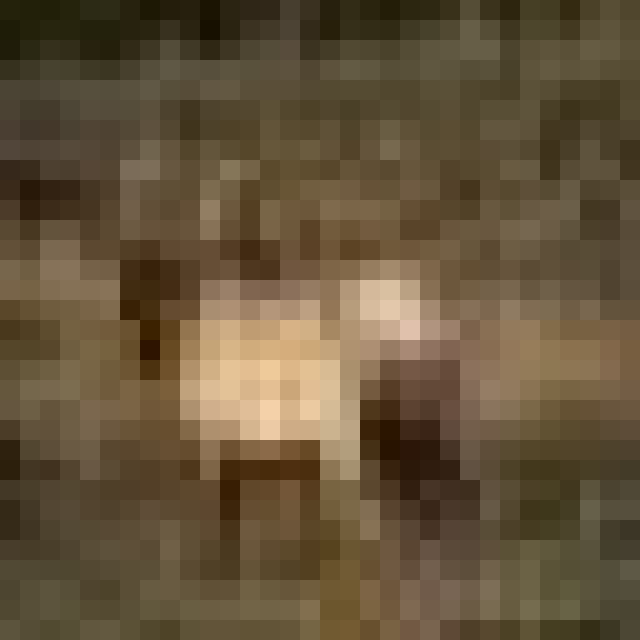}
        label: ``deer''
        \caption{clean}\label{fig:bd-example-deer}
      \end{subfigure}%
      \begin{subfigure}{0.33\linewidth}
        \centering
        \includegraphics[width=0.9\textwidth]{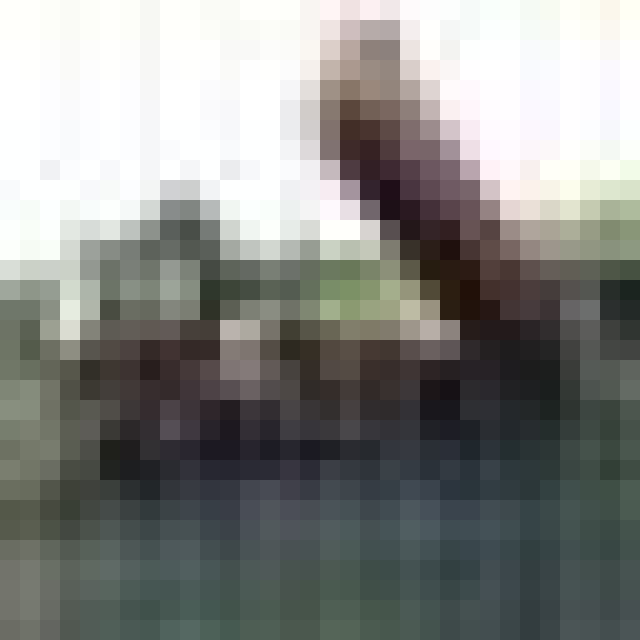}
        label: ``deer''
        \caption{poison}\label{fig:bd-example-poison}
      \end{subfigure}
      \caption{Typical poison attack takes a random sample from the source class (``truck''), adds a trigger ${\bm \Delta}$ to it, and labels it as the target (``deer''). Note the faint vertical striping in \cref{fig:bd-example-poison}.}\label{fig:bd-example}
    \end{figure}
  \end{minipage}
\end{center}

Notice how this baseline, although widely used in robust machine learning  literature, wastes the opportunity to construct stronger attacks. 
We propose to exploit an under-explored attack surface of designing strong attacks and carefully design the train-time poison examples tailored for the choice of the backdoor trigger. 
We want to emphasize that \textit{our goal in proving the existence of such strong backdoor attacks is to motivate continued research into backdoor defenses and inspire practitioners to carefully secure their machine learning pipelines.} 
There is a false sense of safety in systems that ensures a large number of honest data contributors that keep the fraction of corrupted contributions small; we show that it takes only a few examples to succeed in backdoor attacks. 
We survey the related work in \cref{sec:related}. 

\medskip\noindent
{\bf Contributions.} 
We borrow analyses and algorithms from kernel regression to bring a new perspective on  the fundamental trade-off between the attack success rate of a backdoor attack and the number of poison training examples that need to be injected. We
(\textit{i}) use Neural Tangent Kernels (NTKs) to introduce a new computational tool for constructing strong backdoor attacks for training deep neural networks (\cref{sec:ntba,sec:results}); 
(\textit{ii}) use the analysis of the standard  kernel linear regression to interpret  what determines the strengths of a backdoor attack (\cref{sec:intuition}); and (\textit{iii}) investigate the vulnerability of deep neural networks through the lens of corresponding NTKs (\cref{sec:kernel}).

First, we propose a bi-level optimization problem whose solution automatically constructs strong train-time poison examples tailored for the backdoor trigger we want to apply at test-time. 
Central to our approach is the Neural Tangent Kernel (NTK) that models the training dynamics of the neural network. 
Our Neural Tangent Backdoor Attack (NTBA) achieves, for example, an ASR of \SI{72}{\percent} with only 10 poison examples in \cref{fig:asr-vs-eps-periodic}, which is an order of magnitude more efficient. For sub-tasks from CIFAR-10 and ImageNet datasets and two architectures (WideResNet and ConvNeXt), we show the existence of such strong \emph{few-shot} backdoor attacks for two commonly used triggers of the periodic trigger (\cref{sec:results}) and the patch trigger (\cref{apx:patch}).
We show an ablation study showing that every component of NTBA is necessary in discovering such a strong few-shot attack (\cref{sec:ablation}). 
Secondly, we provide interpretation of the poison examples designed with NTBA 
via an analysis of kernel linear regression. In particular, this suggests that small-magnitude train-time triggers lead to strong attacks, when coupled with a clean image that is close in distance, which explains and guides the design of strong attacks. Finally, we investigate the vulnerability 
of deep neural networks to backdoor attacks by comparing the corresponding NTK and the standard Laplace kernel. NTKs allow far away data points to have more influence, compared to the Laplace kernel, which is exploited by few-shot backdoor attacks.

\section{NTBA: Neural Tangent Backdoor Attack}\label{sec:ntba}
We frame the construction of strong backdoor attacks as a bi-level optimization problem and solve it using our proposed Neural Tangent Backdoor Attack (NTBA).
NTBA is composed of the following steps (with details referenced in parentheses):
\begin{enumerate}[label=\textbf{\arabic*.}]
\item \textbf{Model the training dynamics} (\cref{sec:ntba-epoch}): Train the network to convergence on the clean data, saving the network weights and use the \emph{empirical} neural tangent kernel at this choice of weights as our model of the network training dynamics.
\item \textbf{Initialization} (\cref{apx:init}): Use \textit{greedy initialization} to find an initial set of poison images.
\item \textbf{Optimization} (\cref{apx:opt-details,sec:fast-kgrad}): Improve the initial set of poison images using a gradient-based optimizer.
\end{enumerate}
\medskip \noindent
{\bf Background on neural tangent kernels:} 
The NTK of a scalar-valued neural network \(f\) is the kernel associated with the feature map \(\phi\del{\bm x} = \nabla_{\bm{\theta}}f\del{\bm x; \bm{\theta}}\).
The NTK was introduced in \citep{jacot2018neural} which showed that the NTK remains stationary during the training of feed-forward neural networks in the infinite width limit.
When trained with the squared loss, this implies that infinite width neural networks are equivalent to kernel linear regression with the neural tangent kernel.
Since then, the NTK has been extended to other architectures \cite{li2019enhanced,du2019graph,alemohammad2020recurrent,yang2020tensor}, computed in closed form \cite{li2019enhanced,novak2020neural}, and compared to finite neural networks \cite{lee2020finite,arora2019harnessing}.
The closed form predictions of the NTK offer a computational convenience which has been leveraged for data distillation \cite{nguyen2020dataset,nguyen2021dataset}, meta-learning \cite{zhou2021meta}, and subset selection \cite{borsos2020coresets}. 
For finite networks, the kernel is not stationary and its time evolution has been studied in \citep{fort2020deep,long2021properties,seleznova2022analyzing}.
We call the NTK of a finite network with \(\bm{\theta}\) chosen at some point during training the network's \emph{empirical NTK}.
Although the empirical NTK cannot exactly model the full training dynamics of finite networks, \citep{du2018gradient,du2019gradient} give some non-asymptotic guarantees.

\medskip\noindent{\bf Bi-level optimization with NTK:}
Let \(\del{X_{\mathrm{d}}, \bm{y}_{\mathrm{d}}}\) and \(\del{X_{\mathrm{p}}, \bm{y}_{\mathrm{p}}}\) denote the clean and poison training examples, respectively,  \(\del{X_{\mathrm{t}}, \bm{y}_{\mathrm{t}}}\) denote clean test examples, and \(\del{X_{\mathrm{a}}, \bm{y}_{\mathrm{a}}}\) denote test data with the trigger applied and the target label. Our goal is to construct poison examples, \( X_{\mathrm{p}}\), with target label, $\bm{y}_{\mathrm{p}} = y_{\rm target} $, that, when trained on together with clean examples, produce a model which $(i)$ is accurate on clean test data  \(X_{\mathrm{t}}\) and $(ii)$ predicts the target label for poison test data  \(X_{\mathrm{a}}\).
This naturally leads to the   the following  bi-level optimization problem:
\begin{equation}\label{eq:bilevel}
    \min_{X_{\mathrm{p}}}\; \mathcal{L}_{\mathrm{backdoor}}\del*{f\del*{X_{\mathrm{ta}}; \argmin_{\bm{\theta}} \mathcal{L}_{\mathrm{}}\del{f\del{X_{\mathrm{dp}}; \bm{\theta}}, \bm{y}_{\mathrm{dp}}}}, \bm{y}_{\mathrm{ta}}},
\end{equation}
where we denote concatenation with subscripts \(X_{\mathrm{dp}}^\top = \begin{bmatrix}X_{\mathrm{d}}^\top & X_{\mathrm{p}}^\top\end{bmatrix}\) and similarly for \(X_{\mathrm{ta}},y_{\mathrm{ta}} \), and \(y_{\mathrm{dp}} \).
To ensure our objective is differentiable and to permit closed-form kernel predictions, we use the squared loss \(\mathcal{L}\del{\widehat{\bm{y}}, \bm{y}} = \mathcal{L}_{\rm backdoor}\del{\widehat{\bm{y}}, \bm{y}} = \frac{1}{2}\norm[\big]{\widehat{\bm{y}} - \bm{y}}_2^2 \).
Still, such bi-level optimizations are typically challenging to solve \citep{bard1991some,bard2013practical}.
Differentiating directly through the inner optimization \(\argmin_{\bm{\theta}} \mathcal{L}\del{f\del{X_{\mathrm{dp}}; \bm{\theta}}, \bm{y}_{\mathrm{dp}}}\) with respect to the corrupted training data \(X_{\mathrm{p}}\) is impractical for two reasons:
(\emph{i}) backpropagating through an iterative process incurs a significant performance penalty, even when using advanced checkpointing techniques \citep{walther2004advantages} and
(\emph{ii}) the gradients obtained by backpropagating through SGD are too noisy to be useful~\citep{hospedales2020meta}. 
To overcome these challenges,  we propose to use a closed-form kernel to model the training dynamics of the neural network.
This dramatically simplifies and stabilizes our loss, which becomes  
\begin{equation}\label{eq:bd-loss}
  \mathcal{L}_{\mathrm{backdoor}}\del{K_{\mathrm{dp,dpta}}, \bm{y}_{\mathrm{dpta}}} \;\;=\;\; \frac{1}{2}\norm[\big]{\bm{y}_{\mathrm{dp}}^\top K_{\mathrm{dp,dp}}^{-1} K_{\mathrm{dp,ta}} - \bm{y}_{\mathrm{ta}}}_2^2 \;,
\end{equation} 
where we plugged in the closed-form solution of the inner optimization from the {\em kernel linear regression model}, which we can easily differentiate with respect to \(K_{\mathrm{dp,dpta}}\). 
We use \(K : \mathcal{X} \times \mathcal{X} \to \RR\) to denote a kernel function of choice, 
\(K\del{X, X'}\) to denote the \(\abs{X} \times \abs{X'}\) kernel matrix with \(K\del{X, X'}_{i,j} = K\del{X_i, X'_j}\), and 
 subscripts as shorthand for block matrices, e.g. \(K_{\mathrm{a,dp}} = \begin{bmatrix}K\del{X_{\mathrm{a}}, X_{\mathrm{d}}} & K\del{X_{\mathrm{a}}, X_{\mathrm{p}}}\end{bmatrix}\).
This simplification does not come for free, as kernel-designed poisons might not generalize to the neural network training that we desire to backdoor. 
Empirically demonstrating in \cref{sec:results} that there is little loss in transferring our attack to neural network is one of our main goals (see \cref{tab:asr}).

\medskip\noindent
{\bf Greedy initialization.} 
The optimization problem  in \cref{eq:bilevel} is nonconvex.
Empirically, we find that the optimization always converges to a local minima that is close to the initialization of the poison images.
We propose a greedy algorithm to select the initial set of images to start the optimization from.
The algorithm proceeds by applying the trigger function \(P(\cdot)\) to every image in the training set and, incrementally in a greedy fashion, selecting the image that has the greatest reduction in the backdoor loss when added to the poison set.
This is motivated by our analysis in \cref{sec:intuition}, which encourages poisons with small perturbation. 

\subsection{Ablation study}\label{sec:ablation} 



\begin{wraptable}{r}{0.28\textwidth}
  \centering
  \vspace{-0.45cm}
  \caption{Ablation study under the setting of \cref{fig:asr-vs-eps-periodic} with $m=10$.}
  \label{tab:alg-ablation}
  \begin{tabular}[t]{lS[table-format=-2.2,detect-weight,mode=text]}
    \toprule
    ablation & ASR \\
    \midrule
    \textbf{1}\phantom{\textsuperscript{\ensuremath{\prime\prime}}} + \textbf{2} + \textbf{3} & 72.1\%\\
    \textbf{1}\phantom{\textsuperscript{\ensuremath{\prime\prime}}}  \phantom{+}\,+\;\phantom{+}
    \textbf{3} & 12.0\%\\
    \textbf{1}\phantom{\textsuperscript{\ensuremath{\prime\prime}}} + \textbf{2}
    \phantom{+ \textbf{3}} & 16.2\% \\
    \textbf{1}\textsuperscript{\ensuremath{\prime}}\phantom{\textsuperscript{\ensuremath{\prime}}} + \textbf{2} + \textbf{3} & 11.3\% \\
    \textbf{1}\textsuperscript{\ensuremath{\prime\prime}} + \textbf{2} + \textbf{3} & 23.1\%\\
    \bottomrule
  \end{tabular}
\end{wraptable}
We perform an ablation study on the three components at the beginning of this section to demonstrate that they are all necessary.
The alternatives are:
(\textbf{1}\textsuperscript{\ensuremath{\prime}}) the empirical neural tangent kernel but with weights taken from random initialization of the model weights;
(\textbf{1}\textsuperscript{\ensuremath{\prime\prime}}) the infinite-width neural tangent kernel;
(removing \textbf{2}) sampling the initial set of images from a standard Gaussian, (removing \textbf{3}) using the greedy initial poison set without any optimization.
ASR for various combinations  are shown in \cref{tab:alg-ablation}.
The stark difference between our approach (\textbf{1}+\textbf{2}+\textbf{3}) and the rest suggests that all components are important in achieving a strong attack.
Random initialization (\textbf{1}+\textbf{3}) fails as coupled examples that are very close to the clean image space but have different labels is critical in achieving strong attacks as shown in \cref{fig:images}. 
Without our proposed optimization (\textbf{1}+\textbf{2}), the attack is weak.
Attacks designed with different choices of neural tangent kernels (\textbf{1}\textsuperscript{\ensuremath{\prime}}+\textbf{2}+\textbf{3} and \textbf{1}\textsuperscript{\ensuremath{\prime\prime}}+\textbf{2}+\textbf{3}) work well on the kernel models they were designed for, but the attack fails to transfer to the original neural network, suggesting that they are less accurate models of the network training.

\section{Experimental results}\label{sec:results}


We attack a WideResNet-34-5 \cite{zagoruyko2016wide} (\(d \approx 10^7\)) with GELU activations \cite{hendrycks2016gaussian} so that our network will satisfy the smoothness assumption in \cref{sec:fast-kgrad}.
Additionally, we do not use batch normalization which is not yet supported by the neural tangent kernel library we use \cite{novak2020neural}.
Our network is trained with SGD on a 2 label subset of CIFAR-10 \cite{krizhevsky09learning}.
The particular pair of labels is ``truck'' and ``deer'' which was observed in \cite{hayase2021spectre} to be relatively difficult to backdoor since the two classes are easy to distinguish.
We consider two backdoor triggers: the periodic image trigger of \cite{barni2019new} and a \(3 \times 3\) checker patch applied at a random position in the image.
These two triggers represent sparse control over images at test time in frequency and image space respectively.
Results for the periodic trigger are given here while results for the patch trigger are given in \cref{apx:patch}.

To fairly evaluate performance, we split the CIFAR-10 training set into an inner training set and validation set containing \SI{80}{\percent} and \SI{20}{\percent} of the images respectively.
We run NTBA with the inner training set as \(D_{\mathrm{d}}\), the inner validation set as \(D_{\mathrm{t}}\), and the inner validation set with the trigger applied as \(D_{\mathrm{a}}\).
Our neural network is then trained on \(D_{\mathrm{d}} \cup D_{\mathrm{p}}\) and tested on the CIFAR-10 test set.

We also attack a pretrained ConvNeXt \cite{liu2022convnet} finetuned on a 2 label  subset of ImageNet, following the setup of \cite{saha2020hidden} with details given in \cref{apx:imagenet}. 
We describe the computational resources used to perform our attack in \cref{apx:computational-resources}.

\subsection{NTBA makes backdoor attacks significantly more efficient}
\label{sec:results:main}

Our main  results show that (\textit{i}) as expected, there are some gaps in ASR when applying NTK-designed poison examples to neural network training, but
(\textit{ii}) NTK-designed poison examples still manage to be significantly stronger compared to sampling baseline. 
The most relevant metric is the test results of neural network training evaluated on the original validation set with the trigger applied, \(\mathrm{asr}_{\mathrm{nn,te}}\). 
In \cref{tab:asr}, to achieve \(\mathrm{asr}_{\mathrm{nn,te}}=90.7\%\), NTBA requires 30 poisons, which is an order of magnitude fewer than the sampling baseline.
The ASR for backdooring kernel regressions is almost perfect, as it is what NTBA is designed to do; we consistently get high \(\mathrm{asr}_{\mathrm{ntk,te}}\) with only a few poisons. 
Perhaps surprisingly, we show that these NTBA-designed attacks can be used as is to attack regular neural network training and achieve ASR significantly higher than the commonly used baseline in \cref{tab:asr}, \cref{fig:asr-vs-eps-periodic,fig:asr-vs-eps-patch,fig:asr-vs-eps-imagenet,fig:asr-vs-eps-imagenet-pretrained} for WideResNet trained on CIFAR-10 subtasks and ConvNeXt trained on ImageNet subtasks, NTBA tailored for patch and periodic triggers, respectively.
ASR results are percentages and we omit $\%$ in this section.


\begin{table}[h]
    \centering
    \caption{ASR results for NTK and NN (\(\mathrm{asr}_{\blank,\mathrm{ntk}}\) and \(\mathrm{asr}_{\blank,\mathrm{nn}}\)) at train and test time (\(\mathrm{asr}_{\mathrm{tr},\blank}\) and \(\mathrm{asr}_{\mathrm{te},\blank}\)).
    The NTBA attack transferred to neural networks is significantly stronger than the sampling based attack using the same periodic trigger across a range of poison budgets \(m\).
      A graph version of this table is in \cref{fig:asr-vs-eps-periodic}.}\label{tab:asr}
    \vspace{1.5ex}
    \begin{tabular}[b]{r*{5}{S[table-format=-2.2,detect-weight,mode=text]S[table-format=-2.2,detect-weight,mode=text]}}
        \toprule
        & \multicolumn{4}{c}{ours} & sampling\\
        \(m\) & \(\mathrm{asr}_{\mathrm{ntk,tr}}\) & \(\mathrm{asr}_{\mathrm{ntk,te}}\) & \(\mathrm{asr}_{\mathrm{nn,tr}}\) & \(\mathrm{asr}_{\mathrm{nn,te}}\) & \(\mathrm{asr}_{\mathrm{nn,te}}\)\\
        \midrule
        1 & 100.0 & 85.2 & 0.2 & 11.0 & 5.9 \\ 
        3 & 100.0 & 92.8 & 5.6 & 35.2 & 7.6 \\
        10 & 100.0 & 95.2 & 65.2 & 72.1 & 21.3 \\ 
        30 & 100.0 & 96.4 & 94.2 & 90.7 & 49.6 \\
        \bottomrule
    \end{tabular}\hspace{1em}%
    \begin{tabular}[b]{r*{1}{S[table-format=-2.2,detect-weight,mode=text]S[table-format=-2.2,detect-weight,mode=text]}}
        \toprule
        & sampling\\
        \(m\) & \(\mathrm{asr}_{\mathrm{nn,te}}\)\\
        \midrule
         0  & 5.5\\
        100 &79.1\\
        300 &  89.3\\
        1000  & 95.0\\
        \bottomrule
    \end{tabular}
\end{table}

    
\subsection{The attacker does not need to know all the training data}
\label{sec:results:reduce}

\begin{wrapfigure}{r}{0.5\textwidth}
  \centering
  \vspace{-0.45cm}
    \captionof{table}{ASR decreases gracefully with the attacker knowing only $\beta$ fraction of the data.}
    \begin{tabular}[t]{l|S S S S [table-format=-2.2,detect-weight,mode=text]}
    \toprule
         $\beta$ & 1.0  & 0.75 & 0.5 & 0.25\\
         \midrule
           \(\mathrm{asr}_{\mathrm{nn,te}}\) & 96.3 & 94.7 & 78.5 & 73.4\\
         \bottomrule
    \end{tabular}
    \label{tab:data-ablation}
\end{wrapfigure}
In our preceding experiments, the attacker has knowledge of the entire training set and a substantial quantity of validation data.
In these experiments, the attacker is given a \(\beta\) fraction of the 2-label CIFAR-10 subset's train and validation sets.
The backdoor is computed using only this partial data and the neural network is then run on the full data.
NTBA degrades gracefully as the amount of information available to the attacker is reduced. Results for \(m = 10\) are shown in \cref{tab:data-ablation}.


\subsection{Is neural tangent kernel special?}\label{sec:results:kernel}

\begin{wraptable}{r}{0.37\textwidth}
  \centering
  \vspace{-0.45cm}
  \caption{results for directly attacking the NTK and Laplace kernels on CIFAR-10 with a periodic trigger. \(\mathrm{acc}_{\mathrm{tr}}\) refers to clean accuracy after training on corrupted data.}\label{tab:ntk-vs-laplace-cifar}
  \vspace{1.5ex}
  \begin{tabular}[t]{rlll}
    \toprule
    \(m\) & kernel & \(\mathrm{acc}_{\mathrm{tr}}\) & \(\mathrm{asr}_{\mathrm{tr}}\)\\
    \midrule
    1 & NTK & \SI{93}{\percent} &  \SI{100}{\percent}\\
    10 & Laplace & \SI{93}{\percent} & \SI{11}{\percent}\\
    \bottomrule
  \end{tabular}
\end{wraptable} 
Given the extreme vulnerability of NTKs (e.g., \(\mathrm{asr}_{\mathrm{ntk,te}}=85.2\) with one poison in \cref{tab:asr}), it is natural to ask if other kernel models can be similarly backdoored. 
To test this hypothesis, we apply the optimization from NTBA to both NTK and the standard Laplace kernel on the CIFAR-10 sub-task, starting from a random initialization. 
Although the Laplace kernel is given ten times  more poison points, the optimization of NTBA can only achieve \SI{11}{\percent} ASR, even on the training data. 
In contrast, NTBA with the NTK yields a \SI{100}{\percent} train-ASR, with the clean accuracy for both kernels remaining the same. This suggests that Laplace kernel is not able to learn the poison without sacrificing the accuracy on clean data points. 
 In \cref{sec:kernel},  we further investigate what makes NTK (and hence neural networks) special. 







\section{Interpreting the NTBA-designed poison examples}\label{sec:intuition} 
\label{sec:backdoor-klr}

\begin{figure}[h]
  \centering
  \adjustbox{clap}{\hspace{2.5em}\begin{subfigure}[t]{0.1\textwidth}
    \centering
    \includegraphics[width=0.9\textwidth]{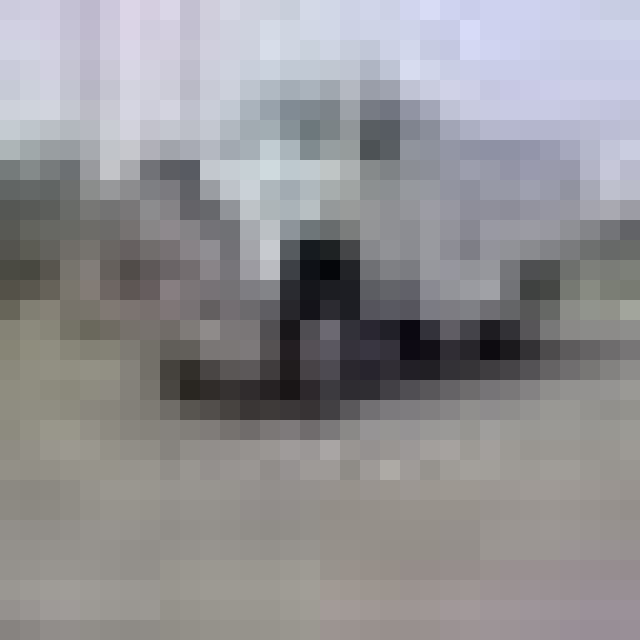}\\[0.0em]
    \includegraphics[width=0.9\textwidth]{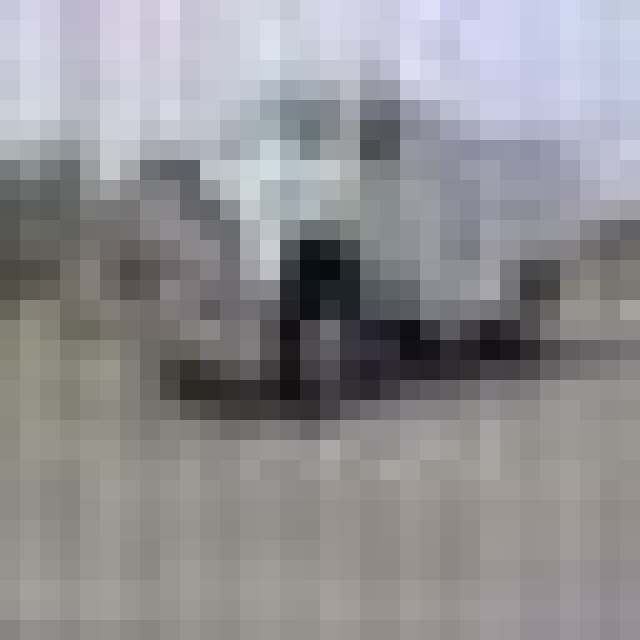}\\[0.0em]
    \includegraphics[width=0.9\textwidth]{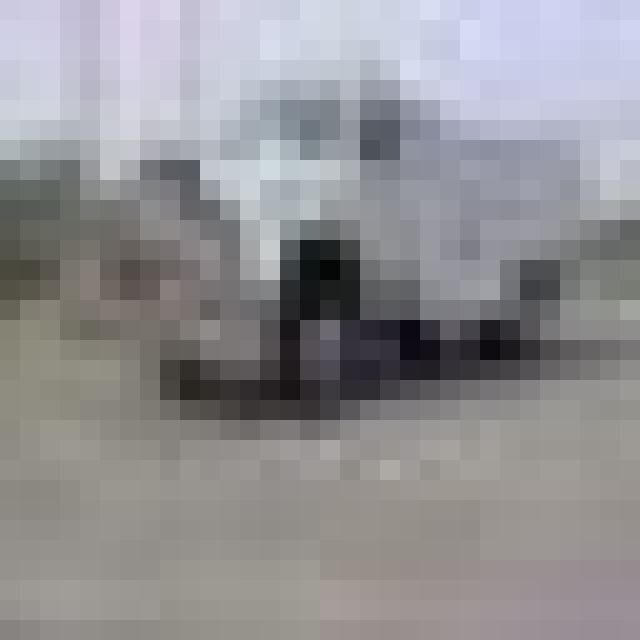}
    \caption{\(m=1\)} \label{fig:6a}
  \end{subfigure}\hspace{0.06in}%
  \put(-56,10){\footnotesize\rotatebox{90}{clean}}
  \put(-56,-34){\footnotesize\rotatebox{90}{greedy}}
  \put(-64,-82){\footnotesize\rotatebox{90}{greedy+}}
  \put(-56,-82){\footnotesize\rotatebox{90}{optimize}}
  \begin{subfigure}[t]{0.2\textwidth}
    \centering
    \includegraphics[width=0.45\textwidth]{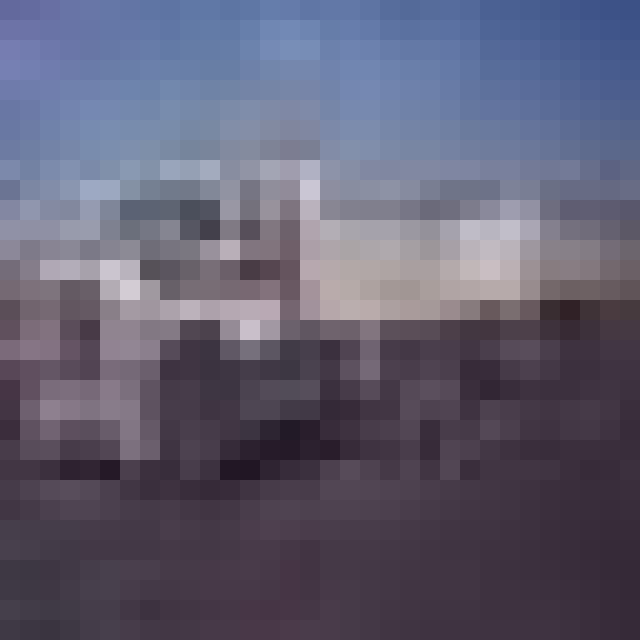}\hspace{0.1em}%
    \includegraphics[width=0.45\textwidth]{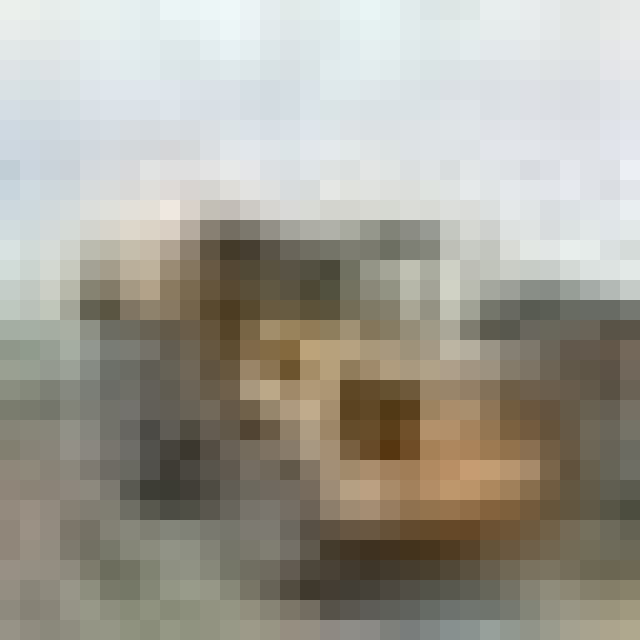}\\[0.0em]
    \includegraphics[width=0.45\textwidth]{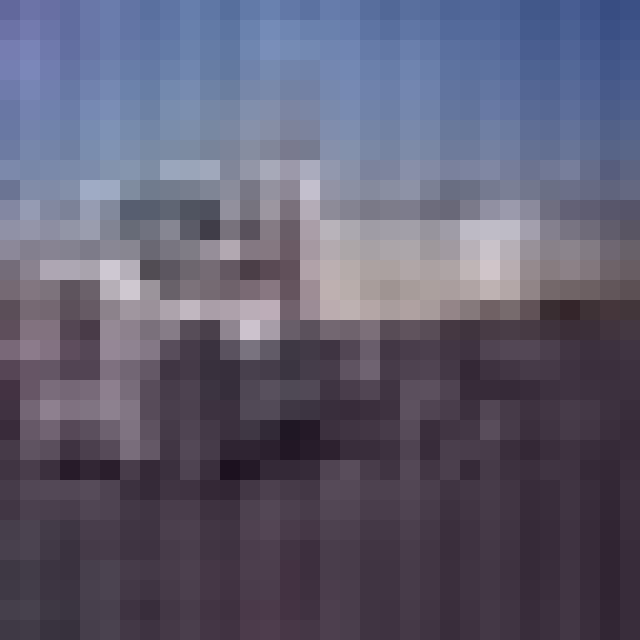}\hspace{0.1em}%
    \includegraphics[width=0.45\textwidth]{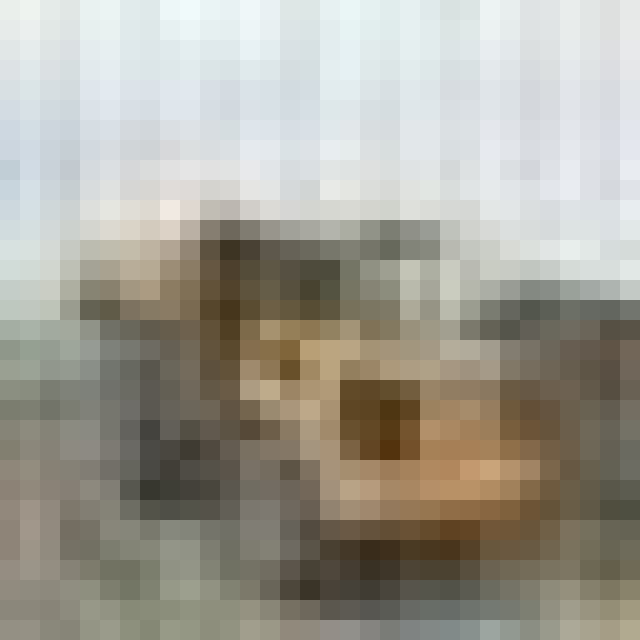}\\[0.0em]
    \includegraphics[width=0.45\textwidth]{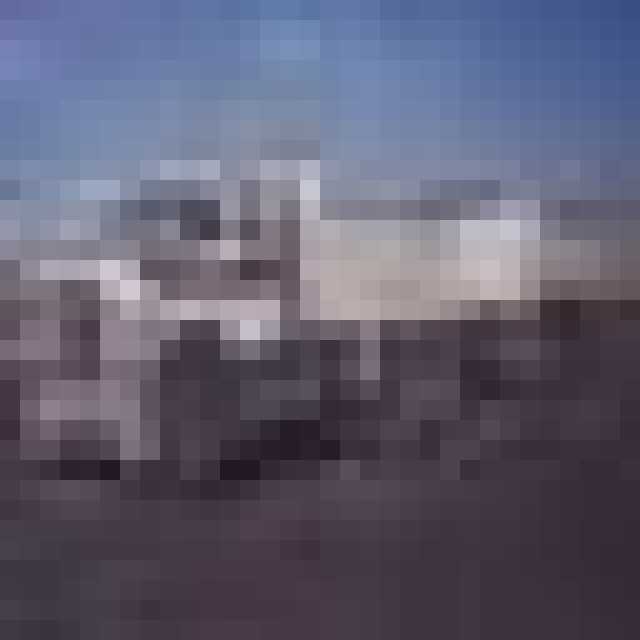}\hspace{0.1em}%
    \includegraphics[width=0.45\textwidth]{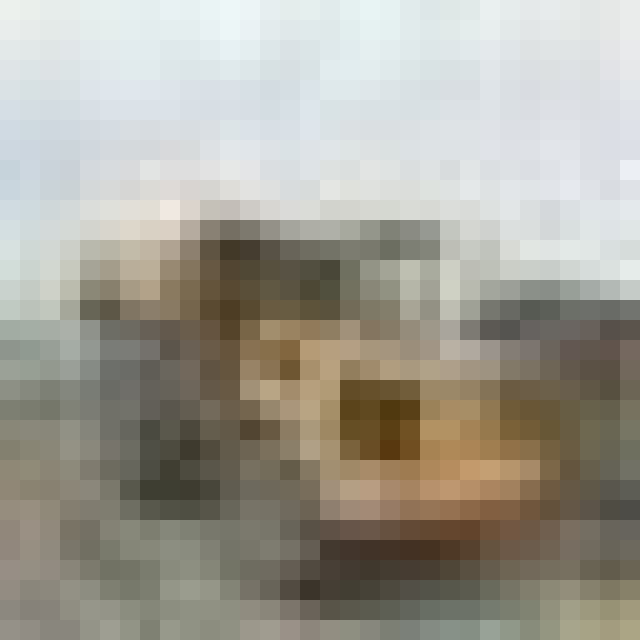}
    \caption{\(m=3\)}
   \end{subfigure}\hspace{-0.06in}%
   \begin{subfigure}[t]{0.7\textwidth}
    \centering
    \includegraphics[width=0.12857\textwidth]{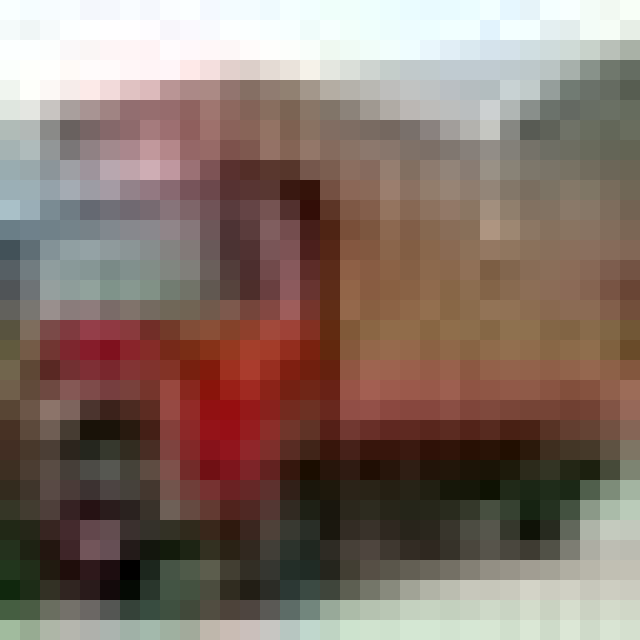}\hspace{0.1em}%
    \includegraphics[width=0.12857\textwidth]{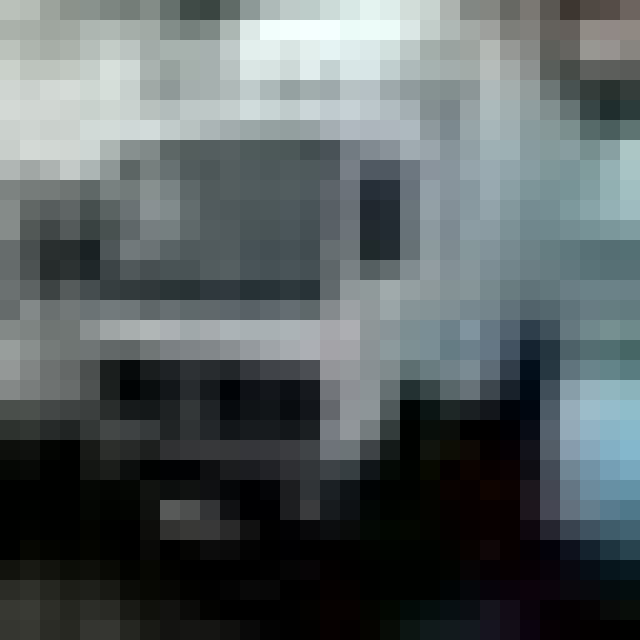}\hspace{0.1em}%
    \includegraphics[width=0.12857\textwidth]{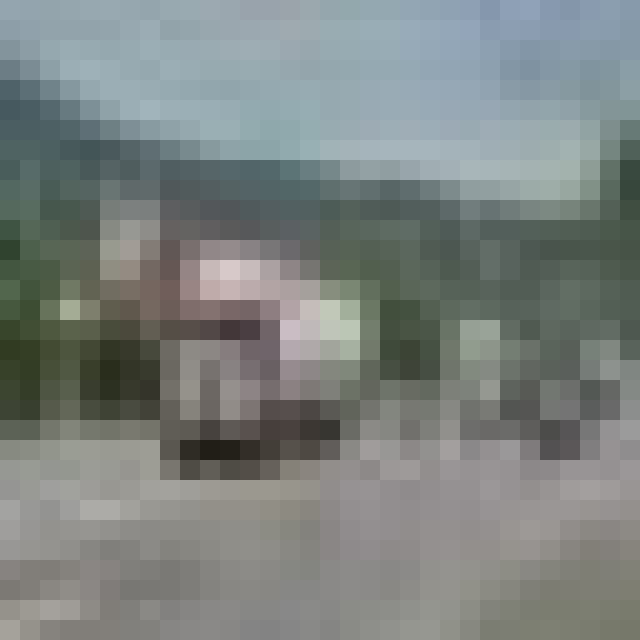}\hspace{0.1em}%
    \includegraphics[width=0.12857\textwidth]{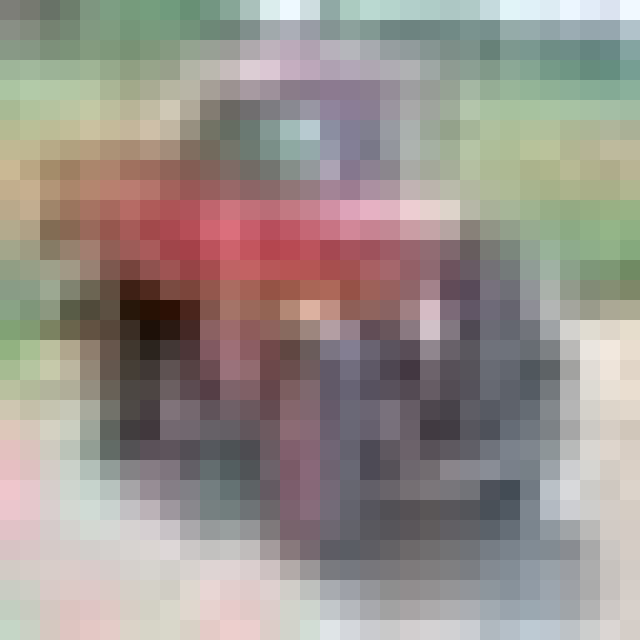}\hspace{0.1em}%
    \includegraphics[width=0.12857\textwidth]{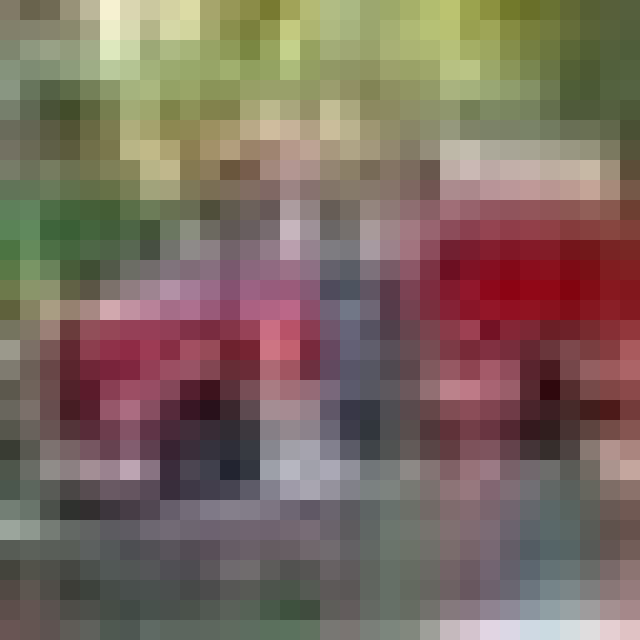}\hspace{0.1em}%
    \includegraphics[width=0.12857\textwidth]{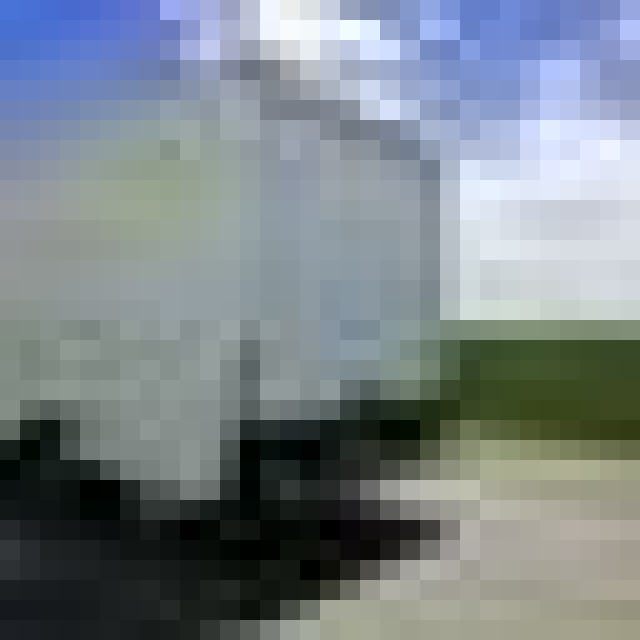}\hspace{0.1em}%
    \includegraphics[width=0.12857\textwidth]{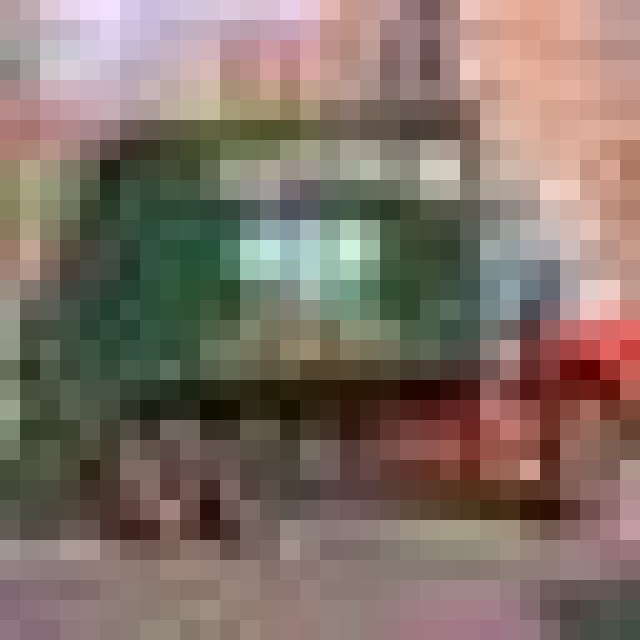}\\[0.0em]
    \includegraphics[width=0.12857\textwidth]{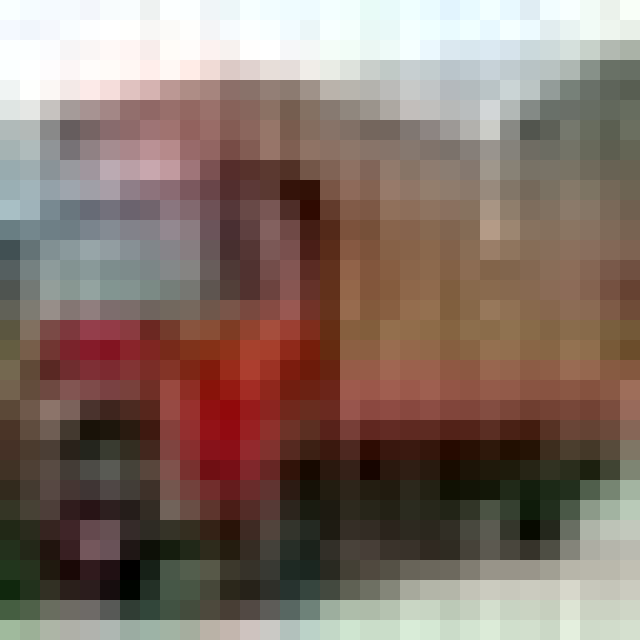}\hspace{0.1em}%
    \includegraphics[width=0.12857\textwidth]{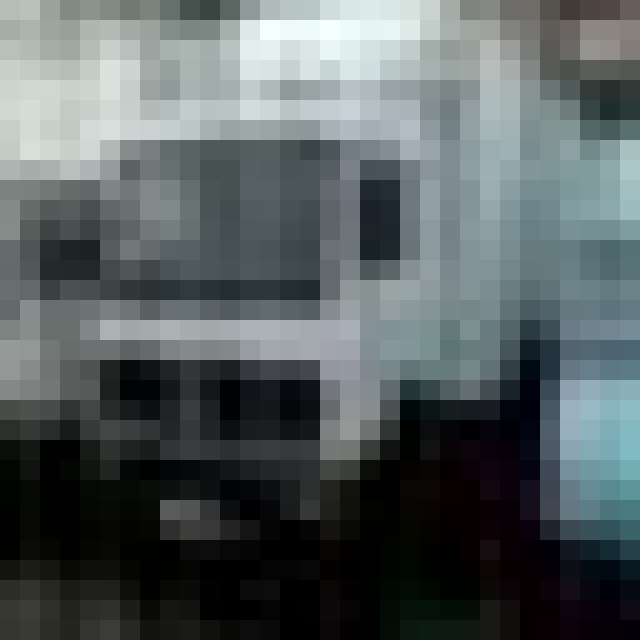}\hspace{0.1em}%
    \includegraphics[width=0.12857\textwidth]{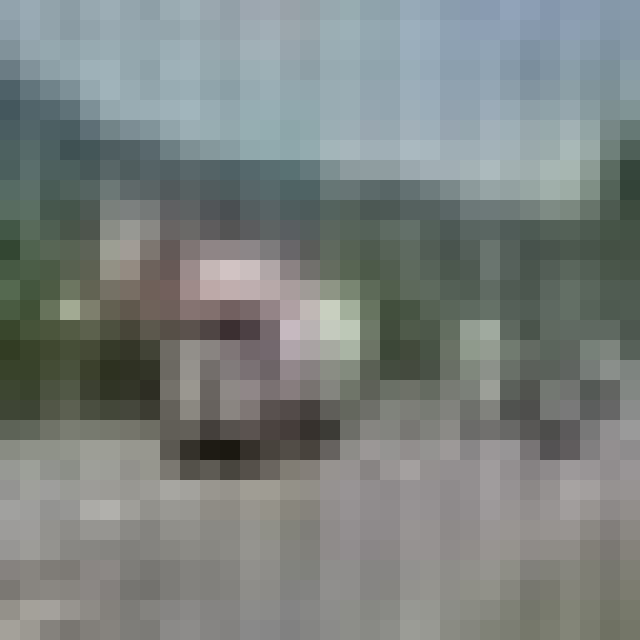}\hspace{0.1em}%
    \includegraphics[width=0.12857\textwidth]{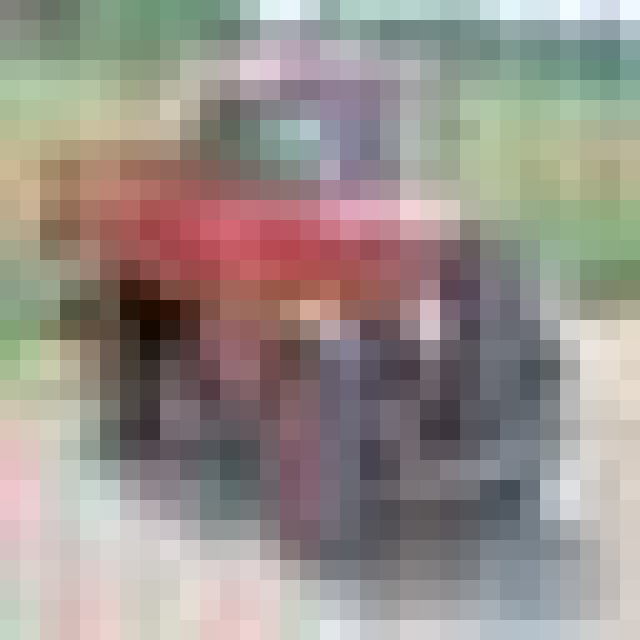}\hspace{0.1em}%
    \includegraphics[width=0.12857\textwidth]{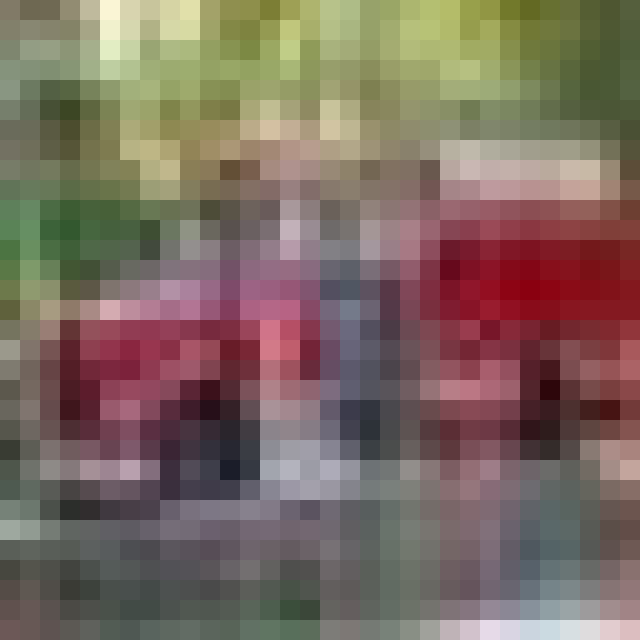}\hspace{0.1em}%
    \includegraphics[width=0.12857\textwidth]{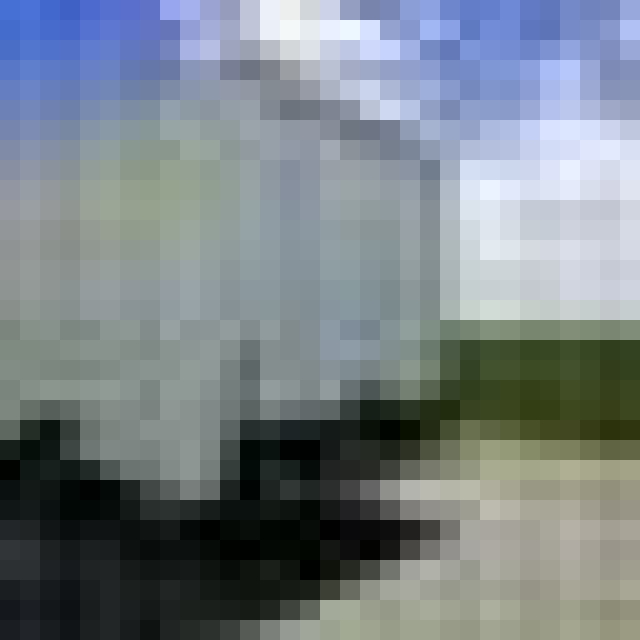}\hspace{0.1em}%
    \includegraphics[width=0.12857\textwidth]{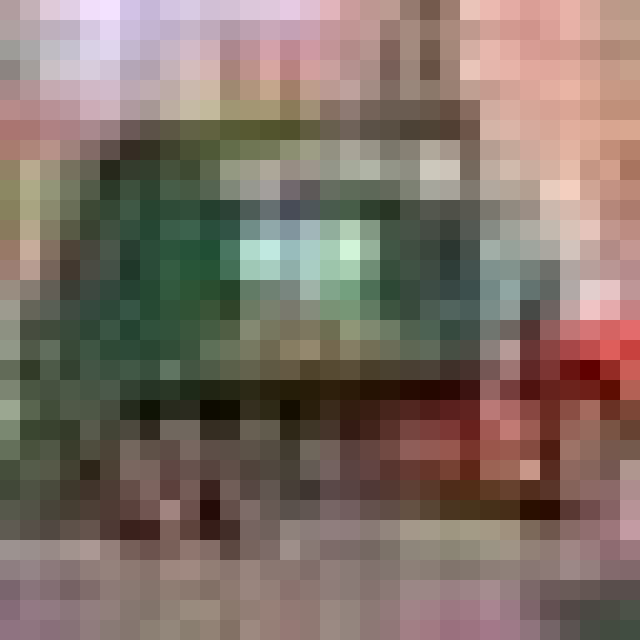}\\[0.0em]
    \includegraphics[width=0.12857\textwidth]{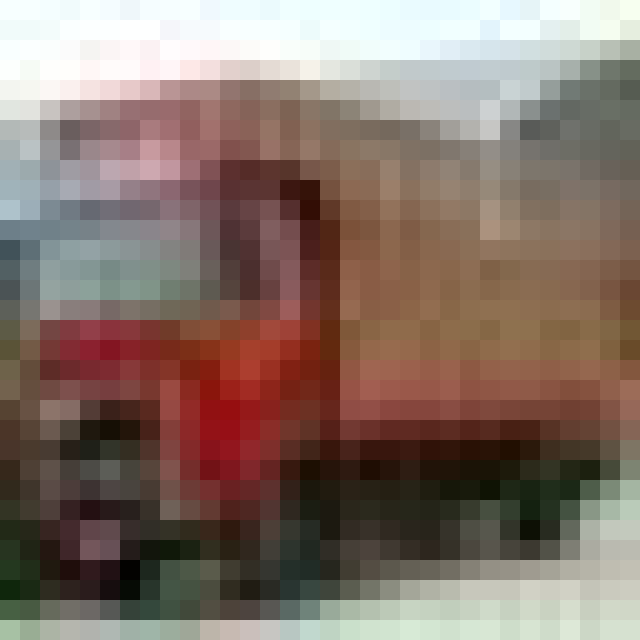}\hspace{0.1em}%
    \includegraphics[width=0.12857\textwidth]{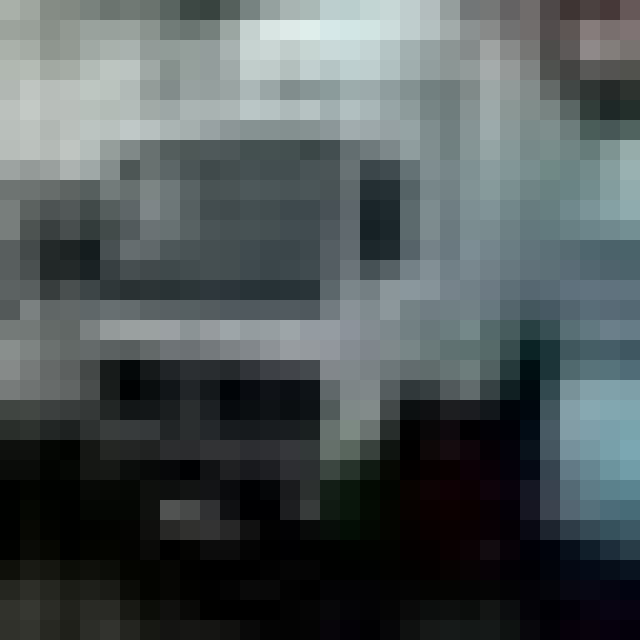}\hspace{0.1em}%
    \includegraphics[width=0.12857\textwidth]{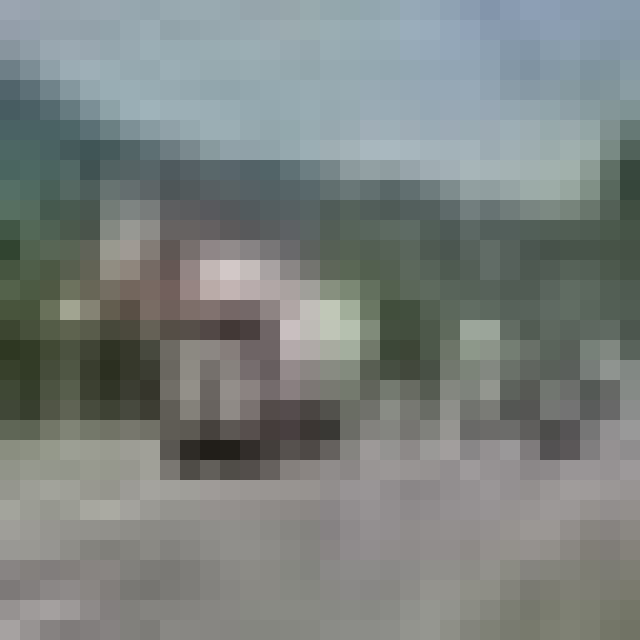}\hspace{0.1em}%
    \includegraphics[width=0.12857\textwidth]{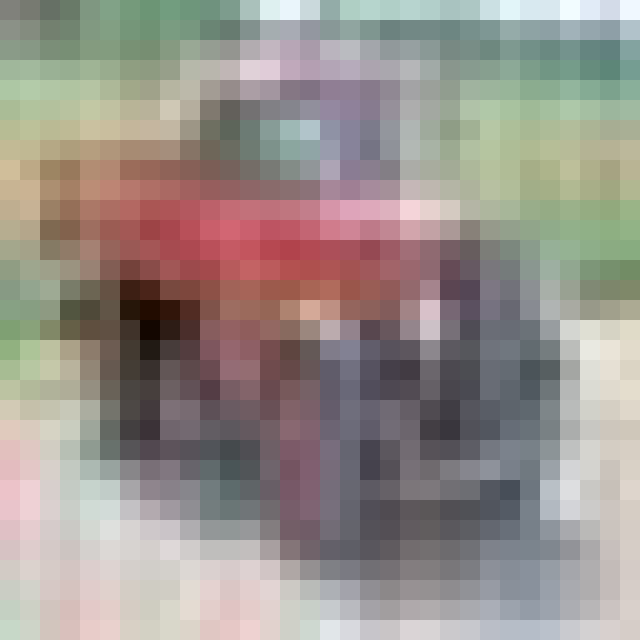}\hspace{0.1em}%
    \includegraphics[width=0.12857\textwidth]{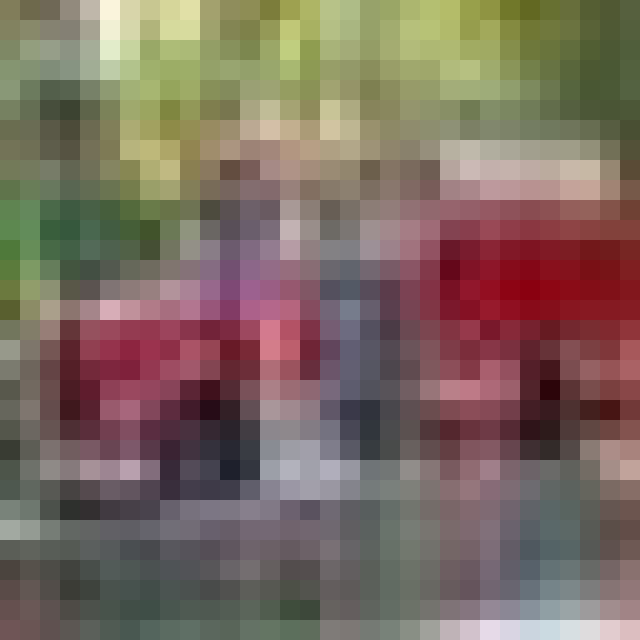}\hspace{0.1em}%
    \includegraphics[width=0.12857\textwidth]{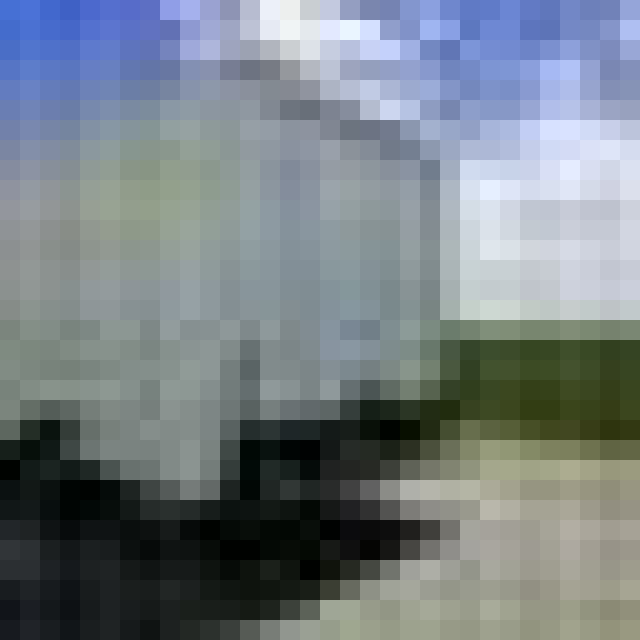}\hspace{0.1em}%
    \includegraphics[width=0.12857\textwidth]{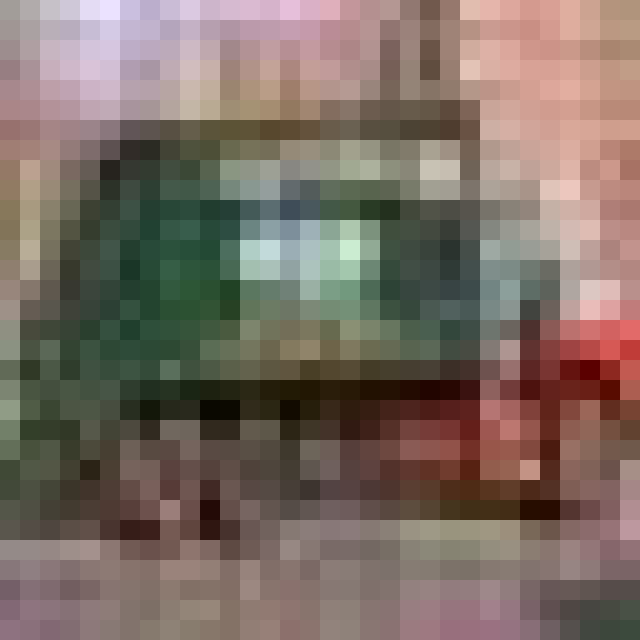}
    \caption{\(m=10\)}
  \end{subfigure}}
  \caption{
  Images produced by NTBA for period trigger and \(m \in \set{1, 3, 10}\). 
  The top row shows the original clean image of the greedy initialization, the middle row shows the greedy initialization that includes the trigger, and the bottom row shows the final poison image after optimization. Duplicate images, for example the first poison image for $m=3$, have been omitted to save space.}\label{fig:images}
\end{figure}

We show the images produced by NTBA in \cref{fig:images}.
Comparing second and third rows of \cref{fig:images}, observe that the optimization generally reduces the magnitude of the trigger. Precise measurements in \cref{fig:trigger-heatmaps,fig:trigger-norms} further show that the magnitude of the train-time trigger learned via NTBA gets smaller as we decrease the number of injected poison examples \(m\). 

We analyze kernel linear regression to show that \emph{backdoor attacks increase in strength as the poison images get closer to the manifold of clean data.}
 This provides an interpretation of the NTBA-designed poison examples.
Given training data \(D_{\mathrm{d}} = \del{X_{\mathrm{d}} \in \RR^{n\times k},  \bm{y}_{\mathrm{d}} \in \set{\pm 1}^n}\) and a generic kernel $K$, the prediction of a kernel linear regression model trained on \(D_{\mathrm{d}}\) and tested on some \(\bm{x} \in \RR^k\) is
\begin{equation}
  f\del{\bm{x}; D_{\mathrm{d}}}\;\; \triangleq \;\; \bm{y}^\top_{\mathrm{d}} K(X_{\mathrm{d}}, X_{\mathrm{d}})^{-1}K\del{X_{\mathrm{d}}, \bm{x}},\label{eq:klr}
\end{equation}
where \(K\del{\blank, \blank}\) denotes the kernel matrix over the data.
For simplicity, suppose we are adding a single poison example \(D_{\mathrm{p}} = \set{\del{\bm{x}_{\mathrm{p}}, y_{\mathrm{p}}}}\) and testing on a single point \(\bm{x}_{\mathrm{a}}\). 
For the attack to succeed, the injected poison example needs to change the prediction of \(\bm{x}_{\mathrm{a}}\) by ensuring that
\begin{equation}
  \underbrace{f\del{\bm{x}_{\mathrm{a}}; D_{\mathsf{d}} \cup \set{\del{\bm{x}_{\mathrm{p}}, y_{\mathrm{p}}}}}}_{\text{poisoned model prediction}}\;\; - \underbrace{f\del{\bm{x}_{\mathrm{a}}; D_{\mathsf{d}}}}_{\text{clean model prediction}} = \frac{\phi\del{\bm{x}_{\mathrm{p}}}\del{I - P}\phi\del{\bm{x}_{\mathrm{a}}}^\top}{\phi\del{\bm{x}_{\mathrm{p}}}\del{I - P}\phi\del{\bm{x}_{\mathrm{p}}}^\top}\del{y_{\mathrm{p}} - f\del{\bm{x}_{\mathrm{p}}; D_{\mathsf{d}}}}\label{eq:lin-add-1-point}
\end{equation}
is sufficiently large,
where \(\phi : \mathcal{X} \to \RR^d\) is a feature map of kernel \(K\) such that \(K\del{\bm{x}, \bm{y}} = \inner{\phi\del{\bm{x}},\phi\del{\bm{y}}}\), and \(P = \Phi^\top\del{\Phi\Phi^\top}^{-1}\Phi\) is the hat matrix of \(\Phi\) (i.e. \(P\) projects onto the span of the rows of \(\Phi\)) where \(\Phi\) is the matrix with rows \(\phi\del{\bm x}\) for \(\bm x \in X_{\mathrm{d}}\).
\Cref{eq:lin-add-1-point} follows from the Schur complement after 
adding one row and column to the kernel matrix \(K(X_{\mathrm{d}} , X_{\mathrm{d}})\)  and adding one dimension to each of \(\bm{y}_{\mathrm{d}}\) and \(K\del{X_{\mathrm{d}}, \bm{x}}\) in \cref{eq:klr}. 
We assume that both \(\bm{x}_{\mathrm{p}}\) and \(\bm{x}_{\mathrm{a}}\) are small perturbations of clean data points, and let 
  \(\bm{\Delta}_{\mathrm{p}} \triangleq  \widetilde{\bm{x}}_{\mathrm{p}} - \bm{x}_{\mathrm{p}} \) and \(\bm{\Delta}_{\mathrm{a}} \triangleq  \widetilde{\bm{x}}_{\mathrm{a}} - \bm{x}_{\mathrm{a}} \) 
  respectively denote the train-time perturbation and the test-time trigger
  for some clean data points  \(\widetilde{\bm{x}}_{\mathrm{p}}, \widetilde{\bm{x}}_{\mathrm{a}} \in X_{\mathrm{d}}\). 
  In the naive periodic attack, both 
  \(\bm{\Delta}_{\mathrm{p}}\) and \(\bm{\Delta}_{\mathrm{a}}\) are the periodic patterns we add.  
  Our goal is to find out which choice of the train-time perturbation, \(\bm{\Delta}_{\mathrm{p}}\), would make the attack stronger (for the given test-time trigger  \(\bm{\Delta}_{\mathrm{a}}\)). 
  
  The powerful poison examples discovered via the proposed NTBA show the following patterns. In \cref{fig:trigger-heatmaps}, each pixel shows the norm of the three channels of the perturbation \(\bm{\Delta}_{\mathrm{p}}\) for a single poison example with the same closest  clean image; the corresponding train examples are explicitly shown in \cref{fig:6a}.
The range of the pixel norm 0.2 is after data standardization normalized by the standard deviation for that pixel. 
In \crefrange{fig:trigger-1}{fig:trigger-30}, we see that the \(\bm{\Delta}_{\mathrm{p}}\) aligns with the test-time trigger \(\bm{\Delta}_{\mathrm{a}}\) in \cref{fig:trigger-test}, but with reduced amplitude and some fluctuations. When the allowed number of poisoned examples, $m$, is small, NTBA makes each poison example more powerful by reducing the magnitude of the perturbation $\bm{\Delta}_{\mathrm{p}}$. 
  In \cref{fig:trigger-norms}, the perturbations grow larger as we increase the number of poisoned examples constructed with our proposed attack NTBA.
NTBA uses smaller training-time perturbations to achieve stronger attacks when the number of poison examples is small which is consistent with the following analysis based on the first-order approximation in \cref{eq:intuition}.

\begin{center}
  \vspace{-2ex}
  \begin{minipage}{.55\textwidth}
    \begin{figure}[H]
      \centering
      \begin{subfigure}{0.33\textwidth}
        \centering
        \includegraphics[width=0.7\textwidth]{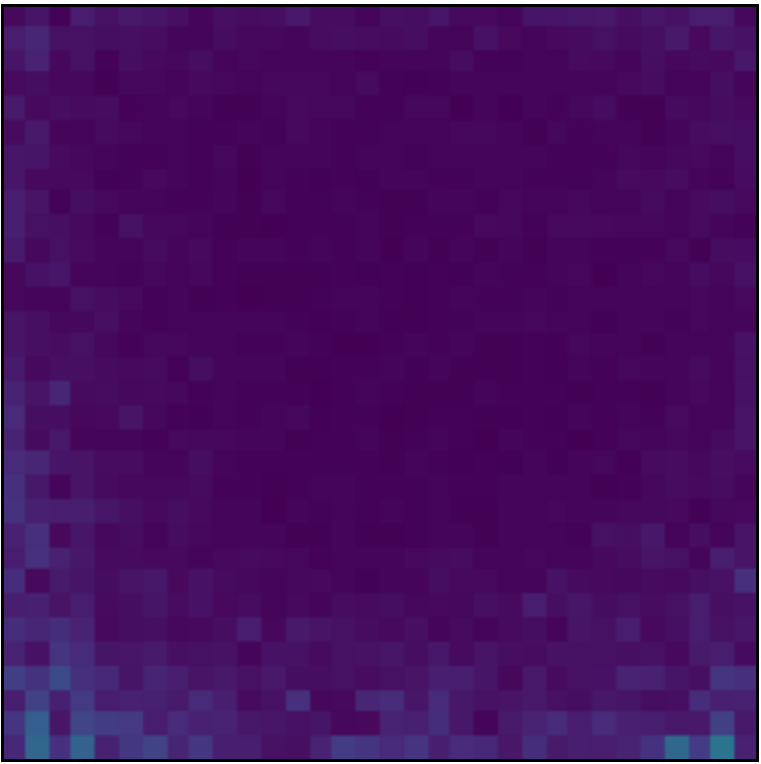}
        \caption{\(m=1\)}\label{fig:trigger-1}
      \end{subfigure}%
      \begin{subfigure}{0.33\textwidth}
        \centering
        \includegraphics[width=0.7\textwidth]{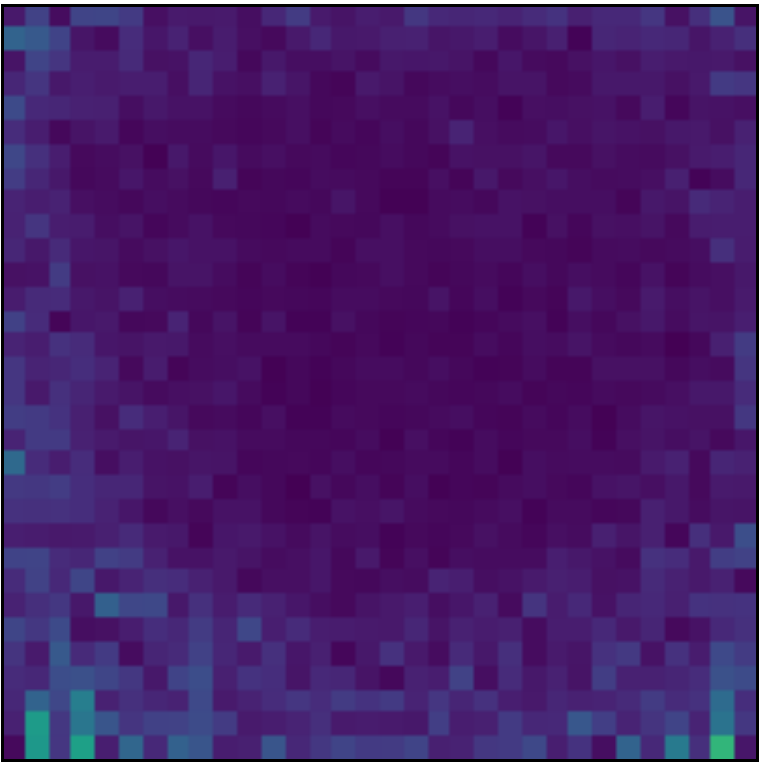}
        \caption{\(m=3\)}\label{fig:trigger-3}
      \end{subfigure}%
      \begin{subfigure}{0.33\textwidth}
        \centering
        \includegraphics[width=0.7\textwidth]{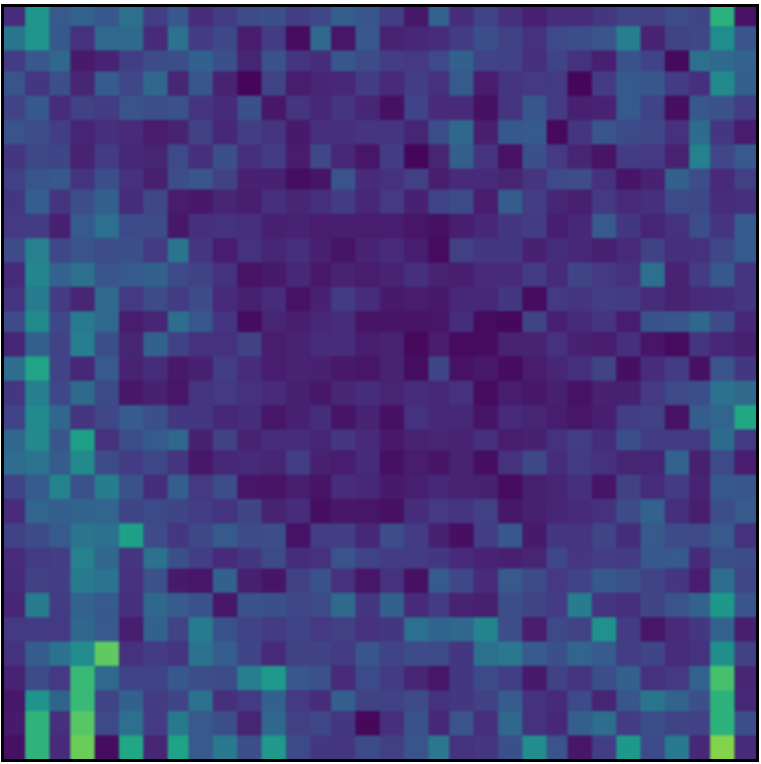}
        \caption{\(m=10\)}\label{fig:trigger-10}
      \end{subfigure}\vspace{0.5em}
      \begin{subfigure}[t]{0.33\textwidth}
        \centering
        \includegraphics[width=0.7\textwidth]{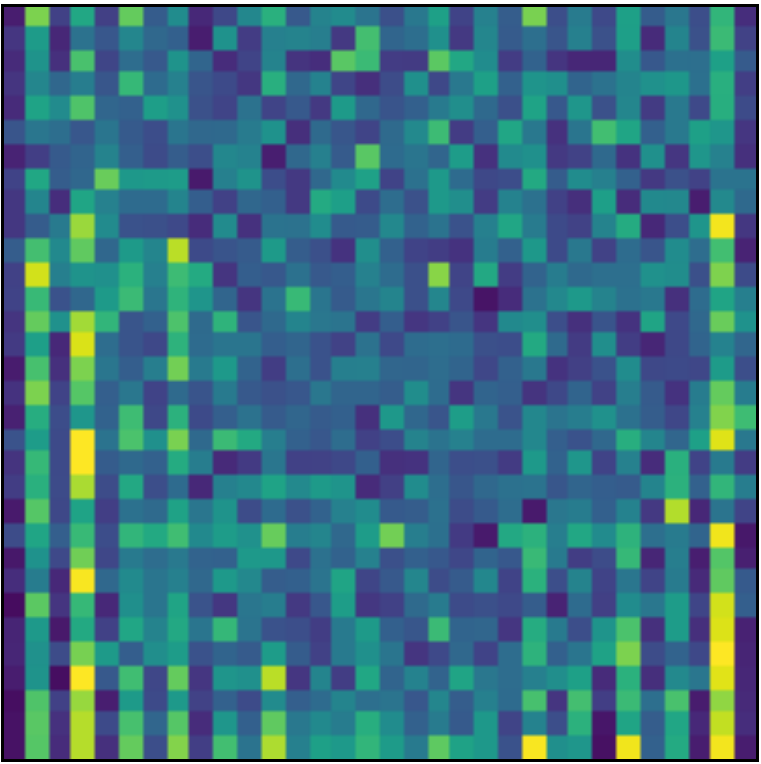}
        \caption{\(m=30\)}\label{fig:trigger-30}
      \end{subfigure}%
      \begin{subfigure}[t]{0.43\textwidth}
        \centering
        \adjustbox{raise=-0.7mm}{
          \includegraphics[width=0.7\textwidth]{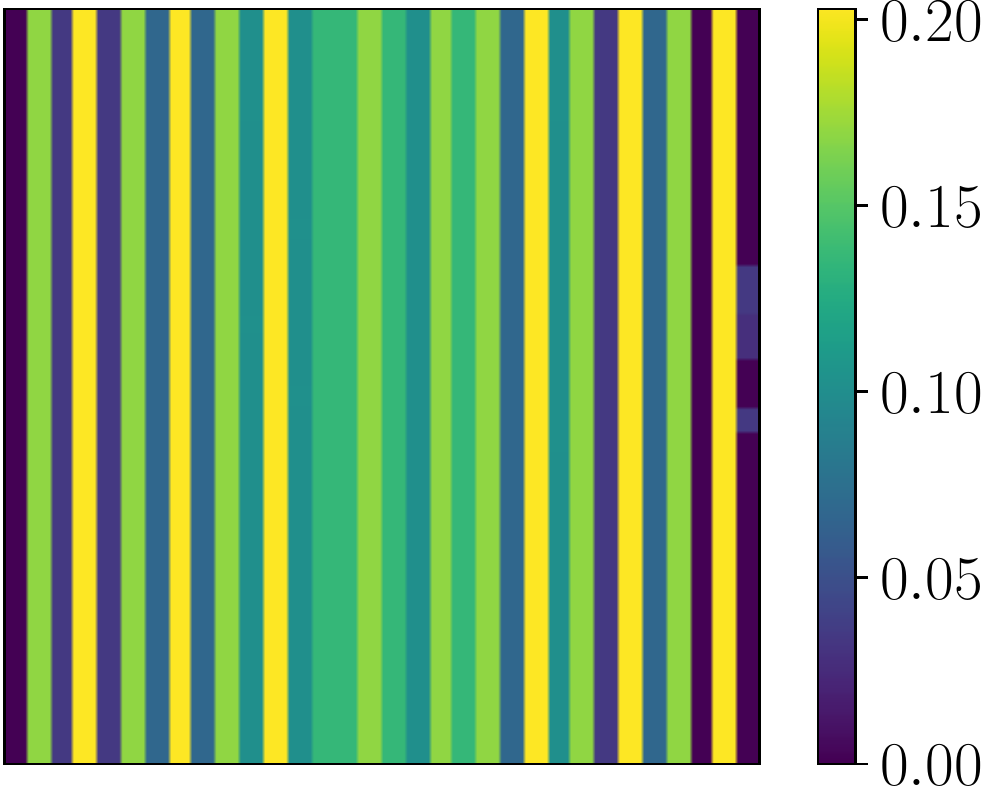}}
        \caption{test $\bm{\Delta}_{\mathrm{a}}$
        }\label{fig:trigger-test}
      \end{subfigure}
      \caption{
      As the number of poison examples,  \(m\), decrease, NTBA makes each poison example stronger  by reducing the magnitude of the pixels of the train-time perturbation $\bm{\Delta}_{\mathrm{p}}$. }\label{fig:trigger-heatmaps} 
    \end{figure}
  \end{minipage}\hspace{1em}
    \begin{minipage}{.4\textwidth}
    \begin{figure}[H]
      \centering
      \includegraphics[width=1.0\textwidth]{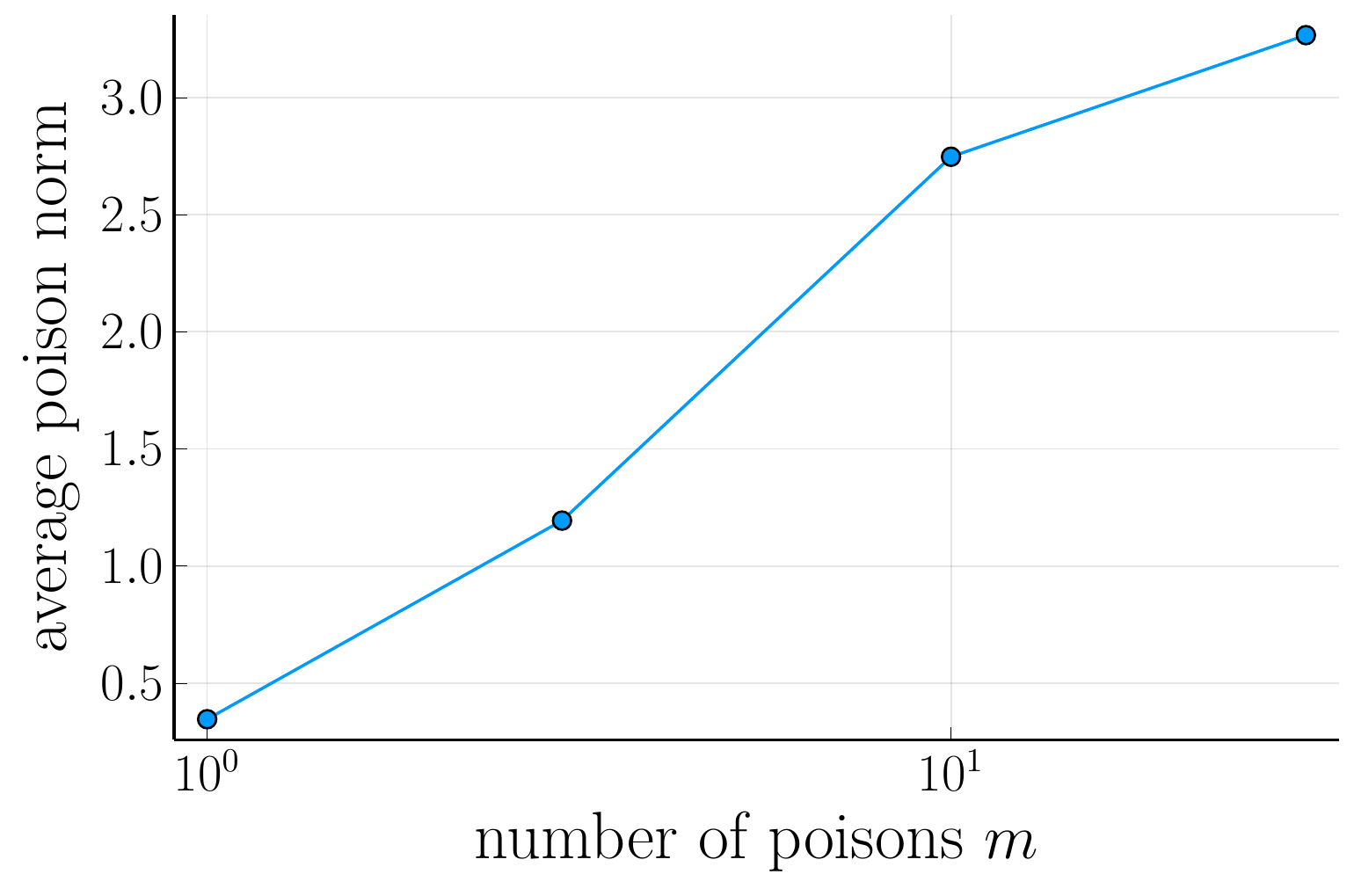}
      \caption{The average norm difference, $\|\bm{\Delta}_{\mathrm{p}}\|$, between each poison image automatically discovered  by NTBA and the closest clean image, after running NTBA with different choices of \(m\).}\label{fig:trigger-norms}
    \end{figure}
  \end{minipage}
\end{center}

To explain this phenomenon, we take Taylor approximations of \(\phi\) at \(\widetilde{\bm{x}}_{\mathrm{p}}\) and \(\widetilde{\bm{x}}_{\mathrm{a}}\) and obtain,
\begin{align*}
  \MoveEqLeft{f\del{\bm{x}_{\mathrm{a}}; D_{\mathsf{d}} \cup \set{\del{\bm{x}_{\mathrm{p}}, y_{\mathrm{p}}}}} - f\del{\bm{x}_{\mathrm{a}}; D_{\mathsf{d}}}}\nonumber\\
  &\approx \frac{\del{\phi\del{\widetilde{\bm{x}}_{\mathrm{p}}} + \Dif\phi\del{\widetilde{\bm{x}}_{\mathrm{p}}}\bm{\Delta}_{\mathrm{p}}}\del{I - P}\del{\phi\del{\widetilde{\bm{x}}_{\mathrm{a}}} + \Dif\phi\del{\widetilde{\bm{x}}_{\mathrm{a}}}\bm{\Delta}_{\mathrm{a}}}^\top}{\del{\phi\del{\widetilde{\bm{x}}_{\mathrm{p}}} + \Dif\phi\del{\widetilde{\bm{x}}_{\mathrm{p}}}\bm{\Delta}_{\mathrm{p}}}\del{I - P}\del{\phi\del{\widetilde{\bm{x}}_{\mathrm{p}}} + \Dif\phi\del{\widetilde{\bm{x}}_{\mathrm{p}}}\bm{\Delta}_{\mathrm{p}}}^\top}\del{y_{\mathrm{p}} - f\del{\bm{x}_{\mathrm{p}}; D_{\mathsf{d}}}}\\
  &= \underbrace{\frac{\inner{\Dif\phi\del{\widetilde{\bm{x}}_{\mathrm{p}}}\bm{\Delta}_{\mathrm{p}},  \Dif\phi\del{\widetilde{\bm{x}}_{\mathrm{a}}}\bm{\Delta}_{\mathrm{a}}}_{\del{I - P}}}{\norm{\Dif\phi\del{\widetilde{\bm{x}}_{\mathrm{p}}}\bm{\Delta}_{\mathrm{p}}}_{\del{I - P}}^2}}_{\triangleq A}
  \underbrace{\del{y_{\mathrm{p}} - f\del{\bm{x}_{\mathrm{p}}; D_{\mathsf{d}}} }}_{ \approx 2} \;,
\end{align*} 
where \(\Dif\phi\del{\widetilde{\bm{x}}}\) denotes the Jacobian of the feature mapping \(\phi\) at \(\widetilde{\bm x}\) and w.l.o.g.~we assume that \(y_{\mathrm{p}} = 1\) and \(f\del{\bm{x}_{\mathrm{p}}; D_{\mathsf{d}}} \approx -1\).
The last step follows because \(\del{I - P}\phi\del{\widetilde{\bm x}_{\mathrm{a}}}= \del{I - P}\phi\del{\widetilde{\bm x}_{\mathrm{p}}} = \bm 0\).
Note that if \(A = 1\), then \(f\del{\bm{x}_{\mathrm{a}}; D_{\mathsf{d}} \cup \set{\del{\bm{x}_{\mathrm{p}}, y_{\mathrm{p}}}}} = y_{\mathrm{p}}\) which would imply a succesful attack for \(\bm{x}_{\mathrm{a}}\).
Since the goal of the attack is to control the prediction whenever \(\bm{\Delta}_{\mathrm{a}}\) is applied to {\em any} clean point \(\widetilde{\bm{x}}\), there may exist some \(\widetilde{\bm{x}}\) where
\(\Dif\phi\del{\widetilde{\bm{x}}_{\mathrm{a}}}\bm{\Delta}_{\mathrm{a}}\) does not align well with \(\Dif\phi\del{\widetilde{\bm{x}}_{\mathrm{p}}}\bm{\Delta}_{\mathrm{p}}\), which would make the numerator of \(A\) small. 
For the backdoor to succeed for these points, \(\norm{\Dif\phi\del{\widetilde{\bm{x}}_{\mathrm{p}}}\bm{\Delta}_{\mathrm{p}}}_{\del{I - P}}\) must be small enough to overcome this misalignment, since the denominator of \(A\) scales as \(\norm{\Dif\phi\del{\widetilde{\bm{x}}_{\mathrm{p}}}\bm{\Delta}_{\mathrm{p}}}_{\del{I - P}}^2\) while the numerator scales as \(\norm{\Dif\phi\del{\widetilde{\bm{x}}_{\mathrm{p}}}\bm{\Delta}_{\mathrm{p}}}_{\del{I - P}}\). 
In particular, for the attack to succeed on a set of poisoned data points $X_{\mathrm{a}}$,  we need 
\begin{equation}
\norm{\Dif\phi\del{\widetilde{\bm{x}}_{\mathrm{p}}}\bm{\Delta}_{\mathrm{p}}}_{\del{I - P}} 
\le 
c\,\del*{\min_{\bm{x}_{\mathrm{a}} \in X_{\mathrm{a}}} \inner*{\frac{\Dif\phi\del{\widetilde{\bm{x}}_{\mathrm{p}}}\bm{\Delta}_{\mathrm{p}}}{\norm{\Dif\phi\del{\widetilde{\bm{x}}_{\mathrm{p}}}\bm{\Delta}_{\mathrm{p}}}_{\del{I - P}}},  \Dif\phi\del{\widetilde{\bm{x}}_{\mathrm{a}}}\bm{\Delta}_{\mathrm{a}}}_{\del{I - P}}} \;, \label{eq:intuition}
\end{equation}
for some constant $c>0$. 
Note that the test-time trigger \(\bm{\Delta}_{\mathrm{a}}\) and therefore the distribution of \(\bm{x}_{\mathrm{a}}\)'s is fixed.
Therefore \cref{eq:intuition} can be satisfied by choosing the train-time perturbation \(\bm{\Delta}_{\mathrm{p}}\) to have small enough norm on the LHS of \cref{eq:intuition}. 
This implies that smaller perturbations in the train-time poison data are able to successfully change the predictions on more examples at test-time, and hence they correspond to a stronger attack.
We can make this connection more realistic by considering multiple poisoned examples injected together.
As the size \(m \triangleq \abs{D_{\mathrm{p}}}\) of the injected poisoned dataset \(D_{\mathrm{p}}\) increases, we may distribute the poisoned examples so that each test point \(\bm{x}_{\mathrm{a}}\) is covered by some poison point \(\bm{x}_{\mathrm{p}} \in X_{\mathrm{p}}\) that aligns well with it. 
Since the worst-case alignment between poison and test data will be higher, the RHS of \cref{eq:intuition} will be larger so the LHS may be larger as well.
This means that for each poison, the size of the trigger \(\norm{\Dif\phi\del{\widetilde{\bm{x}}_{\mathrm{p}}}\bm{\Delta}_{\mathrm{p}}}_{\del{I - P}}\) may be larger (and still achieve a  high attack success rate) when we are adding more poison data.

Two further insights from \cref{eq:intuition} shows the strengths of NTBA.
First, \cref{eq:intuition} suggests that there is potential for improvement by designing train-time perturbations \(\bm{\Delta}_{\mathrm{p}}\) that adapt to the local geometry of the feature map, represented by \(\Dif\phi\del{\widetilde{\bm{x}}_{\mathrm{p}}},\Dif\phi\del{\widetilde{\bm{x}}_{\mathrm{a}}}\), around clean data points 
\(\widetilde{\bm{x}}_{\mathrm{p}}, \widetilde{\bm{x}}_{\mathrm{a}}\). 
We propose using a data-driven optimization to automatically discover such strong perturbations.
Second, our analysis suggests that we need the knowledge of the manifold of clean data to design strong poisoned images that are close to the manifold. 
Since the manifold is challenging to learn from data, we explicitly initialize the optimization near carefully selected clean images 
\(\widetilde{\bm{x}}_{\mathrm{p}}\), allowing the optimization to easily control the size of the difference \(\bm{\Delta}_{\mathrm{p}}\).
We show in our ablation study in \cref{sec:ablation} that both components are critical for designing strong attacks.

\section{Why are NNs so vulnerable to backdoor attacks?}\label{sec:kernel} 

NTBA showcases the vulnerability of DNNs to backdoor attacks. 
We investigate the cause of such vulnerability  by comparing the infinite-width NTK with the standard Laplace kernel. 


\medskip\noindent{\bf NTK gives more influence to far away data points.} 
The Laplace kernel gives more influence to  points that are closer. For example, Laplace-kernel linear regression converges to a \(1\)-nearest neighbor predictor in the limit as the bandwidth \(\sigma \to 0\), which is naturally robust against few-shot backdoor attacks. 
In contrast, we conjecture that the NTK (and hence neural network) gives \emph{more} influence to points as they become more distant. 
We confirm this by visualizing the two kernels with matched bandwidths in the normal and tangent direction to a unit sphere. For details, we refer to \cref{sec:kernel:1d}.

\begin{figure}[h]
    \centering
    \begin{subfigure}{0.45\textwidth}
    \centering
      \includegraphics[width=0.9\linewidth, trim=0 2em 0 2.5em]{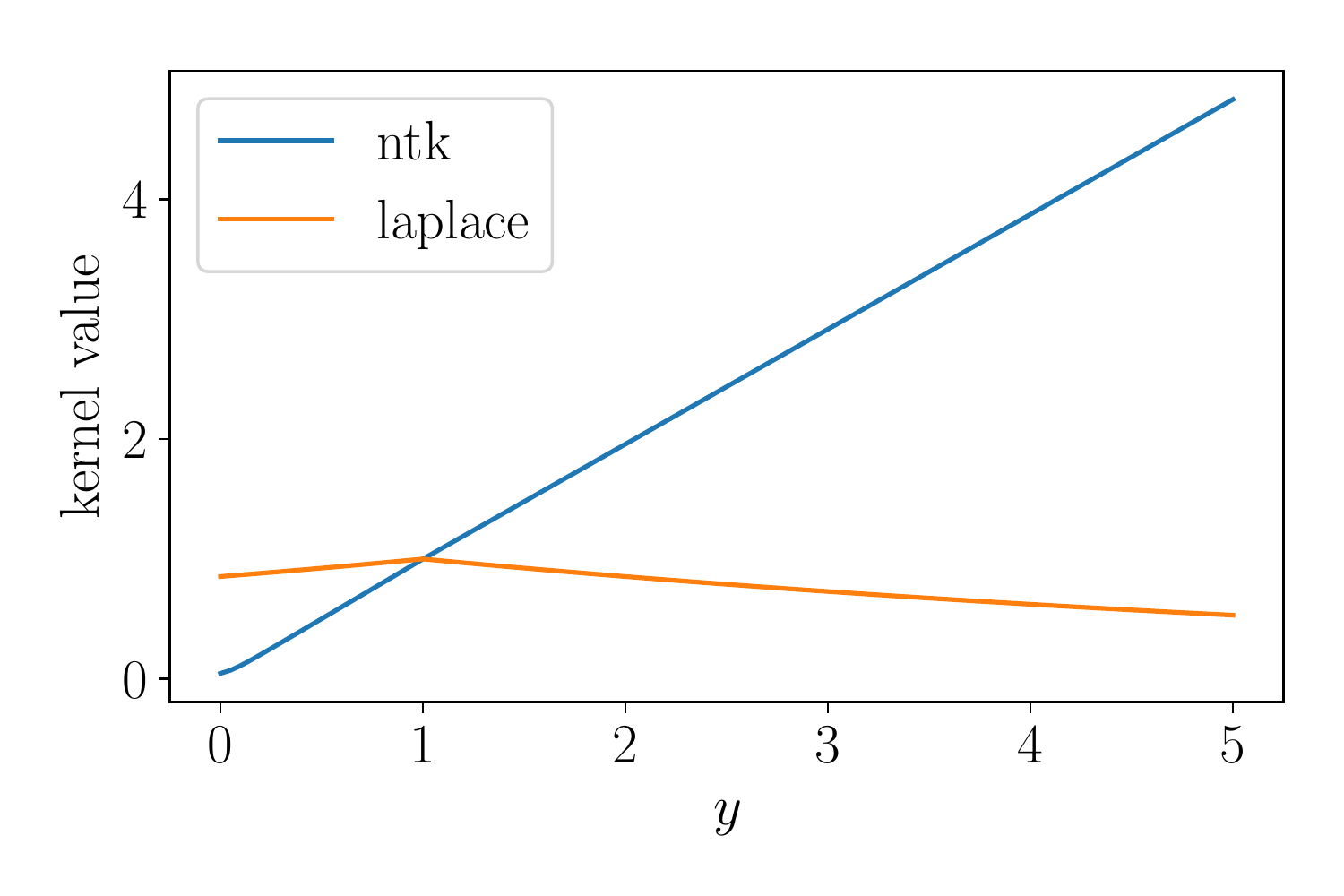}
      \caption{
        Kernel behavior {\em normal} to unit sphere. 
        The plot shows \(K\del{\bm{e}_1, y\bm{e}_1}\) for both the NTK and Laplace kernels where \(\bm{e}_1\) is a unit vector. 
        Note that the NTK increases with \(y\), while the Laplace kernel peaks at \(y = 1\).
      }\label{fig:ntk-vs-laplace-normal}
    \end{subfigure}\hspace{1em}%
    \begin{subfigure}{0.47\textwidth}
      \centering
      \includegraphics[width=0.9\linewidth, trim=0 2em 0 2.5em]{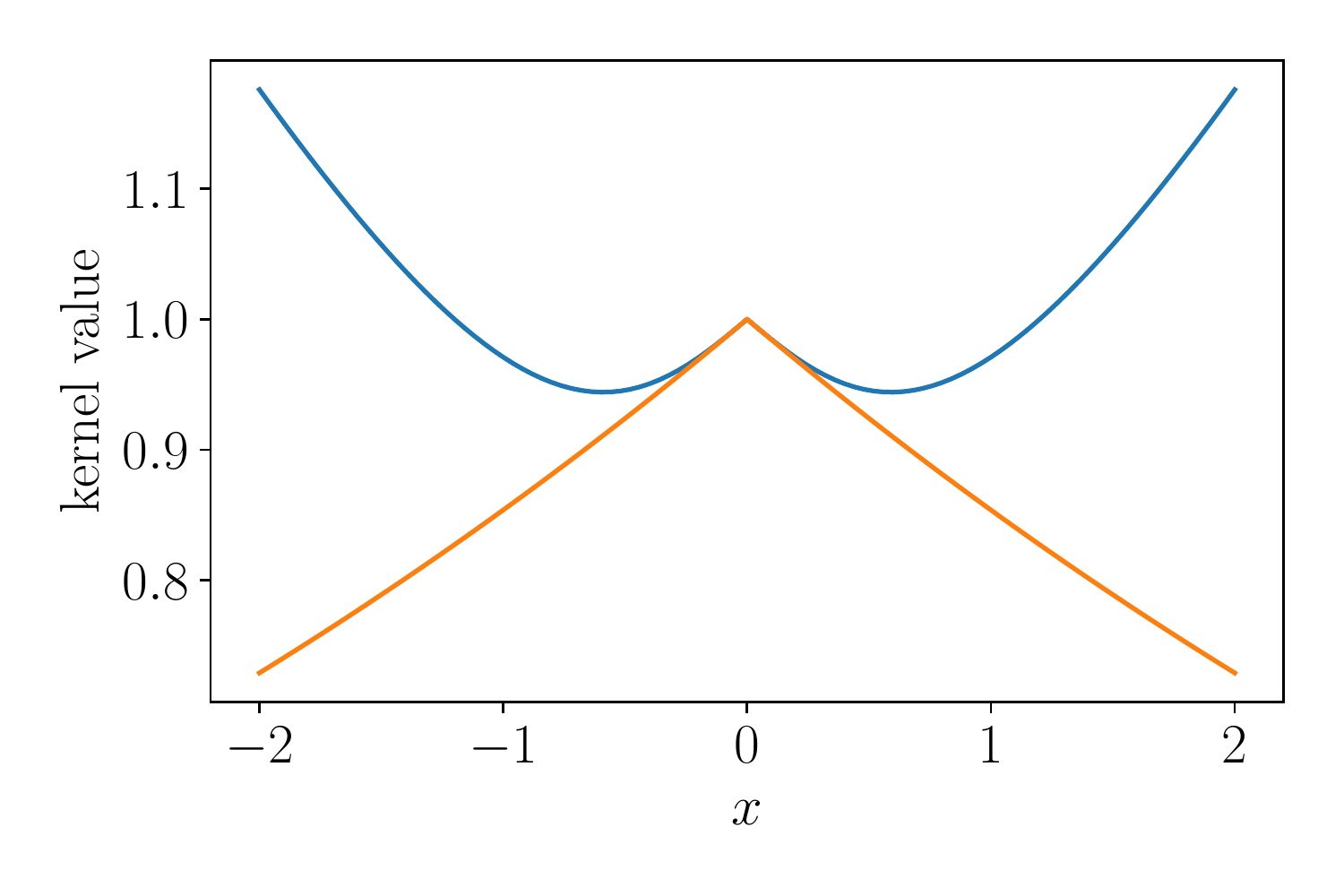}
      \caption{Kernel behavior {\em tangent} to unit sphere. The plot shows \(K\del{\bm{e}_1, \bm{e}_1 + x\bm{e}_2}\) for both the NTK and Laplace kernels where \(\bm{e}_1,\bm{e}_2\) are orthogonal unit vectors.  The two kernel behave similarly near \(x = 0\) but diverge rapidly away from \(0\).}\label{fig:ntk-vs-laplace-tangent}
    \end{subfigure}
   \caption{
    Kernel behavior off the unit sphere shows that  the NTK approaches oblique asymptotes as either \(\abs{x}\) or \(y\), increases, while the Laplace kernel decreases in the same limit.
    }\label{fig:ntk-vs-laplace}
\end{figure}

\medskip\noindent{\bf NTK is more vulnerable to few-shot backdoor attacks.}
We demonstrate with a toy example that NTK is more influenced by far away points, which causes it to be more vulnerable to some few-shot backdoor attacks. 
We use a synthetic backdoor dataset in 3 dimensions \(\del{x, y, z}\) consisting of clean data  \(\del{\begin{bmatrix} \widetilde{x} & 1 & 0 \end{bmatrix}^\top, 1}\) and  \(\del{\begin{bmatrix} \widetilde{x} & -1 & 0\end{bmatrix}^\top, -1}\) for  \(\widetilde{x} \in \set{-100, -99,\dots, 100}\).
Here, the \(x\) dimension represents the diversity of the dataset, the \(y\) dimension represents the true separation between the two classes, and the \(z\) dimension is used to trigger the backdoor attack.
We choose test-time trigger \(P\del{\bm v} = \bm v + \begin{bmatrix} 0 & 0 & 1\end{bmatrix}^\top\) for a clean negative labelled point \(\bm{v}\)
and add a single train-time poison data point  \(\del{0, -1,\widetilde{z}}\).
For the Laplace kernel, we compute the best choice of \(\widetilde{z}\) which is \(\widetilde{z} = 1\).
For the NTK, the backdoor increases in strength as \(\widetilde{z} \to 0^+\) (we chose \(\widetilde{z} = \num{1e-6}\)).

\begin{figure}[h] 
  \centering
  \begin{subfigure}[t]{0.45\textwidth}
    \centering
    \includegraphics[width=0.9\textwidth, trim=0 2em 0 4em]{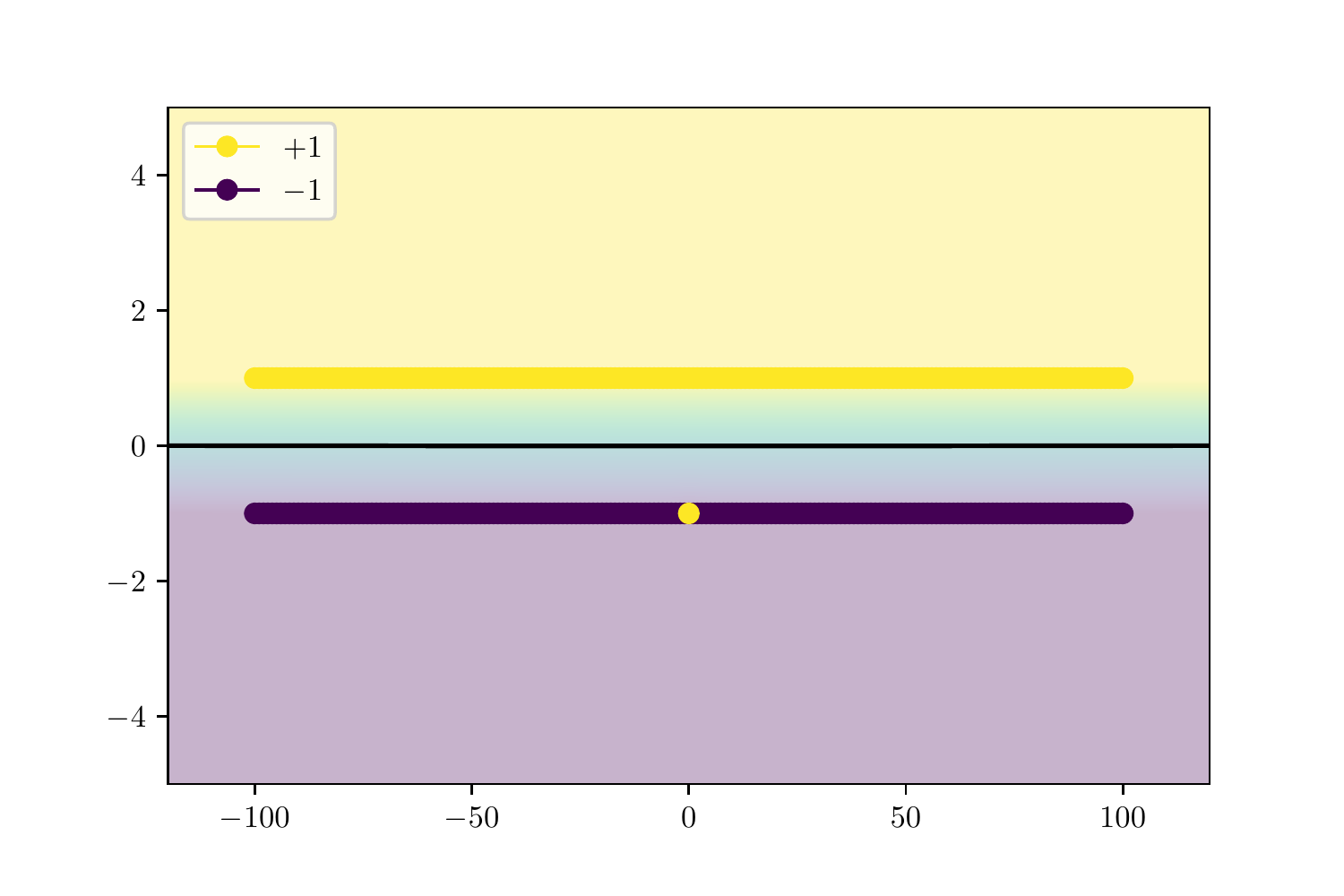}
    \put(-22,-1){\footnotesize $x$}
    \put(-176,100){\footnotesize $y$}
    \caption{NTK, decision boundary at $z=0$}
    \label{fig:overhead-boundary-ntk-clean}
  \end{subfigure}%
  \begin{subfigure}[t]{0.45\textwidth}
    \centering
    \includegraphics[width=0.9\textwidth, trim=0 2em 0 4em]{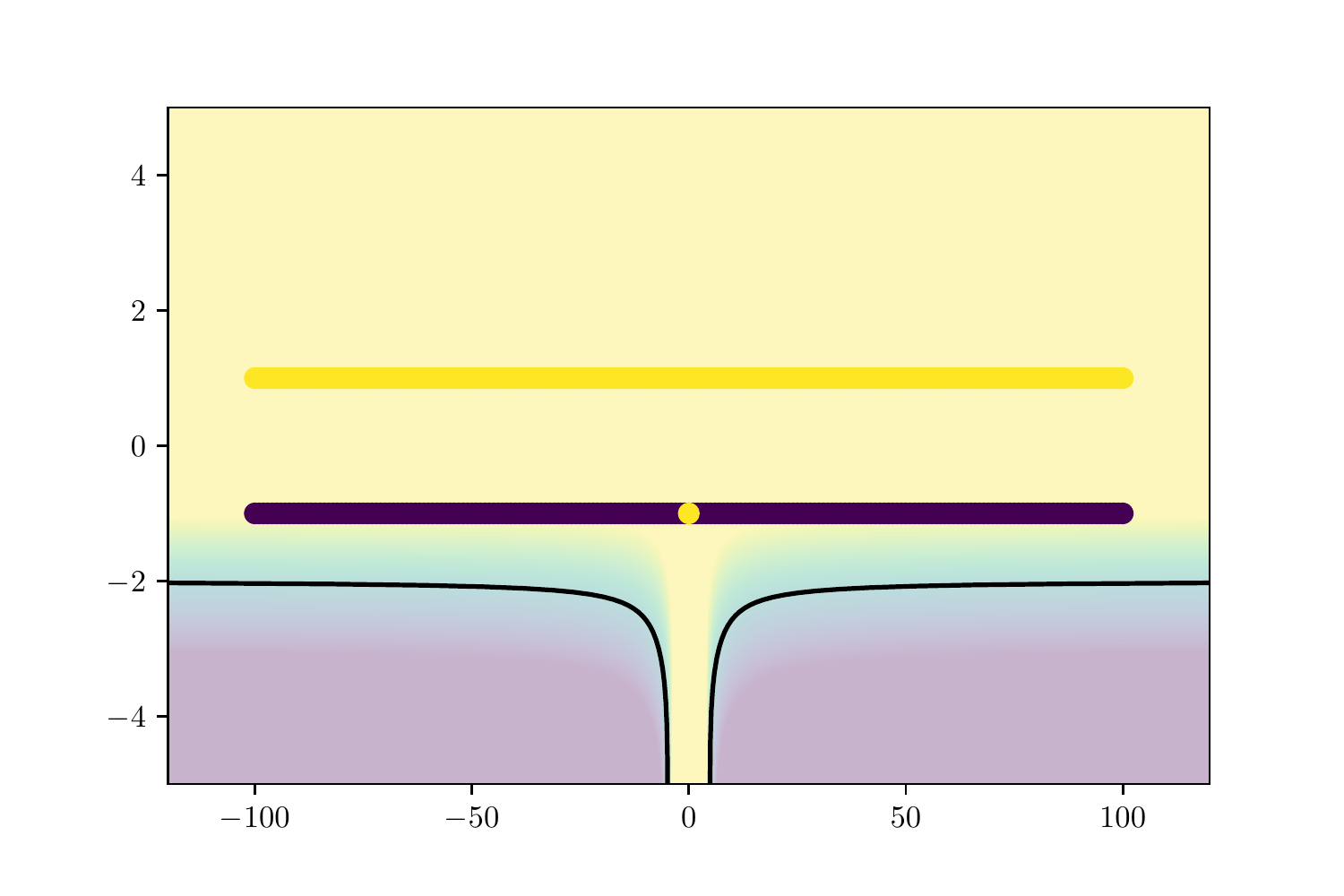}
    \put(-22,-1){\footnotesize $x$}
    \put(-176,100){\footnotesize $y$}
    \caption{NTK, decision boundary at $z=1$}
    \label{fig:overhead-boundary-ntk-poison}
  \end{subfigure}
  \begin{subfigure}[t]{0.45\textwidth}
    \centering
    \includegraphics[width=0.9\textwidth, trim=0 2em 0 2em]{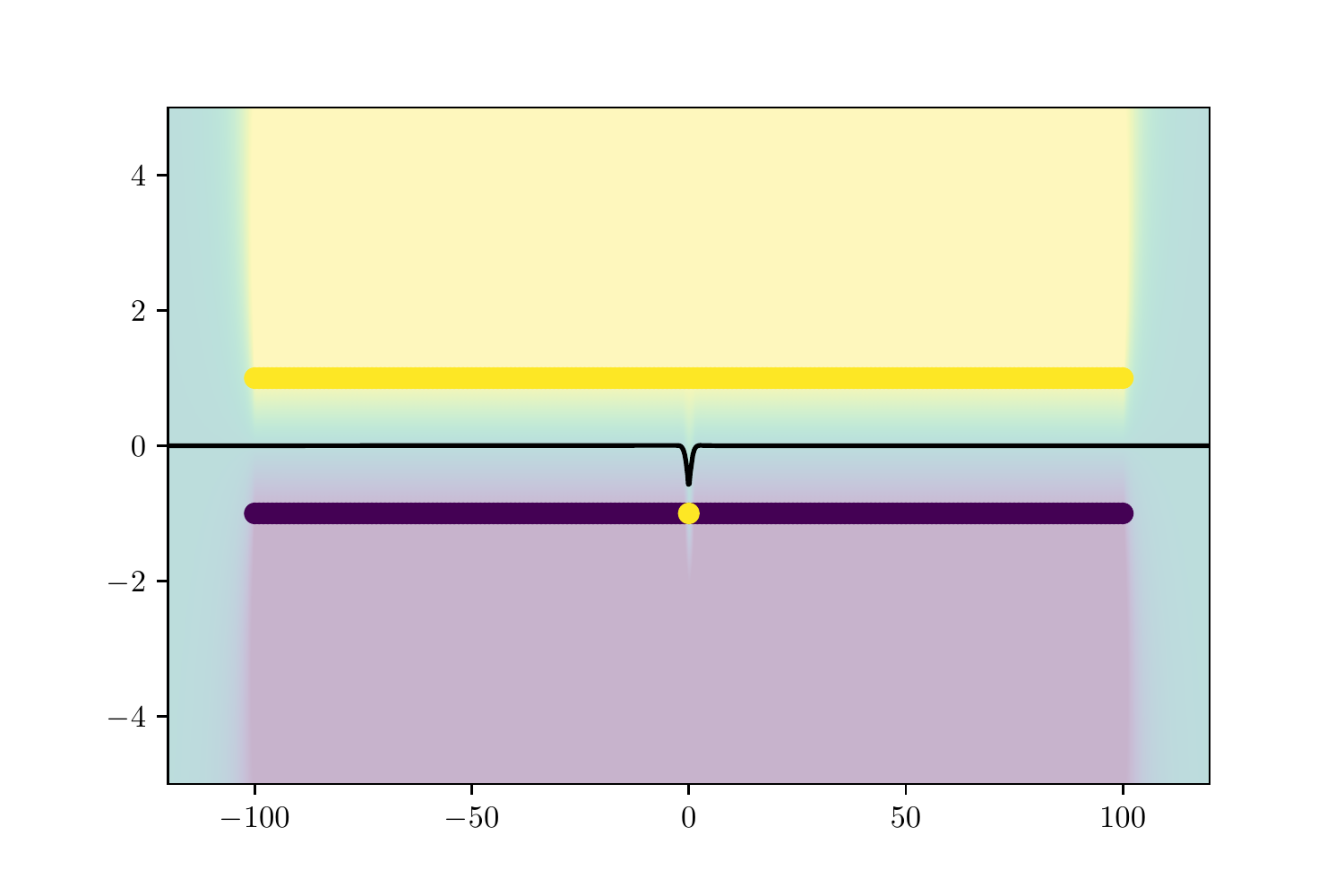}
    \put(-22,-1){\footnotesize $x$}
    \put(-176,100){\footnotesize $y$}
    \caption{Laplace, decision boundary at $z=0$}\label{fig:overhead-boundary-laplace-clean}
  \end{subfigure}%
  \begin{subfigure}[t]{0.45\textwidth}
    \centering
    \includegraphics[width=0.9\textwidth, trim=0 2em 0 2em]{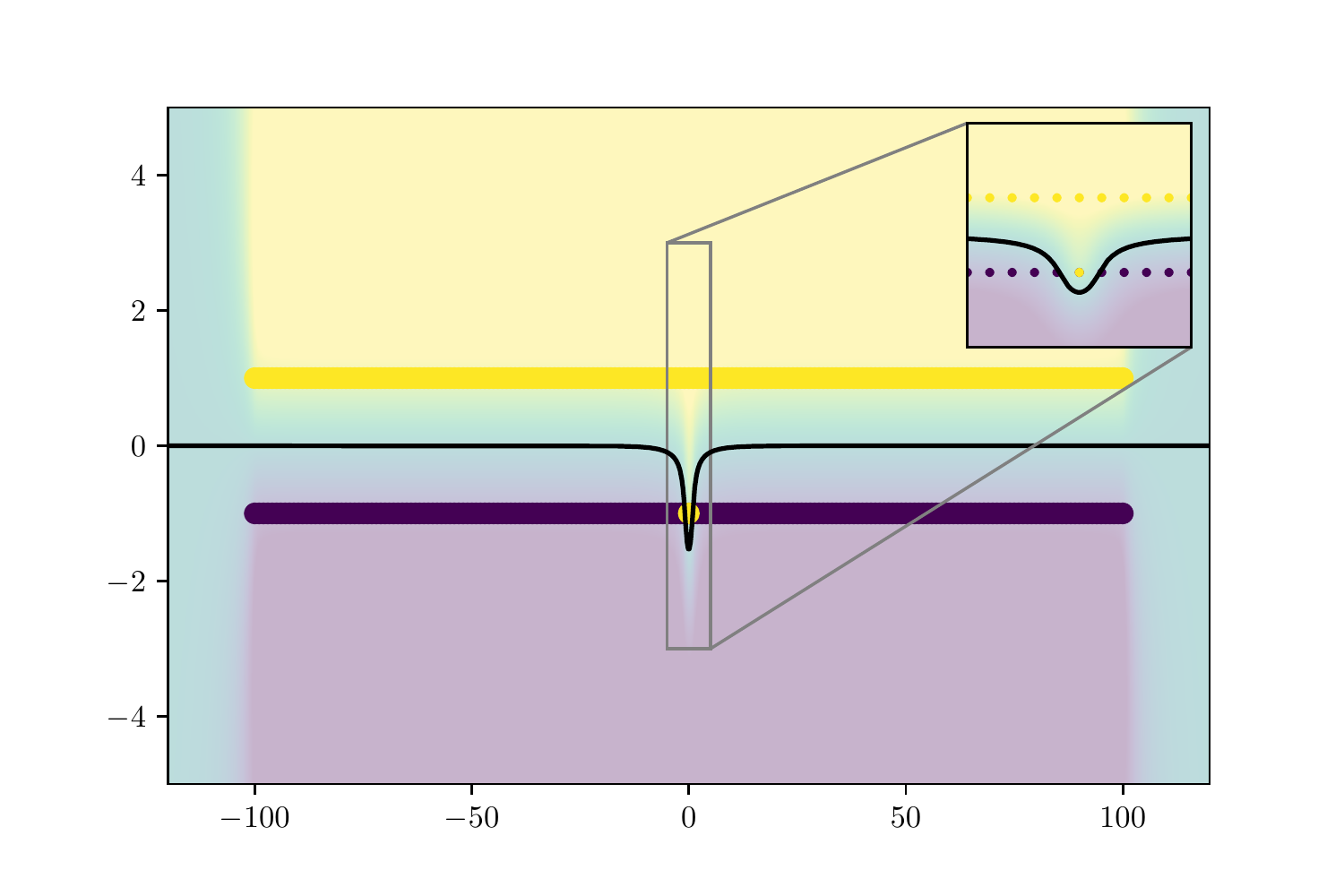}
    \put(-22,-1){\footnotesize $x$}
    \put(-176,100){\footnotesize $y$}
    \caption{Laplace, decision boundary at $z=1$}\label{fig:overhead-boundary-laplace-poison}
  \end{subfigure}
  \caption{%
    The decision boundaries at $z=0$ (black solid line) and corresponding predictions (background shading) on the $z=0$ plane are similar for NTK and Laplace kernel, explaining the similar clean accuracy  in \cref{tab:ntk-vs-laplace-cifar}.  The decision boundary at $z=1$ shows that the trigger fails to generalize to test examples for Laplace kernel. 
    All points in the training dataset are shown regardless of their \(z\)-coordinate.
    Note that the solid bars are actually discrete points with overlapping markers and the yellow point at \(\del{0, -1}\) is the single poison point.
  }\label{fig:overhead-boundary}
\end{figure}

In \cref{fig:overhead-boundary-laplace-poison} we see that the backdoor is not successful for the Laplace kernel, only managing to flip the prediction of a single backdoor test point.
This is because the influence of the poison point rapidly drops off as \(\abs{x}\) increases. 
For \(\abs{x} > 10\) the poison has a negligible effect on the predictions of the model.
In contrast, we see in \cref{fig:overhead-boundary-ntk-poison} that the NTK was successfully backdoored and the predictions of all test points can be flipped by the trigger \(P(\cdot)\).
This is due to the influence of the poison point remains high even from a great distance. 

\section{Conclusion}
\label{sec:conclusion}

We study the fundamental trade-off in backdoor attacks between the number of poisoned examples that need to be injected and the resulting attack success rate and bring a new perspective on backdoor attacks, borrowing  tools from kernel methods. 
Through an ablation study in \cref{tab:alg-ablation}, we demonstrate that every component in the Neural Tangent Backdoor Attack (NTBA) is necessary in finding train-time poison examples that are significantly more powerful. 
We experiment on  CIFAR and ImageNet subsets with WideResNet-34-5 and ConvNeXt architectures for periodic triggers and patch triggers, and show that, in some cases, NTBA requires an order of magnitude smaller number of poison examples to reach a target attack success rate compare to the baseline.

Next, we borrow the analysis  of kernel linear regression to provide an interpretation of the NTBA-designed poison examples. The strength of the attack increases as we decrease the magnitude of the trigger used in the poison training example, especially when it is coupled with a clean data that is close in the image space.
Finally, we compare neural tangent kernel and the Laplace kernel to investigate why the NTK is so vulnerable to backdoor attacks. 
Although this attack may be used for harmful purposes, our goal is to show the existence of strong backdoor attacks to motivate continued research into backdoor defenses and inspire practitioners to carefully secure their machine learning pipelines. 
The main limitation of our approach is a lack of scalability, as the cost of computing the NTK predictions \cref{eq:bd-loss} scales cubically in the number of datapoints.
In the future, we plan to apply techniques for scaling the NTK \citep{meanti2020kernel,rudi2017falkon,zandieh2021scaling} to our attack.

\section*{Acknowledgement} 
JH is supported in part by 
NSF Graduate Research Fellowships Program (GRFP) and Microsoft. 
SO is supported in part by NSF grants CNS-2002664, IIS-1929955, DMS-2134012, CCF-2019844 as a part of NSF Institute for Foundations of Machine Learning (IFML), and CNS-2112471 as a part of NSF AI Institute for Future Edge Networks and Distributed Intelligence (AI-EDGE). 

\bibliography{ref}
\bibliographystyle{style/iclr2023_conference}

\newpage
\appendix

\section{Related work}
\label{sec:related} 

We survey relevant train-time attacks. 

\subsection{Backdoor attacks}

Backdoor attacks as presented in \cref{sec:intro} are introduced in \cite{gu2017badnets}.
In backdoor attacks, the two most important design choices are the choice of trigger \(P\) and the method of producing the poison data \(X_{\mathrm{p}}\).
Many works design \(P\) to appear benign to humans \cite{gu2017badnets,barni2019new,liu2020reflection,nguyen2020wanet} or directly optimize \(P\) to this end \cite{li2020invisible,doan2021lira}.
Poison data \(X_{\mathrm{p}}\) has been constructed to include no mislabeled examples \cite{turner2019label,zhao2020clean} and optimized to evade detection through visual inspection \cite{saha2020hidden} and statistical inspection of latent representations \cite{shokri2020bypassing,doan2021backdoor,xia2022enhancing,chen2017targeted}.
Such backdoor attacks have been demonstrated in a wide variety of settings, including federated learning \cite{wang2020attack,bagdasaryan2020backdoor,sun2019can}, transfer learning \cite{yao2019latent,saha2020hidden}, and generative models \cite{salem2020baaan,rawat2021devil}.
However, our goal of designing strong \emph{few-shot backdoor attacks} has not been addressed with an exception of an influential earlier work of \cite{koh2022stronger}.
We consider the same threat model  as in \citep{koh2022stronger} where the attacker has information about the network's architecture and training data.
However, our results are incomparable to those of \citep{koh2022stronger} which focuses on linear models.
The KKT attack of \citep{koh2022stronger} leveraging decoy parameters cannot be used when the input dimension is far smaller than the parameter dimension and the influence attack of \citep{koh2022stronger} cannot scale to large models, such as the WideResNet we use in our experiments.

Few-shot data attacks have been studied in contexts other than backdoor attacks. 
In \emph{targeted} backdoor attacks, the attacker aims to control the network's output on a specific test instance \cite{shafahi2018poison,barni2019new,guo2020practical,aghakhani2021bullseye}.
\emph{Data poisoning attacks} are similar to backdoor attacks with the alternate goal of reducing the generalization performance of the resulting model.
Poison data \(X_{\mathrm{p}}\) has been optimized to produce stronger data poisoning attacks using influence functions \cite{koh2022stronger,yang2017generative,munoz2017towards} and the neural tangent kernel \cite{yuan2021neural}.


Following \cite{gu2017badnets}, there has also been substantial work on detecting and defending against backdoor attacks.
When the defender has access to known-clean data, they can filter the data using outlier detection
\cite{liang2018enhancing,lee2018simple,steinhardt2017certified}, retrain the network so it forgets the backdoor \cite{liu2018fine}, or train a new model to test the original for a backdoor \cite{kolouri2020universal}.
Other defenses assume \(P\) is an additive perturbation with small norm \cite{wang2019neural,chou2020sentinet}, rely on smoothing \cite{wang2020certifying,weber2020rab}, filter or penalize outliers without clean data \cite{gao2019strip,sun2019can,steinhardt2017certified,blanchard2017machine,pillutla2019robust,tran2018spectral,hayase2021spectre} or use Byzantine-tolerant distributed learning techniques \cite{blanchard2017machine,alistarh2018byzantine,chen2018draco}.
Backdoors cannot be detected in planted neural networks in general \cite{goldwasser2022planting}.


\section{Implementation details}

In \cref{sec:ntba}, we give a brief description of the Neural Tangent Backdoor Attack.
Further details regarding the implementation are given here.

\subsection{Efficient gradient calculation}\label{apx:ntba-grad}

We propose several techniques to make the backward pass described in \cref{sec:ntba} more efficient, which is critical for scaling NTBA to the neural networks that we are interested in.

\subsubsection{Custom batching in backwards pass}
In order to efficiently minimize the loss \(\mathcal{L}_{\mathrm{backdoor}}\) with respect to \(X_{\mathrm{p}}\), we require the gradient \(\spd{\mathcal{L}_{\mathrm{backdoor}}}{X_{\mathrm{p}}}\).
One straightforward way to calculate the gradient is to rely on the JAX autograd system to differentiate the forward process.
Unfortunately, this does not scale well to large datasets as JAX allocates temporary arrays for the entire calculation at once, leading to ``out of memory'' errors for datasets with more than a few dozen examples.
Instead, we write out the backward process in the style of \cite{nguyen2021dataset} and manually contract the gradient tensors, as shown in \cref{alg:bd-grad}.

The kernel matrix \(K_{\mathrm{d,dta}}\) does not depend on \(X_{\mathrm{p}}\) and so we calculate it once at the beginning of our optimization.
Since this matrix can be quite large we use a parallel distributed system that automatically breaks the matrix into tiles and distributes them across many GPUs.
The results are then collected and assembled into the desired submatrix.
We use the technique of \cite{novak2021fast} to compute the kernel matrix tiles which gave a factor of 2 speedup over the direct method of computing the inner products of the network gradients.

\begin{algorithm}
  \caption{Backdoor loss and gradient}\label{alg:bd-grad}
  \DontPrintSemicolon
  \KwIn{Kernel matrix \(K_{\mathrm{d,dta}}\), data \(\del{X_{\mathrm{dta}}, \bm{y}_{\mathrm{dta}}}\) and \(\del{X_{\mathrm{p}}, \bm{y}_{\mathrm{p}}}\).}
  \KwOut{Backdoor design loss \(\mathcal{L}_{\mathrm{backdoor}}\) and gradient \(\pd{\mathcal{L}_{\mathrm{backdoor}}}{X_{\mathrm{p}}}\).}
  Compute Kernel matrix \(K_{\mathrm{p,pdta}}\) from \(X_{\mathrm{dta}}\) and \(X_{\mathrm{p}}\) using \cite{novak2021fast}.\;
  Compute the loss \(\mathcal{L}_{\mathrm{backdoor}}\) via \cref{eq:bd-loss}.\;
  Compute the gradient matrix \(\pd{\mathcal{L}_{\mathrm{backdoor}}}{K_{\mathrm{p,pdta}}}\) by automatic differentiation \cite{jax2018github} of \cref{eq:bd-loss}.\;
  Compute the tensor \(\pd{K_{\mathrm{p,pdta}}}{X_{\mathrm{p}}}\) as described in \cref{sec:fast-kgrad}.\;\label{alg:bd-grad:kgrad}
  \KwRet \(\mathcal{L}_{\mathrm{backdoor}}\), tensor contraction \(\pd{\mathcal{L}_{\mathrm{backdoor}}}{X_{\mathrm{p}}} = \del{\pd{\mathcal{L}_{\mathrm{backdoor}}}{K_{\mathrm{p,pdta}}}}_{i,j}\del{\pd{K_{\mathrm{p,pdta}}}{X_{\mathrm{p}}}}_{i,j,l}\).\;\label{alg:bd-grad:contract}
\end{algorithm}

Additionally, the form of \cref{alg:bd-grad} admits a significant optimization where lines~\ref{alg:bd-grad:kgrad}~and~\ref{alg:bd-grad:contract} can be fused, so that slices of \(\spd{K_{\mathrm{p,pdta}}}{X_{\mathrm{p}}}\) are computed, contracted with slices of \(\spd{\mathcal{L}_{\mathrm{backdoor}}}{K_{\mathrm{p,pdta}}}\), and discarded in batches.
Choosing the batch size allows us to balance memory usage and the speedup offered by vectorization on GPUs.
Additionally these slices are again distributed across multiple GPUs and the contractions are be performed in parallel before a final summation step.

\subsubsection{Efficient empirical neural tangent kernel gradients}\label{sec:fast-kgrad}

In \cref{alg:bd-grad}, the vast majority of the total runtime is spent in the calculation of slices of \(\spd{K_{\mathrm{p,pdta}}}{X_{\mathrm{p}}}\) on line~\ref{alg:bd-grad:kgrad}.
Here we will focus on calculating a single \(1 \times 1 \times k\) slice of \(\spd{K_{\mathrm{p,pdta}}}{X_{\mathrm{p}}}\).
Letting \(\Dif_{\bm{x}}\) denote the partial Jacobian operator w.r.t. argument \(\bm{x}\), the slice we are computing is exactly
\begin{equation}
  \text{\(\Dif_{\bm{x}} K\del{\bm{x}, \bm{y}}\) where \(K\del{\bm{x}, \bm{y}} = \inner{\Dif_{\bm{\theta}} \del{\bm{x}; \bm{\theta}}, \Dif_{\bm{\theta}} f\del{\bm{y}; \bm{\theta}}}\)}\label{eq:kgrad}
\end{equation}
for some \(\bm{x}, \bm{y} \in \RR^k\).\footnote{Extra care must be taken to compute \(\spd{K_{\mathrm{p,p}}}{X_{\mathrm{p}}}\). These details are omitted for simplicity.}

Let \(\Dif_{\bm{x}}^{\rightarrow}\) and \(\Dif_{\bm{x}}^{\leftarrow}\) respectively denote that the Jacobian will be computed using forward or reverse mode automatic differentiation.
Since \(K\) is scalar-valued, it is natural to compute \cref{eq:kgrad} as \(\Dif_{\bm{x}}^{\leftarrow} \inner{\Dif_{\bm{\theta}}^{\leftarrow} f\del{\bm{x}; \bm{\theta}}, \Dif_{\bm{\theta}}^{\leftarrow} f\del{\bm{y}; \bm{\theta}}}\).
However this approach is very slow and requires a large amount of memory due to the intermediate construction of a \(k \times d\) tensor representing \(\Dif_{\bm{x}} \Dif_{\bm{\theta}} f\del{\bm{x}; \bm{\theta}}\).
Instead, assuming that \(f\) is twice continuously differentiable, we can exchange the partial derivatives and compute \( \del{\Dif_{\bm{\theta}}^{\rightarrow} \Dif_{\bm{x}}^{\leftarrow}f\del{\bm{x}; \bm{\theta}}}^\top \del{\Dif_{\bm{\theta}}^{\leftarrow} f\del{\bm{y}; \bm{\theta}}}\) which runs the outermost derivative in forward mode as a Jacobian vector product.
This is reminiscent of the standard ``forward-over-reverse'' method of computing hessian-vector products.

In our experiments, this optimization gave a speedup of over \(5\times\) in terms of kernel gradients per second relative to the triple reverse baseline.
We expect that further speedups may be obtained by leveraging techniques similar to those of \cite{novak2021fast} and leave this direction for future work.

\subsection{Efficient greedy poison set selection}\label{apx:init}

Here we describe the greedy initial poison set selection algorithm from \cref{sec:ntba} in detail.
In \cref{eq:lin-add-1-point}, we note that we can write the difference in prediction on a test point \(\bm{x}_{\mathrm{a}}\) after a single poison point \(\bm{x}_{\mathrm{p}}\) has been added to the training set of a kernel regression model in closed form.
At each step of our greedy algorithm, we apply a vectorized form of \cref{eq:lin-add-1-point} in order to evaluate the predictions for the entire set \(X_{\mathrm{a}}\) under the addition of each poison in \(X_{\mathrm{p}}\).
We choose the candidate set of images \(X_{\mathrm{p}}\) to be the set of all clean images with the trigger added.
For convenience, we write \cref{eq:bd-loss} in terms of \(X_{\mathrm{dpta}}, \bm{y}_{\mathrm{dpta}}\),
\begin{equation}
  \mathcal{L}_{\mathrm{backdoor}}\del{X_{\mathrm{dpta}}, \bm{y}_{\mathrm{dpta}}} = \frac{1}{2}\norm[\big]{\bm{y}_{\mathrm{dp}}^\top K_{\mathrm{dp,dp}}^{-1} K_{\mathrm{dp,ta}} - \bm{y}_{\mathrm{ta}}}_2^2,
\end{equation}
making the evaluation of the kernel matrices implicit.
Then we write the greedy set selection explicitly in \cref{alg:greedy-init}.

\begin{algorithm}[h]
  \caption{Greedy subset selection}\label{alg:greedy-init}
  \DontPrintSemicolon
  \KwIn{Kernel matrix blocks \(K_{\mathrm{d,dta}}\), data subsets \(\del{X_{\mathrm{dpa}}, \bm{y}_{\mathrm{dpa}}}\), \(m \in \NN\).}
  \KwOut{Data subset \(X_{\mathrm{p}}', \bm{y}_{\mathrm{p}}'\) with \(\abs{X_{\mathrm{p}}'} = \abs{\bm{y}_{\mathrm{p}}'} = m\).}
  Initialize \(X^{\del{0}} \) and \(\bm{y}^{\del{0}}\) to be an empty matrix and vector respectively.\;
  \For{\(i \in \sbr{m}\)}{
    \(\del{\bm{x}, y} = \argmin_{\del{\bm x, y} \in D_{\mathrm{p}} \setminus D_{i-1}} \mathcal{L}_{\mathrm{backdoor}}\del*{\begin{bmatrix}X_{\mathrm{dta}} \\ X^{\del{i-1}} \\ \bm{x}\end{bmatrix}, \begin{bmatrix}\bm{y}_{\mathrm{dta}}\\ \bm{y}^{\del{i-1}} \\ y\end{bmatrix}}\)\label{alg:greedy-init:argmin}\;
    \(X^{\del{i}}\gets \begin{bmatrix} X^{\del{i-1}} \\ \bm{x}\end{bmatrix}\) and \(\bm{y}^{\del{i}} \gets \begin{bmatrix}\bm{y}^{\del{i-1}} \\ y\end{bmatrix}\)\;
  }
  \KwRet \(X_{\mathrm{p}}' = X^{\del{m}}, \bm{y}_{\mathrm{p}}' = \bm{y}^{\del{m}}\)\;
\end{algorithm}

The optimization of \cref{alg:greedy-init:argmin} can be computed efficiently by precomputing the \(K_{\mathrm{dp,dpta}}\) matrix and applying the vectorized form of \cref{eq:lin-add-1-point} which can be done in \(\mathcal{O}\del{n^3 + mn^2}\) where \(n = \abs{X_{\mathrm{d}}}\) and \(m = \abs{X_{\mathrm{p}}}\).

\subsection{Optimization details}\label{apx:opt-details}

We use L-BFGS-B by adapting the wrapper of \cite{virtanen2020scipy} for use with JAX.
We found that simple first order methods such as gradient descent with momentum and Adam \cite{kingma2015adam} converged very slowly with small learning rates and were unable to achieve good minima with larger learning rates.
In contrast, the strong Wolfe line search of L-BFGS-B appears to choose step sizes which lead to relatively rapid convergence for our problem.

\subsection{Computational resources}\label{apx:computational-resources}

All neural networks were trained on a single Nvidia 2080 Ti.
We ran NTBA optimization on a machine with four Nvidia A100 GPUs for a duration between 5 hours and 12 hours depending on the number of poisons being optimized.
Before optimization begins, we precompute the \(K_{\mathrm{d,dta}}\) matrix using Nvidia A100 GPUs, requiring a total of 2 GPU hours for double precision.

\section{Supplementary experimental results}\label{apx:supp-exp}

We report further experimental results complimenting those of \cref{sec:results}.

\subsection{Results for patch trigger on CIFAR-10}\label{apx:patch}

\begin{figure}[h]
  \centering
\adjustbox{clap}{\hspace{2.5em}
  \begin{subfigure}[t]{0.3\textwidth}
    \centering
    \includegraphics[width=0.3\textwidth]{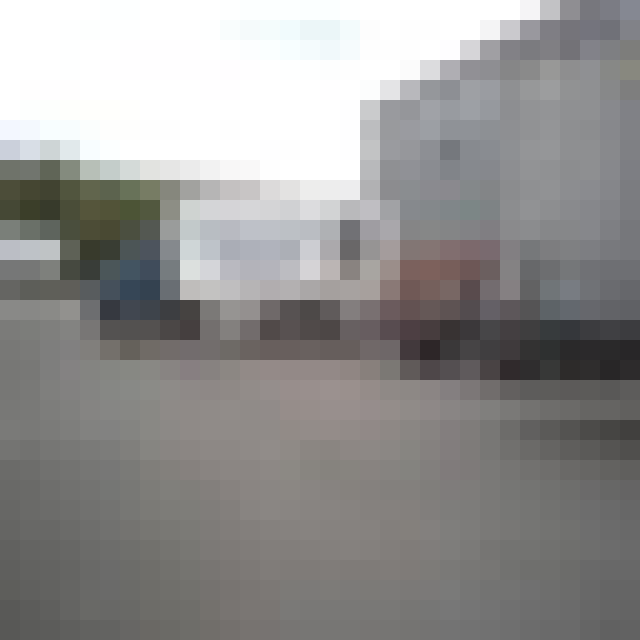}\hspace{0.1em}%
    \includegraphics[width=0.3\textwidth]{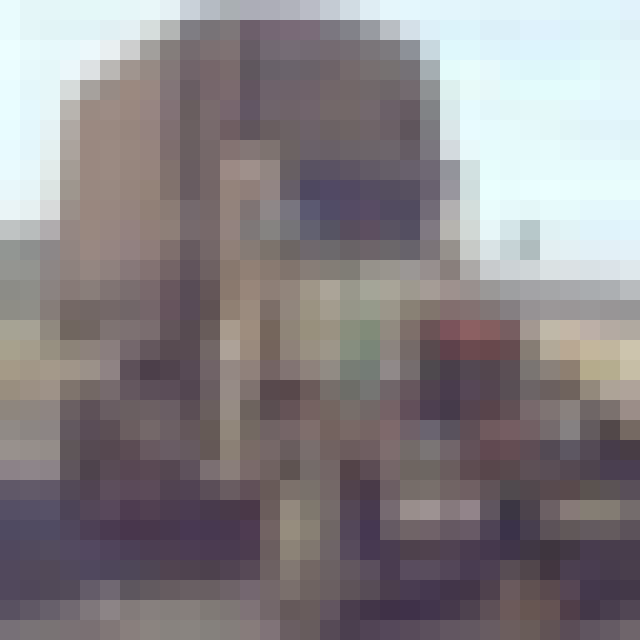}\hspace{0.1em}%
    \includegraphics[width=0.3\textwidth]{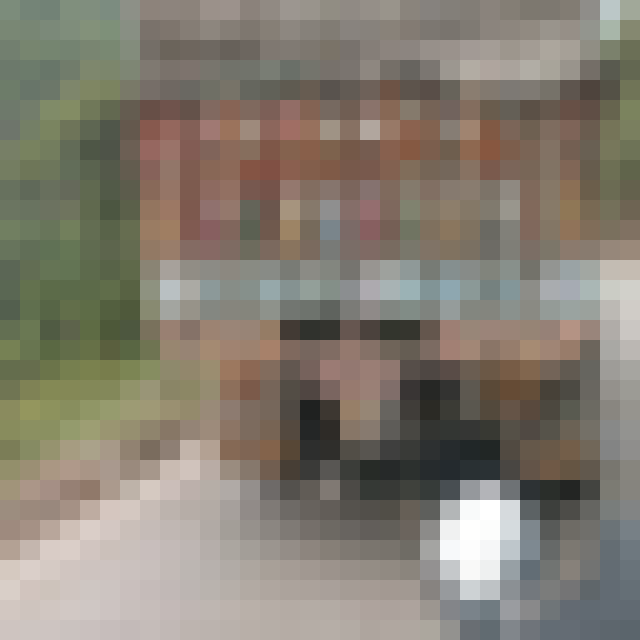}
    \put(-139,10){\footnotesize\rotatebox{90}{clean}}
    \\[0.0em]
    \includegraphics[width=0.3\textwidth]{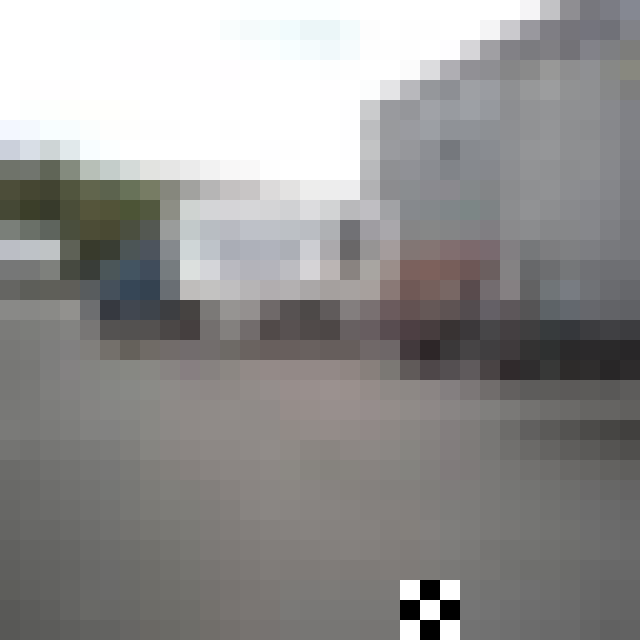}\hspace{0.1em}%
    \includegraphics[width=0.3\textwidth]{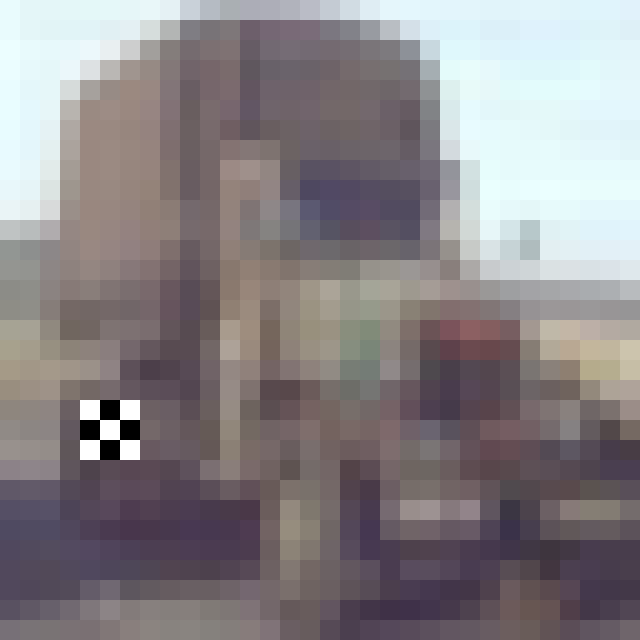}\hspace{0.1em}%
    \includegraphics[width=0.3\textwidth]{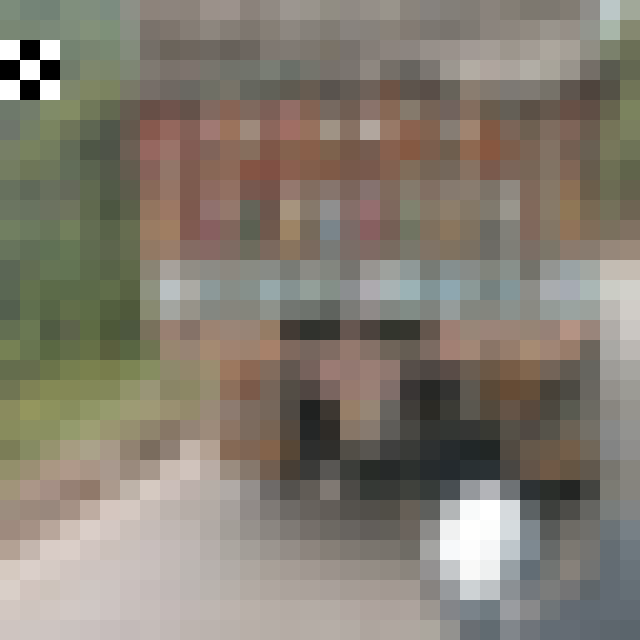}
    \put(-139,10){\footnotesize\rotatebox{90}{greedy}}
    \\[0.0em]
    \includegraphics[width=0.3\textwidth]{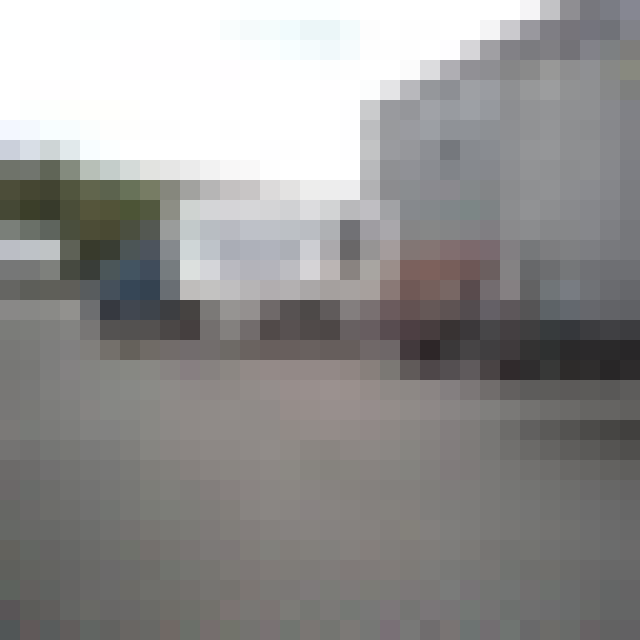}\hspace{0.1em}%
    \includegraphics[width=0.3\textwidth]{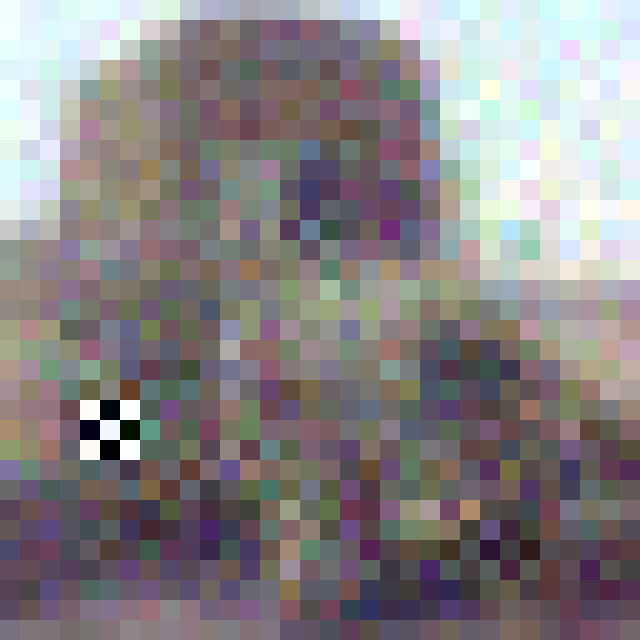}\hspace{0.1em}%
    \includegraphics[width=0.3\textwidth]{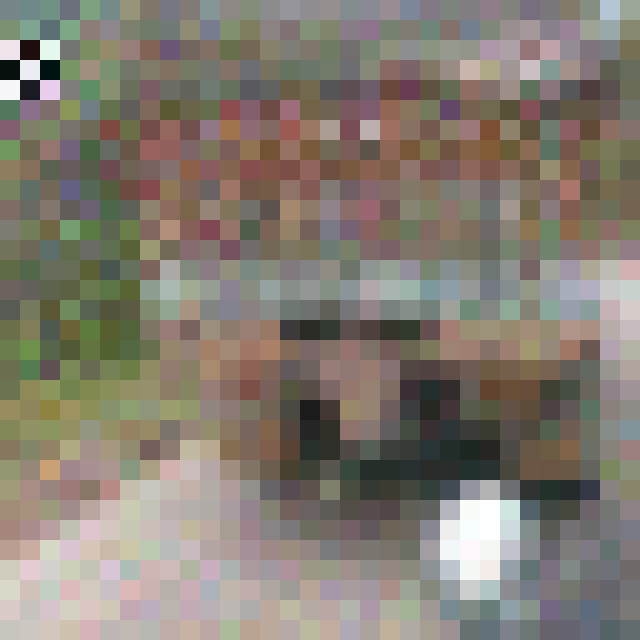}
    \put(-149,5){\footnotesize\rotatebox{90}{greedy+}}
    \put(-139,5){\footnotesize\rotatebox{90}{optimize}}
    \caption{\(m=3\)}
  \end{subfigure}\hspace{-0.06in}%
  \begin{subfigure}[t]{0.7\textwidth}
    \centering
    \includegraphics[width=0.12857\textwidth]{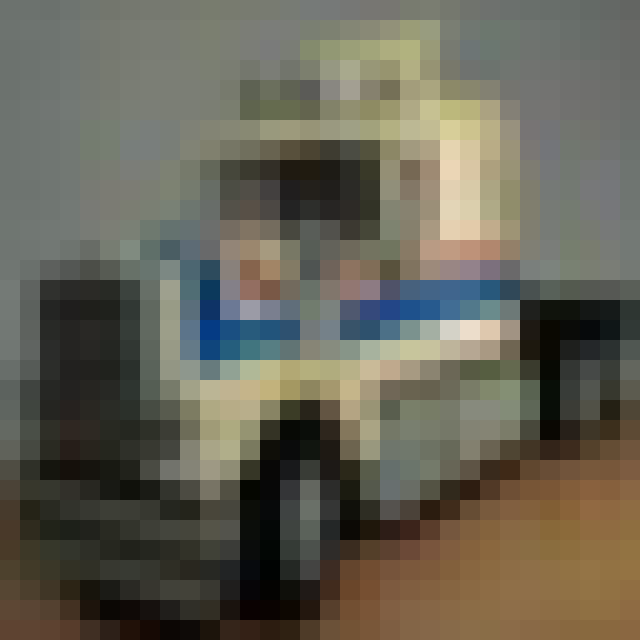}\hspace{0.1em}%
    \includegraphics[width=0.12857\textwidth]{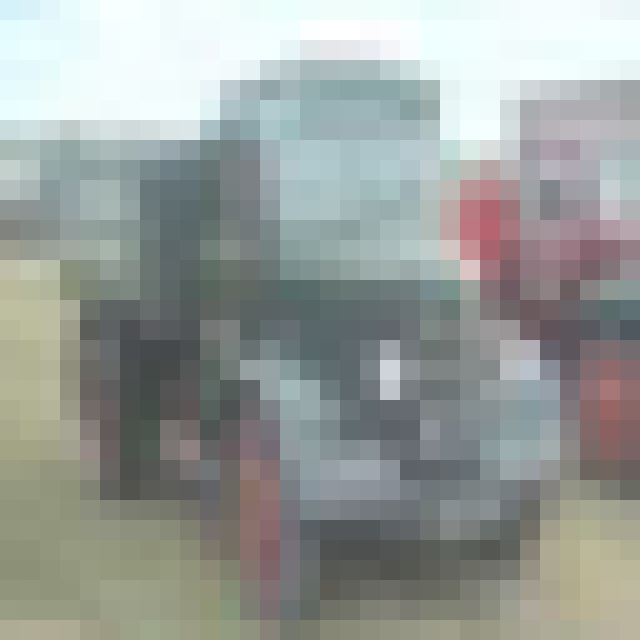}\hspace{0.1em}%
    \includegraphics[width=0.12857\textwidth]{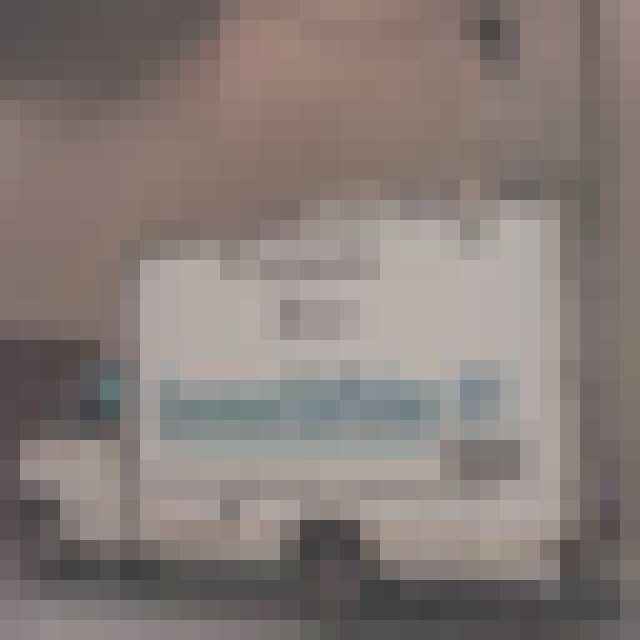}\hspace{0.1em}%
    \includegraphics[width=0.12857\textwidth]{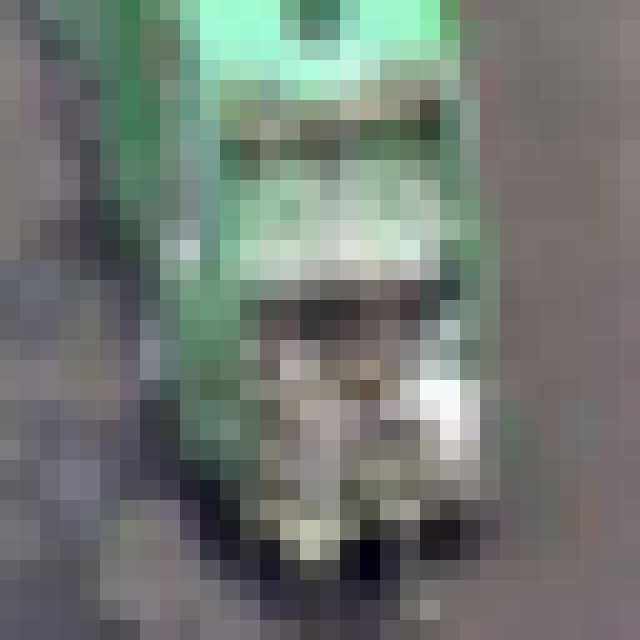}\hspace{0.1em}%
    \includegraphics[width=0.12857\textwidth]{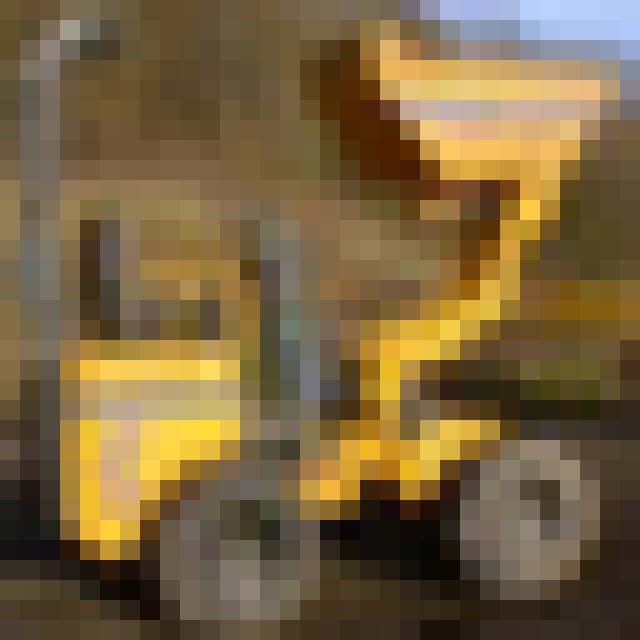}\hspace{0.1em}%
    \includegraphics[width=0.12857\textwidth]{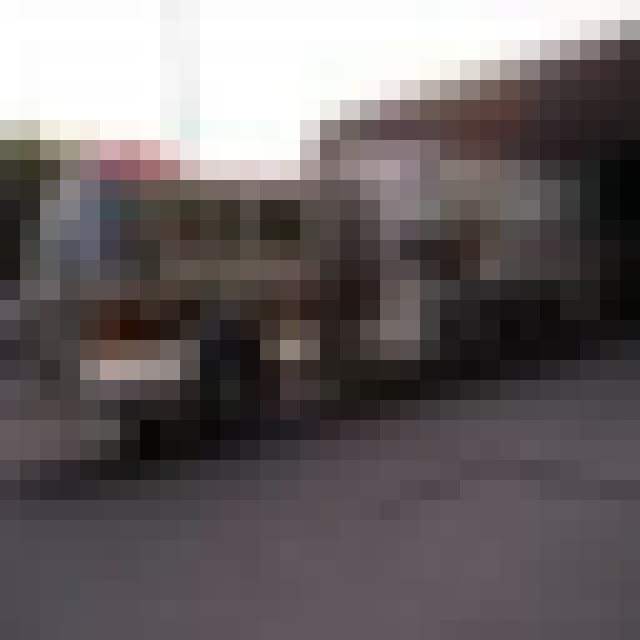}\hspace{0.1em}%
    \includegraphics[width=0.12857\textwidth]{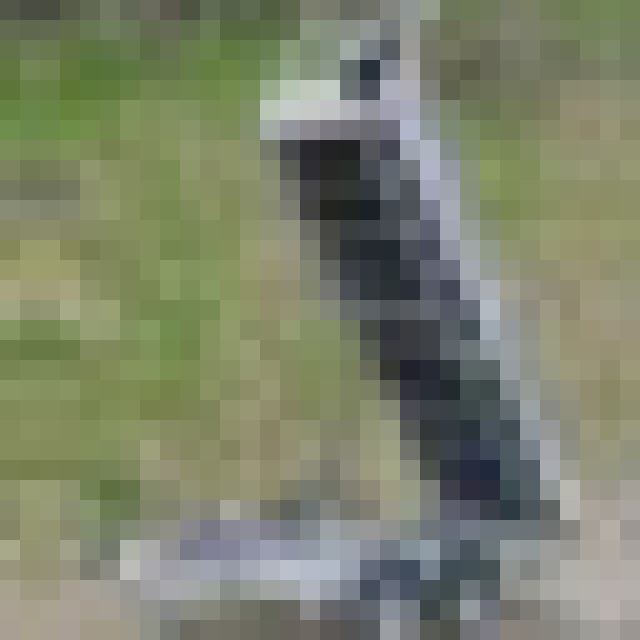}\\[0.0em]
    \includegraphics[width=0.12857\textwidth]{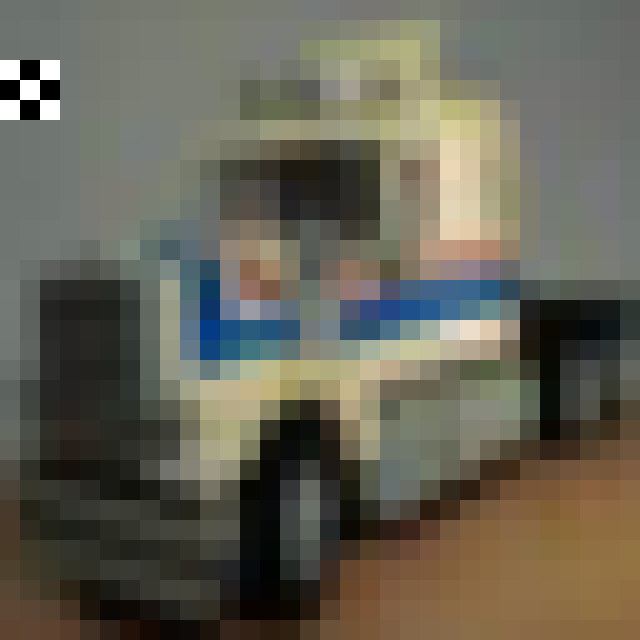}\hspace{0.1em}%
    \includegraphics[width=0.12857\textwidth]{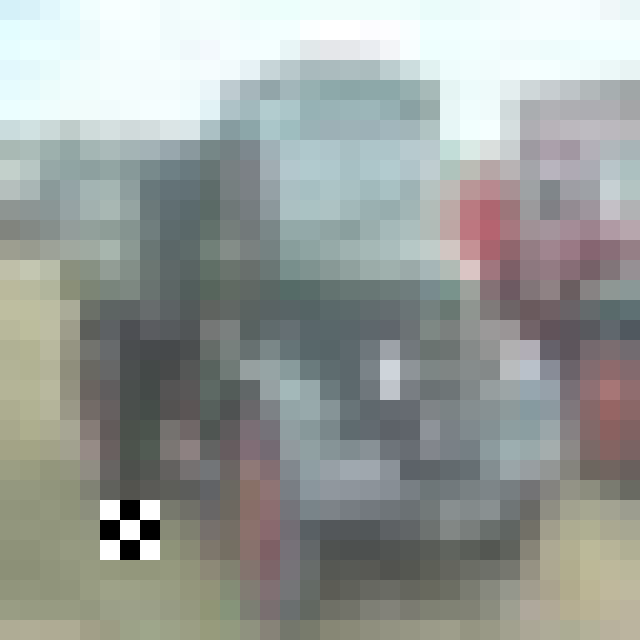}\hspace{0.1em}%
    \includegraphics[width=0.12857\textwidth]{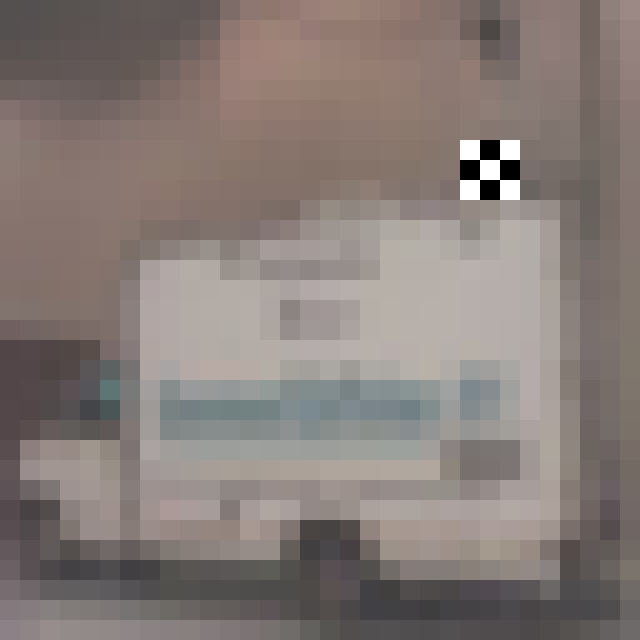}\hspace{0.1em}%
    \includegraphics[width=0.12857\textwidth]{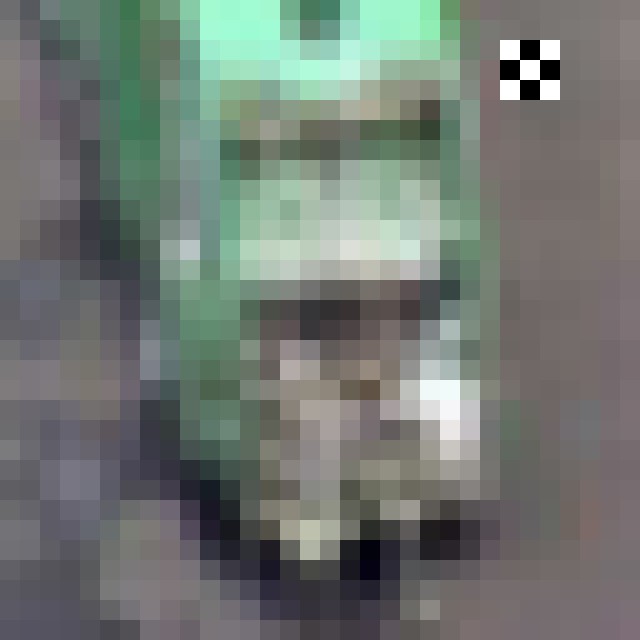}\hspace{0.1em}%
    \includegraphics[width=0.12857\textwidth]{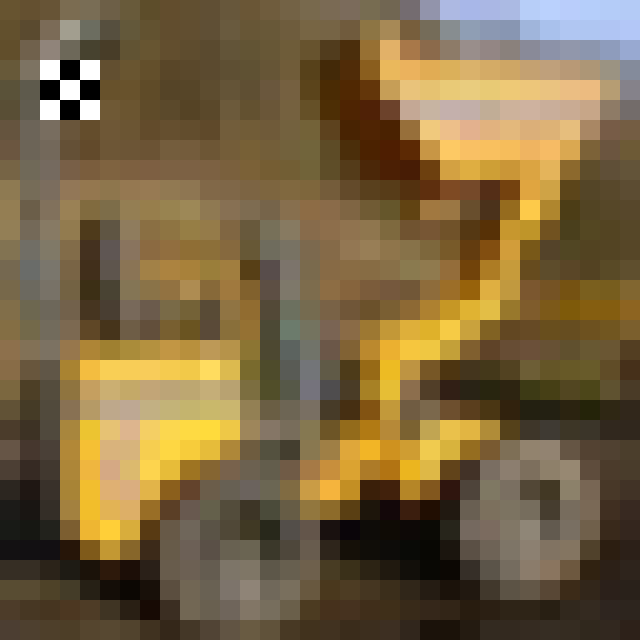}\hspace{0.1em}%
    \includegraphics[width=0.12857\textwidth]{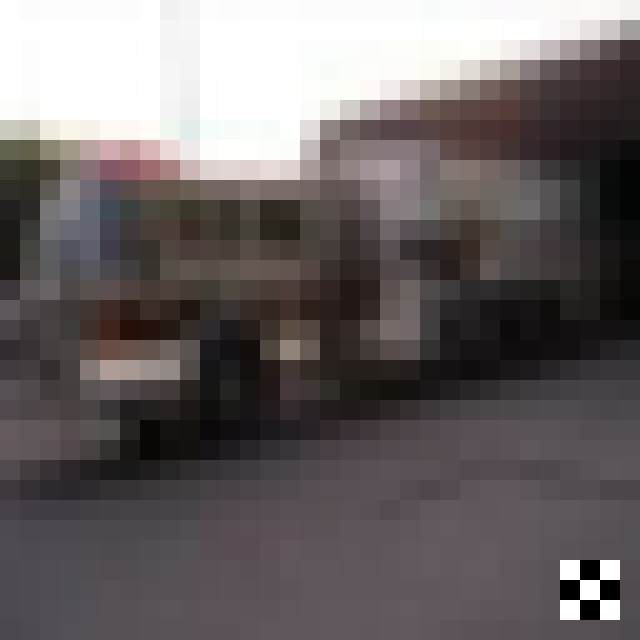}\hspace{0.1em}%
    \includegraphics[width=0.12857\textwidth]{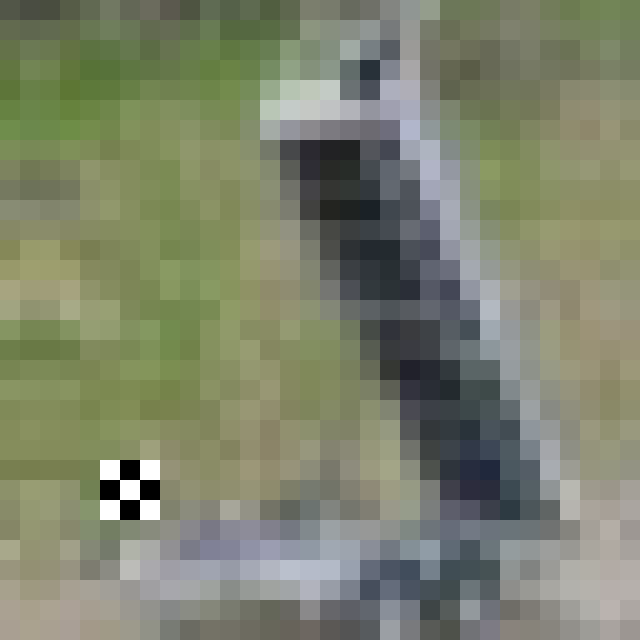}\\[0.0em]
    \includegraphics[width=0.12857\textwidth]{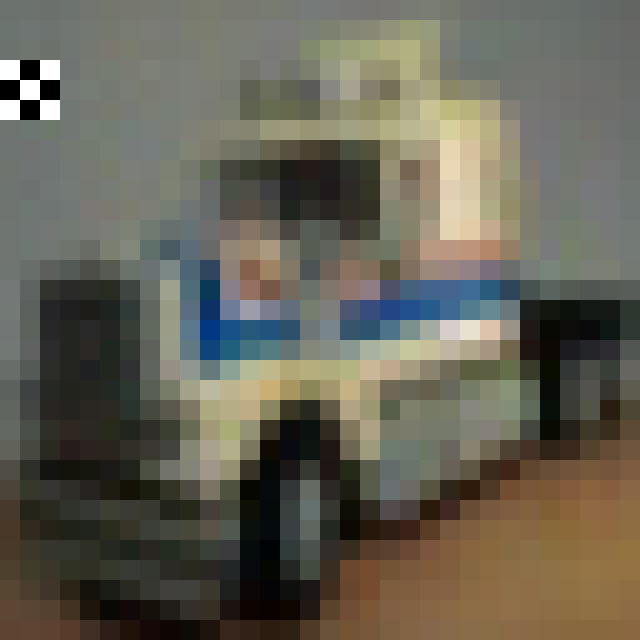}\hspace{0.1em}%
    \includegraphics[width=0.12857\textwidth]{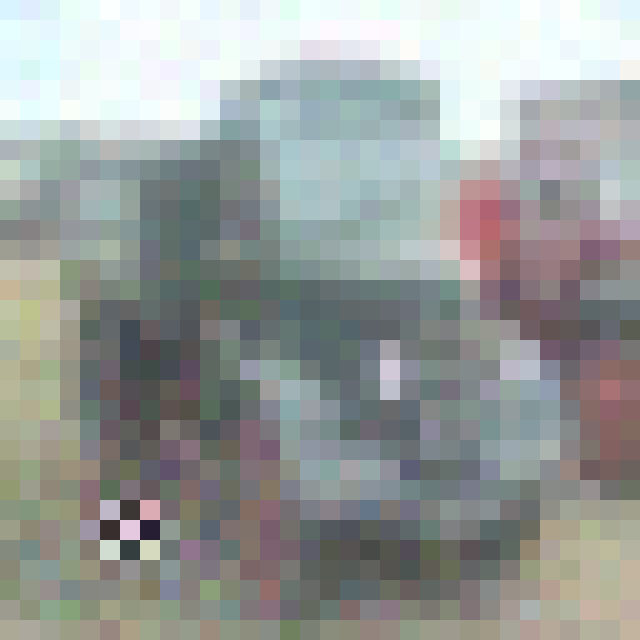}\hspace{0.1em}%
    \includegraphics[width=0.12857\textwidth]{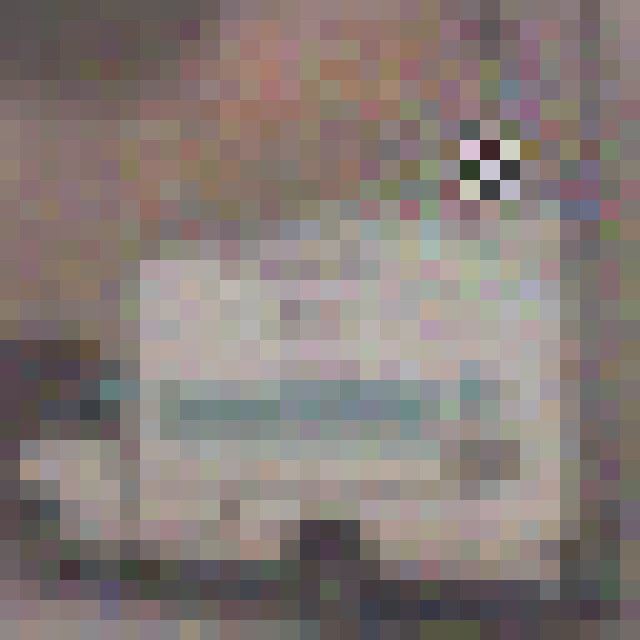}\hspace{0.1em}%
    \includegraphics[width=0.12857\textwidth]{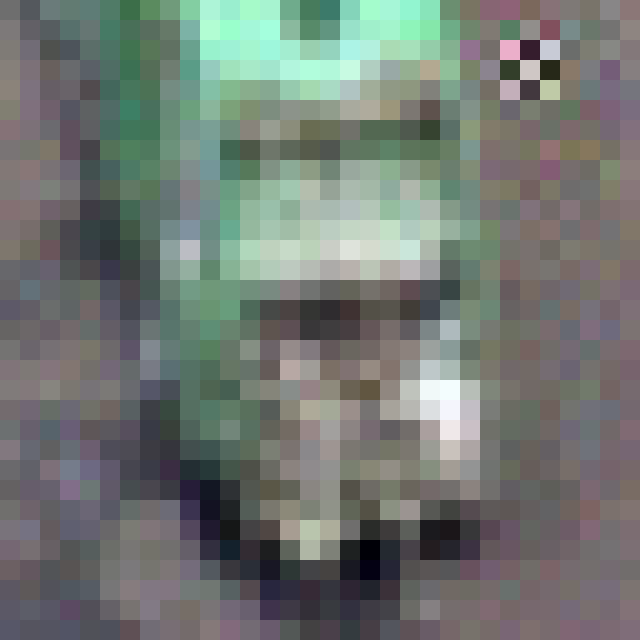}\hspace{0.1em}%
    \includegraphics[width=0.12857\textwidth]{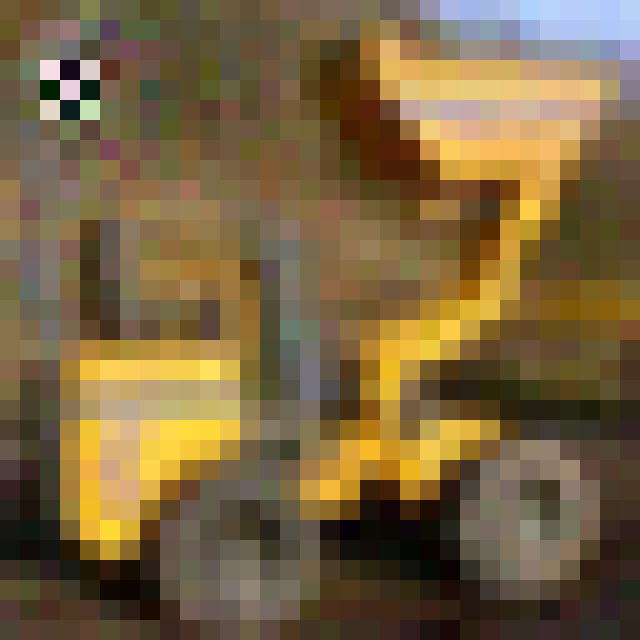}\hspace{0.1em}%
    \includegraphics[width=0.12857\textwidth]{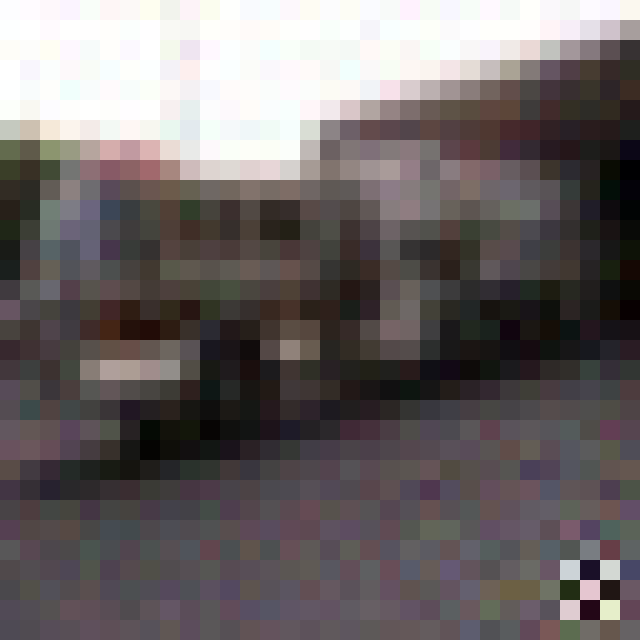}\hspace{0.1em}%
    \includegraphics[width=0.12857\textwidth]{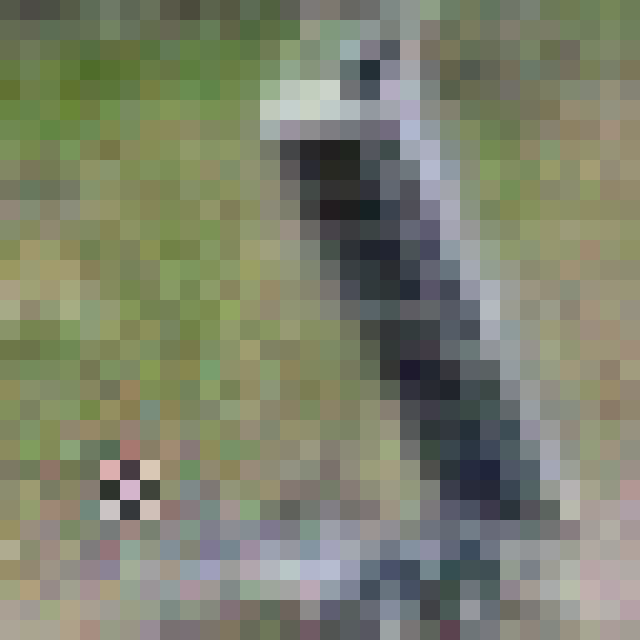}
    \caption{\(m=10\)}
  \end{subfigure}}
  \caption{
    Images produced by backdoor optimization for the patch trigger and \(m \in \set{3, 10}\).
    The top row shows the original clean image, the middle row shows the image with the trigger applied, and the bottom row shows the poisoned image after optimization. Duplicate images have been omitted to save space.}\label{fig:images-patch}
\end{figure}

\begin{figure}[h]
  \centering
  \includegraphics[width=0.4\textwidth]{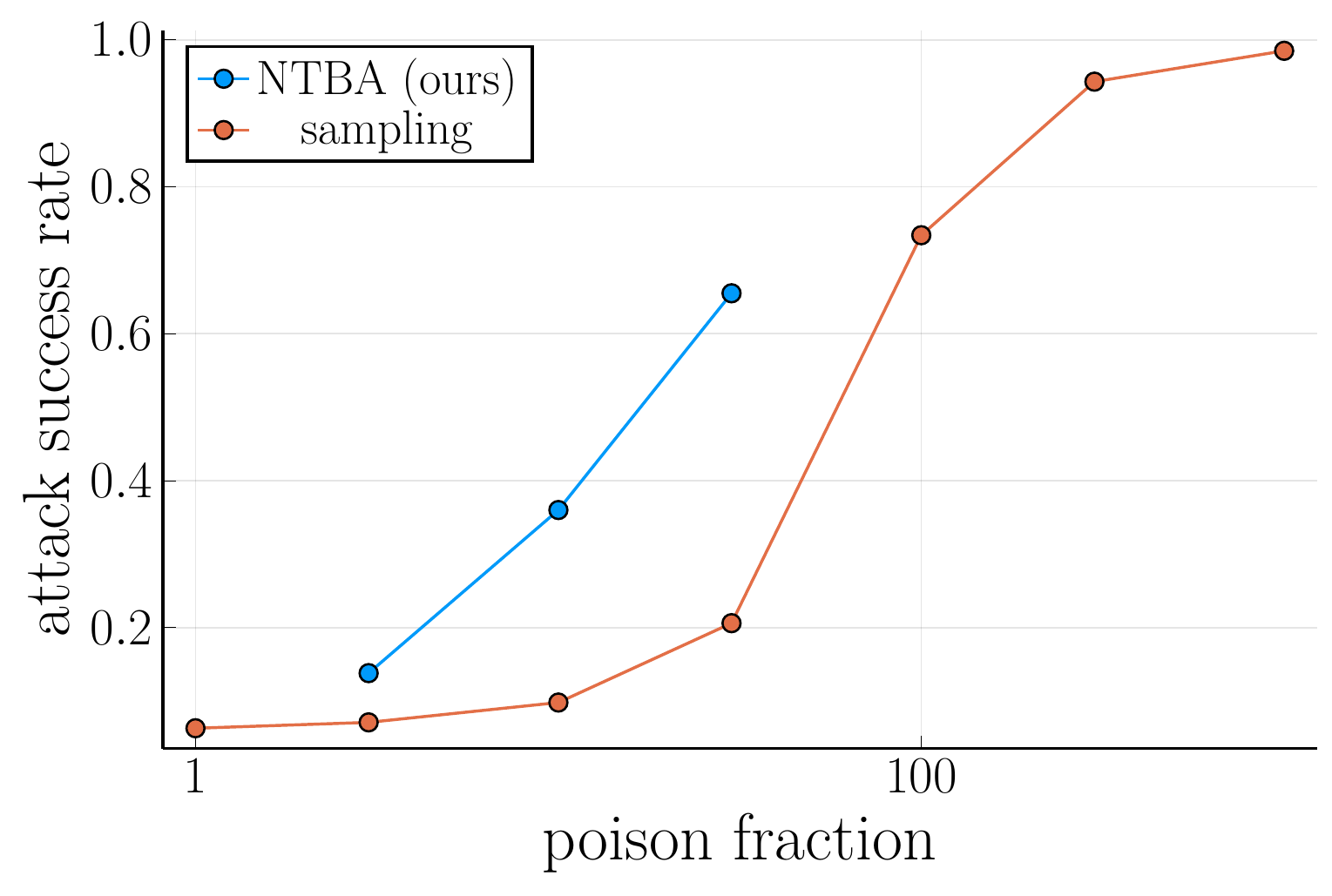}
  \caption{The trade-off between the number of poisons and ASR for the patch trigger.}\label{fig:asr-vs-eps-patch}
\end{figure}

We repeat the experiments of \cref{sec:results:main} using a \(3 \times 3\) checkered patch as the backdoor trigger.
Example images for this attack are shown in \cref{fig:images-patch}.
We plot the ASR vs. the number of poisoned images in \cref{fig:asr-vs-eps-patch} with numerical results reported in \cref{tab:asr-patch}.

We note that for some images in \cref{fig:images-patch}, the trigger becomes partially faded out after optimization while for other images the trigger remains unchanged.
We believe this may be due to the optimization getting stuck in a local minima nearby some images, preventing it from erasing the triggers as we would expect according to the analysis in \cref{sec:intuition}.
This may partly explain why the attacks computed for the patch trigger are not as strong as those computed for the periodic trigger.

\begin{table}[h]
  \centering
  \caption{ASR of NTBA ($\mathrm{asr}_{\mathrm{nn,te}}$) is significantly higher than the ASR for the  baseline of the sampling based attack using the same patch trigger, across a range of poison budgets \(m\).
    Clean accuracy \(\mathrm{acc}_{\mathrm{nn,te}}\) remains above \SI{92.6}{\percent} in all cases.}\label{tab:asr-patch}
  \vspace{1.5ex}
  \begin{tabular}[t]{r*{5}{S[table-format=-2.2,detect-weight,mode=text]S[table-format=-2.2,detect-weight,mode=text]}}
    \toprule
    & \multicolumn{4}{c}{ours} & sampling\\
    \(m\) & \(\mathrm{asr}_{\mathrm{ntk,tr}}\) & \(\mathrm{asr}_{\mathrm{ntk,te}}\) & \(\mathrm{asr}_{\mathrm{nn,tr}}\) & \(\mathrm{asr}_{\mathrm{nn,te}}\) & \(\mathrm{asr}_{\mathrm{nn,te}}\)\\
    \midrule
    3 & 99.9 & 74.1 & 0.9 & 13.8 & 7.1\\
    10 & 99.0 & 79.8 & 37.7 & 36.0 & 9.8\\
    30 & 93.8 & 82.3 & 66.8 & 65.5 & 20.6\\
    \bottomrule
  \end{tabular}\hspace{1em}%
  \begin{tabular}[t]{r*{1}{S[table-format=-2.2,detect-weight,mode=text]S[table-format=-2.2,detect-weight,mode=text]}}
    \toprule
    & sampling\\
    \(m\) & \(\mathrm{asr}_{\mathrm{nn,te}}\)\\
    \midrule
    0  & 6.2\\
    1  & 6.3\\
    100 & 73.4\\
    300 & 94.3\\
    1000 & 98.5\\
    \bottomrule
  \end{tabular}
\end{table}

\subsection{Results for periodic trigger on ImageNet}\label{apx:imagenet}

\begin{figure}[h]
  \centering
  \includegraphics[width=0.5\textwidth]{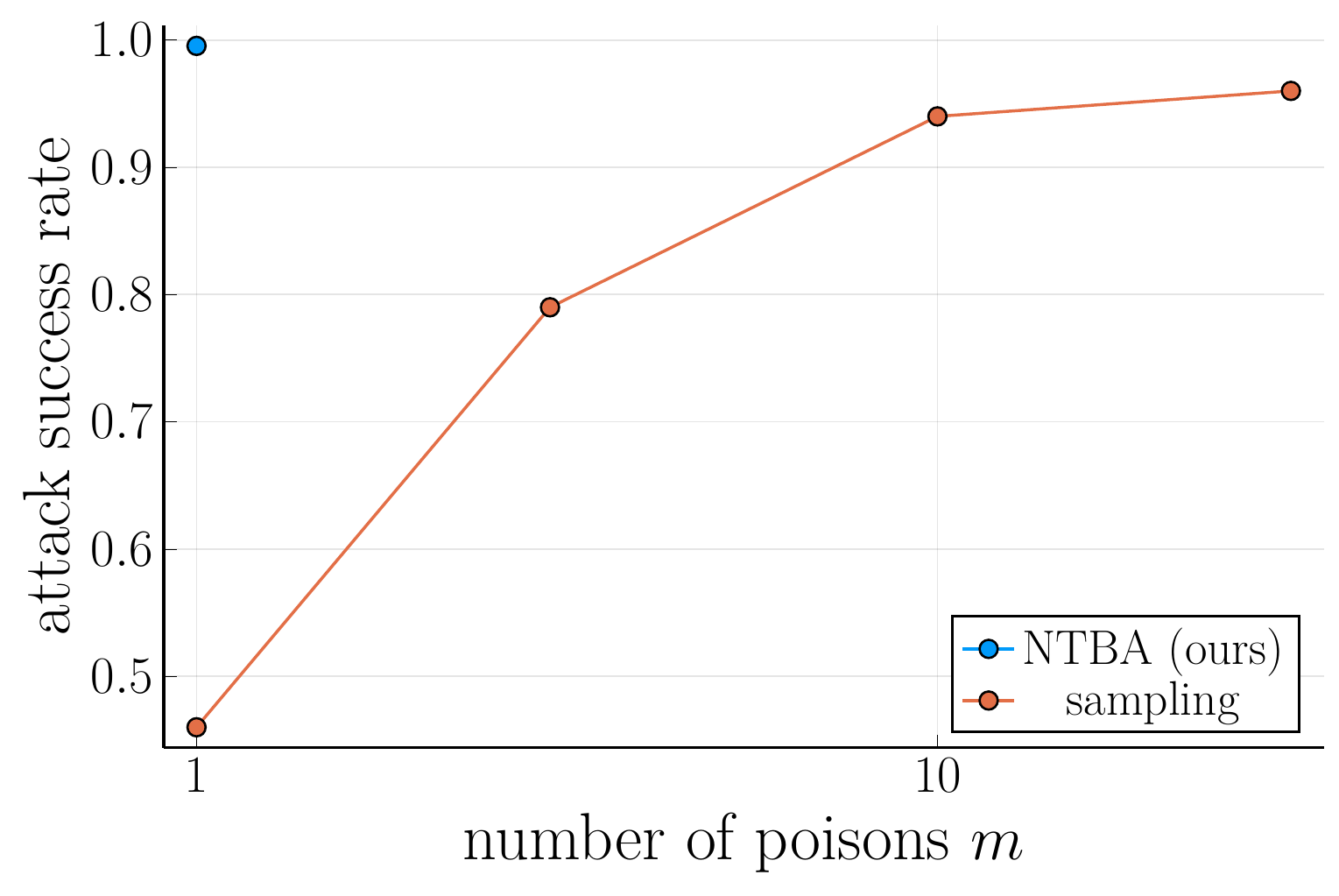}
  \caption{The trade-off between the number of poisons and ASR for ConvNeXt trained from scratch.}\label{fig:asr-vs-eps-imagenet}
\end{figure}

\begin{figure}[h]
  \centering
  \includegraphics[width=0.5\textwidth]{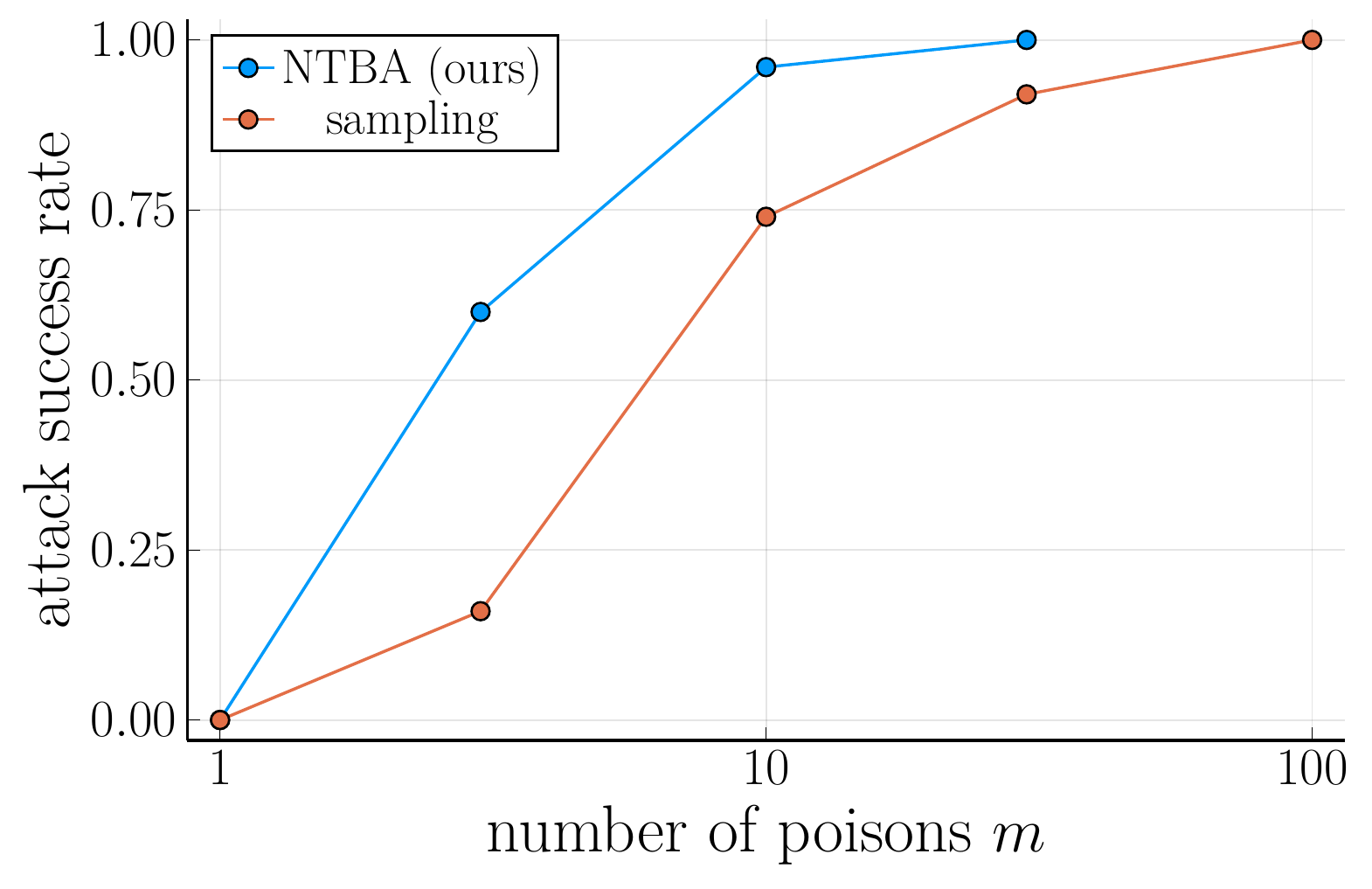}
  \caption{The trade-off between the number of poisons and ASR for ConvNeXt pretrained on ImageNet.}\label{fig:asr-vs-eps-imagenet-pretrained}
\end{figure}

We also use NTBA to attack a ConvNeXt-tiny \cite{liu2022convnet} (\(d\approx \num{2.8e7}\)) trained on a 2 label subset of ImageNet.
We use ``slot'' as the source label and ``Australian terrier'' as the target label following the examples from \cite{saha2020hidden}.
We consider both the case where the ConvNeXt is initialized randomly and trained from scratch and the case where it has been pretrained on ImageNet and fine-tuned as in \cite{saha2020hidden}.
The results for these two settings are shown in \cref{fig:asr-vs-eps-imagenet,fig:asr-vs-eps-imagenet-pretrained} respectively.
When trained from scratch, the clean accuracy of the ConvNeXt remains above 90\% in all cases.
When pretrained and fine-tuned, the ConvNeXt achieves at least 99\% clean accuracy in all cases.

We note that ConvNeXt is surprisingly vulnerable to backdoors when trained from scratch, as even a single random poisoned image is sufficient to achieve 50\% ASR and NTBA is able to achieve 100\% ASR with a single image.
With pretraining, the ConvNeXt becomes slightly more resistant to backdoors, but the periodic attack remains quite strong. We give numerical results in \cref{tab:asr-imagenet}.

\begin{table}[h] 
  \centering
  
  \begin{subtable}[t]{0.45\textwidth}
    \centering
    \begin{tabular}{rrr}
    \toprule
         \(m\) &  NTBA & sampling\\
         \midrule
         0 &  & 10\\
         1 & 100 & 46\\
         3 &  & 78\\
         10 &  & 94\\
         30 &  & 96\\
         \bottomrule
    \end{tabular}
    \caption{trained from scratch}
    \label{tab:asr-imagenet-from-scratch}
  \end{subtable}%
  \begin{subtable}[t]{0.45\textwidth}
    \centering
    \begin{tabular}{rrr}
       \toprule
         \(m\) &  NTBA & sampling\\
         \midrule
         0 &  & 0\\
         1 & 0 & 0\\
         3 & 60 & 16\\
         10 & 96 & 74\\
         30 &  100 & 92\\
         100 &  & 100 \\
         \bottomrule
    \end{tabular}
    \caption{pretrained}
    \label{tab:asr-imagenet-pretrained}
  \end{subtable}
  \caption{
  \(\mathrm{asr}_{\mathrm{nn,te}}\) results for ConvNeXt on ImageNet. 
  Numbers are percentages over the 50 examples from the source label.
  }\label{tab:asr-imagenet}
\end{table}

\subsection{Transfer and generalization of NTBA}\label{sec:transfer}
 
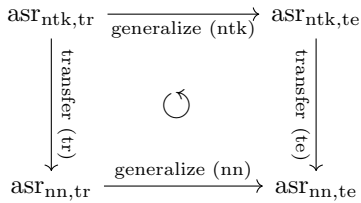
\begin{figure}[h]
 \vspace{-0.2cm}
    \centering
        \begin{tikzcd}[row sep=0.7in,column sep=0.8in]
        {\mathrm{asr}_{\mathrm{ntk,tr}}} \arrow[r, "\textrm{generalize (ntk)}"'] \arrow[d, "\textrm{transfer (tr)}" {anchor=south, rotate=-90}] &  {\mathrm{asr}_{\mathrm{ntk,te}}} \arrow[d, "\textrm{transfer (te)}"' {anchor=north, rotate=-90}] \\
        {\mathrm{asr}_{\mathrm{nn,tr}}} \arrow[r, "\textrm{generalize (nn)}"]\arrow[ur, phantom, "\scalebox{1.5}{$\circlearrowleft$}" description]                                        & {\mathrm{asr}_{\mathrm{nn,te}}}                                       
        \end{tikzcd}
    \captionof{figure}{Relationship between the columns of \cref{tab:asr,tab:asr-patch}.}\label{fig:asr-cd}
  \end{figure}

\cref{fig:asr-cd} illustrates two important steps which separate the performance achieved by the optimization, \(\mathrm{asr}_{\mathrm{ntk,tr}}\), (which consistently achieves \SI{100}{\percent} attack success rate) and the final attack success rate of the neural network, \(\mathrm{asr}_{\mathrm{nn,te}}\): transfer from the NTK to the neural network and generalization from poison examples seen in training to new ones.
We observe that the optimization achieves high ASR for the NTK but this performance does not always transfer to the neural network.

Interestingly, we note that the attack transfers very poorly for training examples, so much so that the generalization gap for the attack is negative for the neural network.
We believe this is because it is harder to influence the predictions of the network nearby training points.
Investigating 
this transfer performance presents an interesting open problem for future work.

\subsection{Choice of weights for the empirical NTK}\label{sec:ntba-epoch}

\begin{figure}[h]
  \centering
  \includegraphics[width=0.4\textwidth]{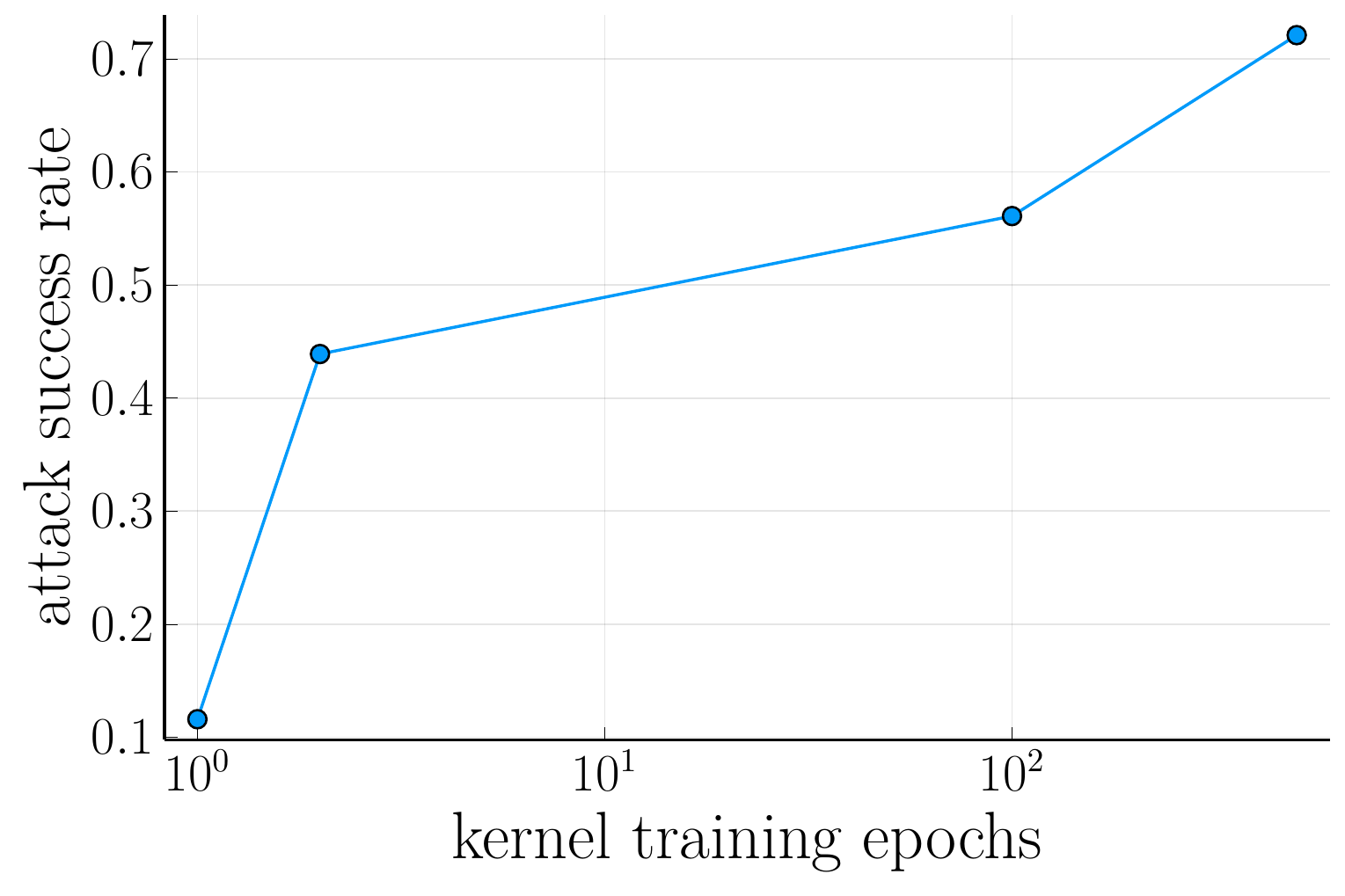}
  \caption{
    Plot showing \(\mathrm{asr}_{\mathrm{nn,tr}}\) vs. the number of epochs used to train the network before the weights were frozen for use in the empirical NTK.
    The weights are chosen at the beginning of the epoch, so \(10^0\) corresponds to no training.
  }\label{fig:asr-vs-epoch}
\end{figure}

In our main experiments we chose to use the weights of the network after full convergence for use with the empirical neural tangent kernel.
In \cref{fig:asr-vs-epoch} we show the results we obtain if we had used the network weights at other points along the training trajectory.
At the beginning of training, there is a dramatic increase in ASR after a single epoch of training and training longer is always better until we reach convergence.
At 500 epochs the loss of the network falls below \num{1e-7}, and the network effectively does not change from then on.
These results mirror those of \cite{fort2020deep,long2021properties}, which find that the empirical neural tangent kernel's test accuracy on standard image classification rapidly improves at the beginning of training and continues to improve as training progresses.

\section{Kernel perspective on the vulnerability of NNs}\label{sec:kernel:1d}

In \cref{fig:ntk-vs-laplace-normal,fig:ntk-vs-laplace-tangent}, as a simple example, we consider the infinite width neural tangent kernel of a 3 layer feed-forward neural network with ReLU activations.
Recently, \citep{geifman2020similarity,chen2021deep} showed that the neural tangent kernel of feed-forward neural networks are equivalent to Laplace kernels \(K^{\mathrm{lap}}\del{\bm{x}, \bm{y}} = \exp\del{\norm{\bm{x} - \bm{y}}/\sigma}\) for inputs lying on the unit sphere.
For our choice of NTK, we compare against a Laplace kernel with \(\sigma \approx 6.33\), that closely matches the NTK around \(x=0\) in \cref{fig:ntk-vs-laplace-tangent}.
For inputs that do not lie on the sphere, the kernels behave differently, which we illustrate in \cref{fig:ntk-vs-laplace}.

\end{document}